\documentclass{article}

\usepackage{fontawesome5}
\usepackage{adjustbox}
\usepackage{siunitx}
\usepackage{xspace}
\usepackage{amsmath}
\usepackage{subcaption}
\usepackage{algorithm}
\usepackage{algpseudocode}

\usepackage[table]{xcolor}
\usepackage{booktabs}
\usepackage{makecell}
\usepackage{pgfmath}
\usepackage{listings}
\usepackage[most]{tcolorbox}

\definecolor{answerColor}{HTML}{91B5DA}
\definecolor{hallucColor}{HTML}{9BCDB8}
\definecolor{concColor}{HTML}{B9ABD6}
\definecolor{tokenColor}{HTML}{B8B8B8}
\definecolor{divColor}{HTML}{E8B294}

\def\metriccell#1#2#3{%
  \pgfmathtruncatemacro{\metricpercent}{%
    10+80*min(1,max(0,(#2-1)/4))}%
  \edef\metricshade{#1!\metricpercent!white}%
  \expandafter\cellcolor\expandafter{\metricshade}%
  #3%
}

\def\answercell#1{\metriccell{answerColor}{#1}{#1}}
\def\answerbest#1{\metriccell{answerColor}{#1}{\textbf{#1}}}

\def\halluccell#1{\metriccell{hallucColor}{#1}{#1}}
\def\hallucbest#1{\metriccell{hallucColor}{#1}{\textbf{#1}}}

\def\conccell#1{\metriccell{concColor}{#1}{#1}}
\def\concbest#1{\metriccell{concColor}{#1}{\textbf{#1}}}

\def\divcell#1{\metriccell{divColor}{#1}{#1}}
\def\divbest#1{\metriccell{divColor}{#1}{\textbf{#1}}}

\def\tokenheat#1#2{%
  \pgfmathtruncatemacro{\tokenpercent}{%
    10+80*min(1,max(0,(736-#1)/(736-149)))}%
  \edef\tokenshade{tokenColor!\tokenpercent!white}%
  \expandafter\cellcolor\expandafter{\tokenshade}%
  #2%
}
\def\tokencell#1{\tokenheat{#1}{#1}}
\def\tokenbest#1{\tokenheat{#1}{\textbf{#1}}}

\usepackage{pifont}
\usepackage[table]{xcolor} 

\definecolor{checkGreen}{HTML}{387A50}
\definecolor{crossRed}{HTML}{C41230}

\providecommand{\cmark}{}
\providecommand{\xmark}{}
\renewcommand{\cmark}{\textcolor{checkGreen}{\ding{51}}}
\renewcommand{\xmark}{\textcolor{crossRed}{\ding{55}}}

 \usepackage[preprint]{neurips_2026}

\usepackage[utf8]{inputenc} 
\usepackage[T1]{fontenc}    
\usepackage{hyperref}       
\usepackage{url}            
\usepackage{booktabs}       
\usepackage{amsfonts}       
\usepackage{nicefrac}       
\usepackage{booktabs}
\usepackage{amssymb}
\usepackage{pifont}
\usepackage{multirow}
\usepackage{booktabs,multirow,makecell}

\usepackage{microtype}      
\usepackage{xcolor}         
\usepackage{graphicx}
\usepackage{booktabs}
\usepackage{multirow}
\usepackage{booktabs,multirow,makecell}
\usepackage{colortbl}
\definecolor{liza}{RGB}{153, 0, 0}
\definecolor{cmured}{RGB}{196, 18, 48}

\title{{\color{cmured}{\color{cmured} ANCHOR}:} {\color{cmured}A}udio-Visually Grou{\color{cmured}n}ded {\color{cmured}C}hain-of-T{\color{cmured}ho}ught {\color{cmured}R}easoning Benchmark}

\author{%
Joel Julin\textsuperscript{1,*} \quad
Souraja Kundu\textsuperscript{1,*} \quad
Liza Dahiya\textsuperscript{1} \quad
George Z. Wei\textsuperscript{1} \quad
Liuyue Xie\textsuperscript{2}
\\
\bfseries
Jingru Yi\textsuperscript{2} \quad
Xu Zhang\textsuperscript{2} \quad
Hao Yang\textsuperscript{2} \quad
Laszlo A. Jeni\textsuperscript{1}
\\[0.75em]
\textsuperscript{1}Carnegie Mellon University \hspace{0.35em}
\textsuperscript{2}Amazon
\\[0.35em]
\textsuperscript{*}Equal contribution
\\[0.5em]
{\small
\faAnchor\,
\href{https://anchorbenchmark.github.io}{%
  \texttt{anchorbenchmark.github.io}%
}
}
}

\begin{document}

\maketitle

\begin{abstract}

Multimodal large language models (MLLMs) perform well on audio-visual question answering, but correct answers do not necessarily imply that their stated reasoning is grounded in relevant perceptual evidence. Existing benchmarks mainly evaluate final-answer accuracy, offering limited insight into whether model-generated rationales correctly identify visual regions, temporal intervals, and audio events supporting predictions. We introduce \textbf{{\color{cmured}ANCHOR}}, a benchmark and evaluation framework for measuring grounded-rationale alignment in multimodal reasoning, comprising 2,319 videos and 8,272 question--answer pairs with structured reasoning traces, temporal evidence, spatial bounding boxes, and salient audio cues, including a human-refined test set of 224 videos and 975 question--answer pairs. We separately measure answer correctness, hallucination, logical divergence, and conciseness, with direct spatial and temporal grounding measurements on a focused subset. Across the full benchmark, twelve contemporary MLLMs show a substantial alignment gap: models with comparable answer performance can differ markedly in whether their explanations are supported by annotated evidence. On a matched subset of 100 videos and 446 question--answer pairs, rankings remain highly stable across human-refined, Gemini-generated, and GPT-generated references (mean Pearson r=0.99), providing evidence against model-reference circularity within this subset. Grounding-cue ablations of three MLLMs on the full human-refined test set reveal a consistent modality hierarchy: removing video reduces accuracy by 38.4--41.8 points, while removing bounding boxes (3.2--7.7 points) and audio (4.2--5.6 points) causes smaller, complementary, model-dependent drops. These findings show answer-only evaluation obscures important grounding failures and position \textbf{\color{cmured}ANCHOR} as a testbed for evaluating accurate and evidence-aligned audio-visual reasoning.

\end{abstract}

\section{Introduction}
\label{sec:intro}
Multimodal Large Language Models (MLLMs) have demonstrated impressive capabilities in Chain-of-Thought (CoT) reasoning across text-centric benchmarks~\cite{wei2022chain, zhang2023multimodal}, showing an emergent ability to decompose complex problems into logical intermediate steps. However, real-world reasoning is inherently multimodal; humans interpret social interactions, infer causality, and recognize emotions by seamlessly integrating visual cues, acoustic signals, and linguistic context.

In contrast, current MLLMs often default to dominant textual inputs, such as subtitles or biased question phrasing, producing answers that may be correct but are weakly grounded in perceptual evidence~\cite{chen2024quantifying, zheng2025unveiling}. This discrepancy exposes a fundamental limitation: existing evaluations predominantly focus on final answer accuracy, which rarely tests whether models genuinely reason over visual and audio signals or merely exploit spurious linguistic correlations.

This matters beyond simple accuracy: a model that reaches the correct conclusion through ``hallucinated'' or ungrounded reasoning is exploiting dataset patterns rather than generalizing, leaving it vulnerable to shortcut learning in high-stakes, complex multimodal scenarios.


\begin{figure}[!t]
    \centering
    \includegraphics[width=0.99\linewidth]{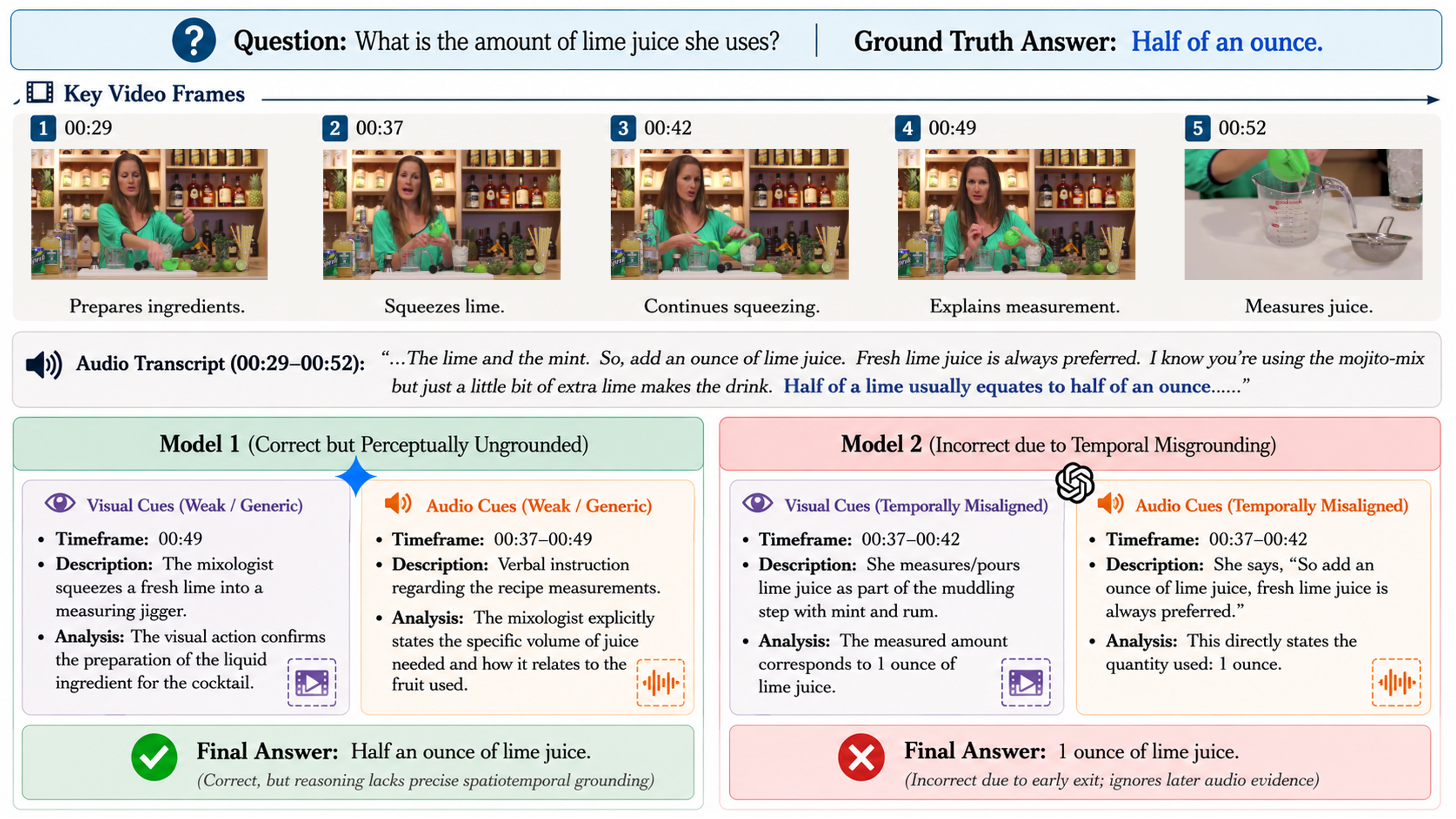}
\caption{\textbf{Limitations in MLLM CoT reasoning.} \textsc{Gemini-3-Flash} answers correctly but lacks perceptual grounding, while \textsc{GPT-5.2} misses late audio evidence and answers incorrectly. \textbf{\color{cmured}ANCHOR} exposes these failures by evaluating alignment between reasoning and sensory evidence.}
    \label{fig:intro_figure}
\vspace{-0.25cm}
\end{figure}

Figure~\ref{fig:intro_figure} illustrates these critical failure modes. In the first case, Gemini-3-Flash~\cite{gemini3flash_google_2025} achieves the correct answer but generates a verbose, uninformative CoT that serves as ``filler'' text rather than providing the causal perceptual evidence required for the conclusion. Conversely, \textsc{GPT-5.2}~\cite{openai_gpt52_2025} demonstrates a temporal grounding failure; while its initial logic is sound, the model performs an ``early exit,'' stopping its audio-visual processing prematurely and providing an incorrect answer based on incomplete evidence. These examples underscore that neither final accuracy nor explanation length guarantees true perceptual faithfulness.

To address this gap, we introduce \textbf{{\color{cmured} ANCHOR}}, a large-scale benchmark designed to explicitly evaluate how MLLMs process, align, and reason over multimodal input streams through structured intermediate representations. Our benchmark comprises 2,319 videos and 8,272 question-answer pairs spanning diverse domains, including social interactions, egocentric perception, and instructional tasks, each densely labeled with spatial bounding boxes (BB) generated via SAM3~\cite{carion2025sam3segmentconcepts} and fine-grained visual and audio event cues. Unlike evaluations that score only the final answer, \textbf{{\color{cmured} ANCHOR}} requires models to generate a \textit{Grounded Chain-of-Thought} that explicitly cites the precise timeframes and evidence supporting each reasoning step, enabling direct assessment of whether a model's reasoning is grounded in sensory evidence or reflects superficial textual pattern-matching.


\section{Related Works}

\subsection{Chain-of-Thought Reasoning.}
Chain-of-thought (CoT) prompting elicits intermediate reasoning steps before a final answer and can improve performance on arithmetic, symbolic, and logical tasks~\cite{wei2022chain}. Related approaches improve robustness through self-consistency~\cite{wang2023selfconsistencyimproveschainthought}, externalize intermediate computations through scratchpads and executable programs~\cite{nye2021workscratchpadsintermediatecomputation,chen2023programthoughtspromptingdisentangling}, or combine reasoning with tool use and structured search~\cite{yao2023reactsynergizingreasoningacting,yao2023treethoughtsdeliberateproblem}.

Most CoT methods, however, are designed for text-only settings. In perceptual tasks involving spatial localization, temporal dynamics, and asynchronous audio--visual cues, a plausible textual rationale does not establish that the model identified the correct evidence~\cite{li2022hero,pramanick2025enrich}. \textbf{\color{cmured}ANCHOR} addresses this limitation by requiring generated rationales to reference supporting video intervals, audio events, and spatially localized entities. These observable rationales can then be compared with annotated perceptual evidence.

\begin{table*}[!htbp]
\centering
\caption{\textbf{Benchmark Comparison.} \textbf{\color{cmured} ANCHOR} is the first to integrate dense spatial grounding (\textbf{BB}) with human-verified reasoning traces (\textbf{CoT}) for evidence-linked evaluation. \textbf{Grounding:} \textbf{Vid} (video), \textbf{Aud} (audio), \textbf{BB} (bounding box). \textbf{Properties:} \textbf{Dom.} (diverse domains), \textbf{Type} (Real-world/Synthetic), \textbf{Audio} (\textbf{Sp}eech, \textbf{So}und, \textbf{Mu}sic), \textbf{Size} (Total QA pairs), \textbf{Dur.} (video duration in secs).}

\scriptsize
\label{tab:full_benchmark_comparison}
\begin{tabular}{l cccc | ccccc}
\toprule
& \multicolumn{4}{c}{\textbf{Grounding Signals}} & \multicolumn{5}{c}{\textbf{Dataset Properties}} \\
\cmidrule(lr){2-5} \cmidrule(lr){6-10}
\textbf{Benchmark} & \textbf{Vid} & \textbf{Aud} & \textbf{BB} & \textbf{CoT} & \textbf{Dom.} & \textbf{Type} & \textbf{Audio} & \textbf{Size} & \textbf{Dur.} \\
\midrule
AVQA~\cite{yang2022avqa} & \cmark & \cmark & \xmark & \xmark & \xmark & R & So & 57K & 10 \\
MMAU~\cite{sakshi2024mmau} & \xmark & \cmark & \xmark & \xmark & \cmark & / & Sp+So+Mu & 10K & / \\
DAVE~\cite{radevski2025dave} & \cmark & \cmark & \xmark & \xmark & \cmark & R+S & So & 2K & $\leq 60$ \\
AVHBench~\cite{sung2024avhbench} & \cmark & \cmark & \xmark & \xmark & \cmark & R+S & So & 5.8K & 5--30 \\
Daily-Omni~\cite{zhou2025daily} & \cmark & \cmark & \xmark & \xmark & \xmark & R & Sp+So+Mu & 1.2K & 30--60 \\
WorldSense~\cite{hong2025worldsense} & \cmark & \cmark & \xmark & \xmark & \cmark & R & Sp+So+Mu & 3.2K & 140 \\
VCR-Bench~\cite{qi2025vcr} & \cmark & \xmark & \xmark & \cmark & \cmark & R & / & 1,034 & 1--3270 \\
OmniVideoBench~\cite{li2510omnivideobench} & \cmark & \cmark & \xmark & \cmark & \cmark & R & Sp+So+Mu & 1k & 4--1955 \\
MAVERIX~\cite{xie2025maverix} & \cmark & \cmark & \xmark & \xmark & \cmark & R & Sp+So+Mu & 2.5K & 5--6620 \\
TreeBench~\cite{wang2025traceable} & \xmark & \xmark & \cmark & \cmark & \cmark & R & / & 405 & / \\
\midrule
\rowcolor{gray!15}
\textbf{\color{cmured} ANCHOR} \textbf{(Ours)} & \cmark & \cmark & \cmark & \cmark & \cmark & R & Sp+So+Mu & 8.2K & 5--6856 \\
\bottomrule
\end{tabular}
\end{table*}

\subsection{Multimodal Understanding Benchmarks.}
Multimodal benchmarks have progressed from static captioning, grounding, and visual question answering~\cite{chen2015microsoftcococaptionsdata,Agrawal_2019,zhu2016visual7wgroundedquestionanswering,plummer2016flickr30kentitiescollectingregiontophrase} to temporal and long-context video understanding~\cite{patraucen,huang2020movienetholisticdatasetmovie,song2024moviechat}. Audio-visual benchmarks further evaluate cross-modal inference over sound, speech, and vision: AVQA targets audio-visual question answering~\cite{yang2022avqa}; DAVE and AVHBench cover diverse audio-visual events~\cite{radevski2025dave,sung2024avhbench}; Daily-Omni and WorldSense emphasize real-world multisensory understanding~\cite{zhou2025daily,hong2025worldsense}; and MMAU and MAVERIX provide broader audio and audio-visual reasoning suites~\cite{sakshi2024mmau,xie2025maverix}. These benchmarks nevertheless focus primarily on final-answer correctness.

Closest to our setting, OmniVideoBench includes reasoning-oriented supervision for omni-modal video understanding~\cite{li2510omnivideobench}. VCR-Bench evaluates video-only CoT using manually annotated stepwise rationales and separates perception from reasoning errors~\cite{qi2025vcr}. TreeBench evaluates box-grounded visual reasoning over static images~\cite{wang2025traceable}. In contrast, \textbf{\color{cmured}ANCHOR} evaluates generated rationales in audio-visual video QA using aligned temporal, spatial, visual, and audio evidence.

Most existing benchmarks therefore measure \emph{what} a model predicts without determining whether its stated rationale identifies the evidence supporting that prediction. \textbf{\color{cmured}ANCHOR} addresses this gap through evidence-linked evaluation, measuring whether generated explanations align with the annotated perceptual evidence rather than relying on final-answer accuracy alone (Table~\ref{tab:full_benchmark_comparison}).
\section{Benchmark Design}
\label{sec:design}
{\color{cmured} \textbf{ANCHOR}} evaluates audio-visually grounded reasoning in SoTA MLLMs. This section describes (i) dataset construction, (ii) statistics, (iii) difficulty rating, and (iv) human-in-the-loop annotation.

\subsection{Dataset Construction Pipeline}
We collect videos whose questions require visual or audio evidence, spanning durations from 4.9 seconds to 114.27 minutes and resolutions from 120p to 1920p. The videos cover diverse categories, including Shopping, Sentiment, Egocentric Agent, Information Query, Social Interaction, Sports, and Gaming, with multiple-choice and open-ended questions testing reasoning, audio-visual correlation, and temporal and spatial understanding.

All videos and QA pairs in {\color{cmured}\textbf{ANCHOR}} are sourced from existing benchmarks rather than newly authored from scratch. We therefore treat the original benchmark answers as reference answers, and focus our contribution on selecting audio-visually grounded examples, assigning difficulty, and augmenting them with aligned reasoning traces, temporal evidence, audio cues, and spatial bounding boxes.

Videos and QA pairs are sourced from nine datasets: AVQA~\cite{yang2022avqa}, Ego4D~\cite{grauman2022ego4d}, THUMOS-14~\cite{THUMOS14}, Animal-Bench~\cite{jing2024animal}, MAVERIX~\cite{xie2025maverix}, COIN~\cite{tang2019coin}, Moviechat~\cite{song2024moviechat}, Video-MME~\cite{fu2025video}, and LVBench~\cite{wang2025lvbench}. We select AVQA and Ego4D pairs requiring non-redundant audio-visual grounding and first-person interaction logic, THUMOS-14 and Animal-Bench examples emphasizing fine-grained temporal boundaries and complex biological motion, and MAVERIX and COIN examples targeting cross-modal reasoning and instructional temporal structure. For LVBench, Moviechat, and Video-MME, we choose videos longer than 500 seconds with 361p to 720p resolution to better cover long-horizon information retrieval and produce less skewed duration and resolution distributions.

\begin{figure}[!t]
    \centering
    \includegraphics[width=\linewidth]{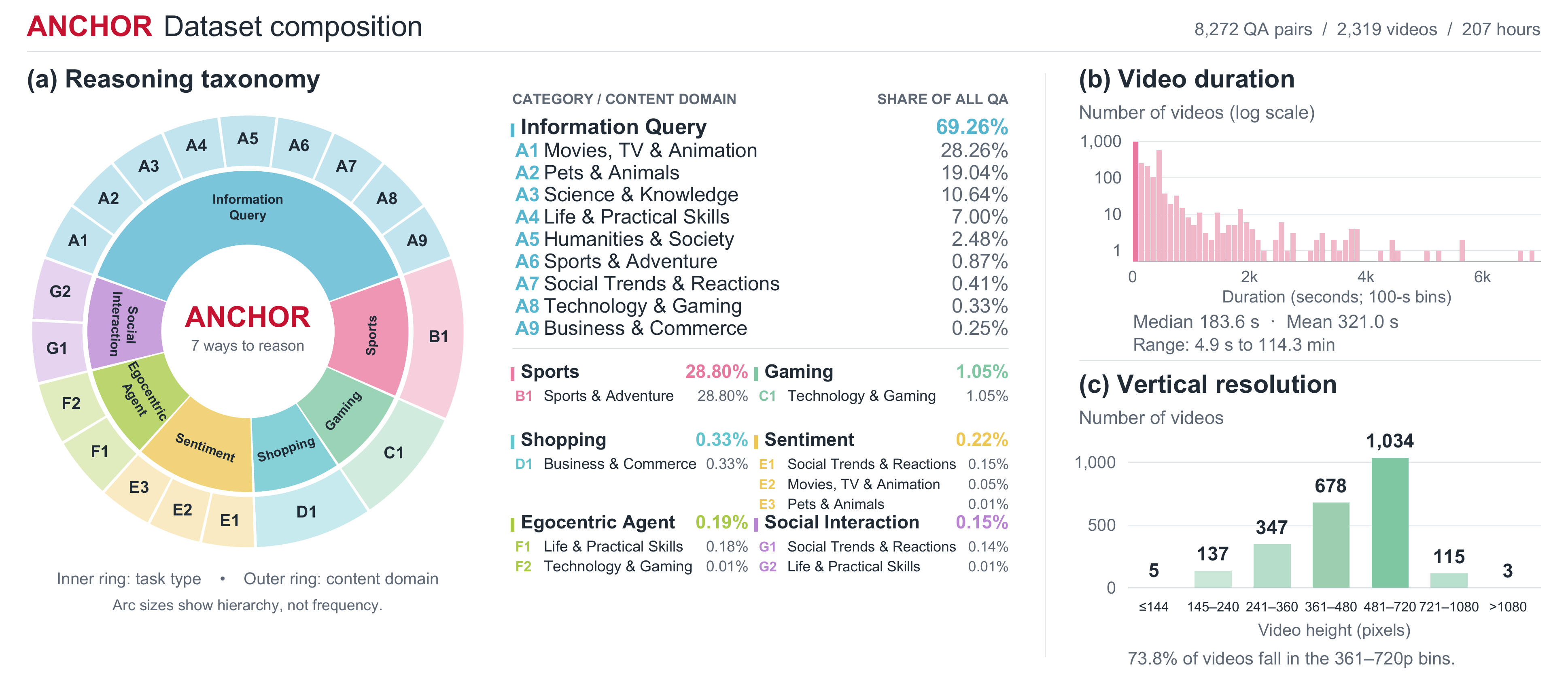}
\caption{\textbf{{\color{cmured}ANCHOR}.} (Left) Taxonomy distribution across top-level categories and subcategories. (Right) Data distributions showing a long-tailed duration profile (Top) and the prevalence of standard vertical resolutions like 480p/720p (Bottom).}
\label{fig:stats}

\end{figure}

\subsection{Dataset Statistics}
The {\color{cmured} \textbf{ANCHOR}} benchmark contains 2,319 videos and 8,272 question--answer pairs, including a human-refined subset of 224 videos and 975 QA pairs. Videos span 745,443 seconds ($\sim$207 hours) in total, averaging 321.45s (median 183.53s; range 4.9s--114.3min), with a long-tailed distribution (Fig.~\ref{fig:stats}): 37.86\% under one minute, 54.12\% one to ten minutes, and 8.02\% over ten minutes.

The dataset includes 21,003,063 processed frames at an average of 28.24 FPS. Frame resolution ranges from 19,200 to 3,686,400 pixels, with mean and median values of 630,347 and 409,920 pixels; vertical resolution ranges from 120p to 1920p, with mean and median heights of 566p and 480p. This range captures both low-resolution web videos and higher-resolution footage, making the benchmark suitable for evaluating robustness across varied visual quality. Figure~\ref{fig:stats} summarizes the category, duration, and resolution distributions, highlighting both the diversity of domains and the heterogeneity of video lengths and resolutions.

\subsection{Difficulty Rating}
\label{sec:diff_rating}
Each video contains multiple QA pairs, including local questions focused on short temporal windows and global information-retrieval questions. We assign each QA pair a difficulty level, easy, medium, or hard, using rollout disagreement and intra-rollout similarity. This procedure is designed to identify questions that are difficult because they require robust multimodal reasoning, while filtering out cases where apparent difficulty arises from unstable model outputs or ambiguous annotations.

Four rollout models (\textsc{Gemini-2.5-Pro}, \textsc{InternVL-3.5-30B-A3B}, \textsc{GPT-4o-mini}, and \textsc{Claude Sonnet-4.5}) each answer every question four times, yielding 16 candidate answers per QA pair. Four judge models assess semantic mismatch against the reference answer and intra-rollout similarity on a 1--5 Likert scale. We label a question hard if at least 10 rollout answers disagree or the similarity score is 5, medium if 5--9 answers disagree or the similarity score is 3--4, and easy otherwise; see Appendix~\ref{app:difficulty}. For models without native audio-video support, we parse videos into frames and transcribe audio using \textsc{GPT-4o-mini-transcribe} for closed-source models and \textsc{Whisper} base for open-source models.

After LLM-based filtering, four human experts review QA pairs labeled hard to distinguish genuinely difficult questions from collective model or judge errors. Around 80\% of LLM-labeled hard QA pairs are confirmed as hard by human judgment, and a subset is provided for annotation correction. Among AI-labeled QA pairs, 27.5\% are easy, 18.7\% are medium, and 53.8\% are hard, making {\color{cmured}\textbf{ANCHOR}} predominantly hard. This distribution intentionally stresses challenging cases where models must integrate visual, audio, temporal, and spatial evidence rather than relying on simple recognition or language priors.

\begin{figure}[!t]
    \centering
    \includegraphics[width=\linewidth]{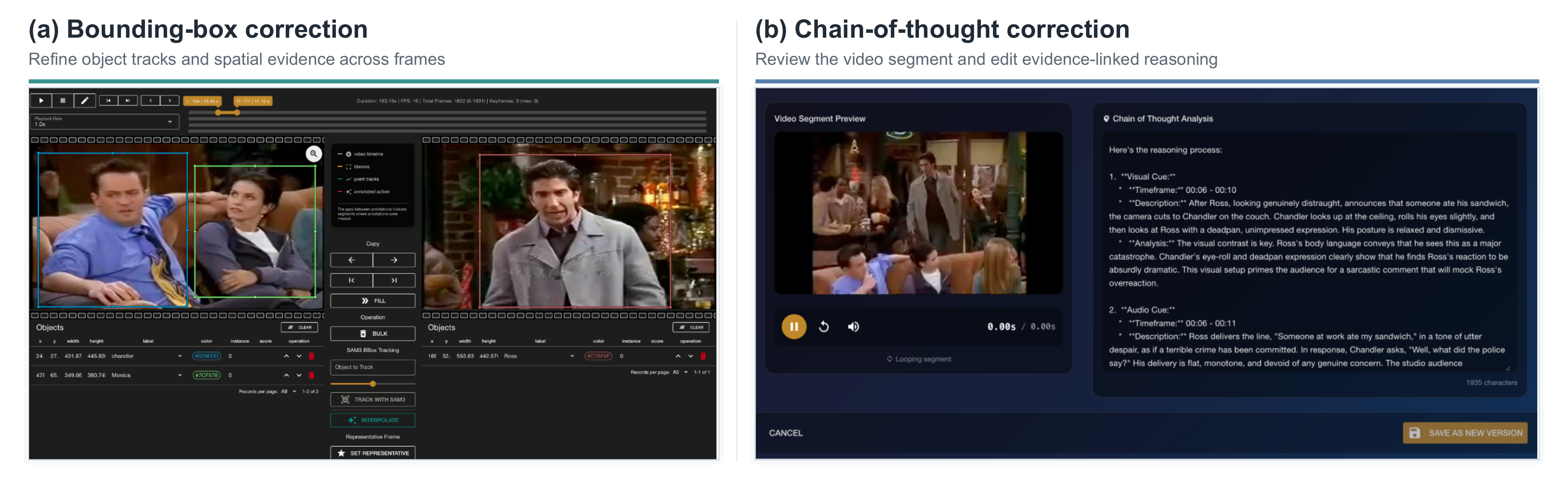}
\caption{\textbf{The {\color{cmured} ANCHOR} Annotation Interface.} Our system synchronizes: \textbf{(a) Spatiotemporal View (Left)}, a frame-level inspector for refining SAM3 bounding box proposals; and \textbf{(b) CoT Editor (Right)}, a structured suite where reasoning steps are explicitly linked to visual and acoustic evidence.}
\label{fig:annotation_tool}
\vspace{-0.3cm}
\end{figure}

\subsection{Human-in-the-Loop Video Annotation}
\label{sec:video_annotation}

Given a video clip with QA annotations, preliminary SAM3 bounding boxes~\cite{carion2025sam3segmentconcepts}, and pre-processed CoT traces, annotators refine both perceptual grounding signals and the correctness, coherence, and faithfulness of the reasoning process. For examples categorized as ``hard,'' this refinement emphasizes factual accuracy, rigorous multimodal grounding, and human interpretability.

We first generate pseudo ground-truth reasoning traces by prompting \textsc{Gemini-2.5-Pro}~\cite{comanici2025gemini} with each video's question and ground-truth answer. The model produces a structured CoT describing the visual and audio cues, including precise timeframes and object descriptions, needed to reach the answer. These traces are paired with automatically generated SAM3 bounding boxes to identify the spatial regions referenced in the explanation; see Appendix~\ref{app:cot_schema} and Appendix~\ref{app:bb_target_selection} for the CoT schema and bounding-box target-selection details.

To make refinement scalable across videos with varying duration, scene complexity, and perceptual event density, we extend the VidAT Video Annotation Tool~\cite{zhang2020vidat} with a unified interface for synchronized video, audio, bounding-box, timeline, and CoT editing, shown in Figure~\ref{fig:annotation_tool}. For bounding-box refinement, annotators can manually or SAM3-assist creation and correction, adjust spatial boundaries, propagate boxes across frames via temporal interpolation, and remove redundant or spurious detections. For CoT refinement, annotators edit reasoning traces while observing the corresponding perceptual evidence and explicitly link reasoning steps to visual regions or audio events.

We conducted an IRB-approved controlled annotation study with $26$ participants, who collectively refined $224$ videos and $975$ QA pairs. Participants completed approximately $10$ videos per hour on average, with each video containing multiple QA pairs. When they identified logical discrepancies or misaligned spatiotemporal grounding, they corrected bounding boxes, adjusted audio-visual event spans, and rewrote reasoning steps to ensure strict alignment with the raw perceptual evidence. This human-in-the-loop process yields a reliable gold standard for evaluating grounded reasoning.

\label{sec:focus_subset} Beyond the full human-refined subset, we construct a focused validation subset by uniformly randomly sampling $100$ videos ($446$ QA pairs) from this set. We use this fixed subset for two targeted analyses requiring additional reference annotations or model outputs beyond those collected for the full human-refined set: a reference-source circularity check (\ref{app:circularity}) and a direct spatial/temporal grounding evaluation (\ref{app:direct_grounding}).
\section{Experiments}

\begin{table}[!htbp]
\centering
{%
\scriptsize
\setlength{\tabcolsep}{5pt}
\renewcommand{\arraystretch}{1.15}
\begin{tabular}{lccccc}
\toprule
\textbf{Model} &
\makecell{\textbf{Answer}\\$\uparrow$} &
\makecell{\textbf{Halluc.}\\$\uparrow$} &
\makecell{\textbf{Conc.}\\$\uparrow$} &
\makecell{\textbf{Tokens}\\$\downarrow$} &
\makecell{\textbf{Div.}\\$\uparrow$} \\
\midrule
Qwen3-VL-30B-A3B-Instruct &
\answercell{3.45} & \halluccell{1.79} & \conccell{3.40} &
\tokencell{287} & \divcell{2.50} \\

Qwen3-omni &
\answercell{3.34} & \halluccell{2.03} & \conccell{3.90} &
\tokencell{291} & \divcell{2.65} \\

InternVL3.5 &
\answercell{2.14} & \halluccell{1.53} & \conccell{3.48} &
\tokencell{283} & \divcell{1.85} \\

Llava-OneVision &
\answercell{1.85} & \halluccell{1.53} & \conccell{4.25} &
\tokenbest{202} & \divcell{1.58} \\

GLM-4.6-V-Flash &
\answercell{3.29} & \halluccell{1.94} & \conccell{2.50} &
\tokencell{309} & \divcell{2.46} \\

Kimi-VL-A3B-Instruct &
\answercell{3.05} & \halluccell{1.93} & \conccell{2.83} &
\tokencell{336} & \divcell{2.31} \\

Nemotron-3-Nano-Omni-30B &
\answercell{3.37} & \hallucbest{2.21} & \conccell{3.69} &
\tokencell{206} & \divcell{2.59} \\
\midrule
Gemini-3-Flash &
\answerbest{3.83} & \halluccell{2.20} & \concbest{4.47} &
\tokencell{203} & \divbest{2.93} \\

GPT-5.2 &
\answercell{3.77} & \halluccell{2.05} & \conccell{2.72} &
\tokencell{342} & \divcell{2.83} \\

Grok-4.1-Fast-Reasoning &
\answercell{3.39} & \halluccell{1.48} & \conccell{1.30} &
\tokencell{644} & \divcell{2.39} \\

Claude-Sonnet-4.6 &
\answercell{3.37} & \halluccell{1.66} & \conccell{1.50} &
\tokencell{571} & \divcell{2.55} \\

Amazon Nova Lite &
\answercell{3.30} & \halluccell{1.63} & \conccell{2.01} &
\tokencell{545} & \divcell{2.42} \\
\bottomrule
\end{tabular}
}
\vspace{6pt}
\caption{\textbf{CoT evaluation.}
MLLMs evaluated under truncated-video input settings.
Answer correctness (Answer), faithfulness (Halluc.), and
reasoning consistency (Div.) are scored on a 1--5 Likert scale
by three judges (\textsc{GPT-5-nano}, \textsc{Grok-4.1-fast},
\textsc{Claude-Haiku-4.5}), with higher scores indicating
better performance.
Conciseness (Conc.) is a token-based quintile score
(1--5, higher = more concise) derived from average output
token count (Tokens) relative to the global distribution
across all models.
Each metric has a distinct hue; stronger shading indicates
higher scores or fewer tokens.
Shading uses a shared 1--5 score scale and a shared
149--736 token range across both evaluation tables.
Bold marks the best result in each column.}
\label{tab:multi_judge_cot}
\end{table}

\begin{table}[!htbp]
\centering
{%
\scriptsize
\setlength{\tabcolsep}{5pt}
\renewcommand{\arraystretch}{1.15}
\begin{tabular}{lccccc}
\toprule
\textbf{Model} &
\makecell{\textbf{Answer}\\$\uparrow$} &
\makecell{\textbf{Halluc.}\\$\uparrow$} &
\makecell{\textbf{Conc.}\\$\uparrow$} &
\makecell{\textbf{Tokens}\\$\downarrow$} &
\makecell{\textbf{Div.}\\$\uparrow$} \\
\midrule
Gemini-3.1-Pro &
\answercell{3.49} & \hallucbest{2.49} & \concbest{4.80} &
\tokenbest{149} & \divbest{2.97} \\

Gemini-3-Flash &
\answerbest{3.59} & \halluccell{2.08} & \conccell{4.09} &
\tokencell{200} & \divcell{2.86} \\
\midrule
Grok-4 &
\answercell{3.01} & \halluccell{1.61} & \conccell{1.40} &
\tokencell{736} & \divcell{2.50} \\

Grok-4.1-Fast-Reasoning &
\answercell{2.85} & \halluccell{1.35} & \conccell{1.59} &
\tokencell{653} & \divcell{2.28} \\
\midrule
Nova-2-Pro &
\answercell{2.93} & \halluccell{1.96} & \conccell{3.62} &
\tokencell{243} & \divcell{2.48} \\

Nova-2-Lite &
\answercell{3.03} & \halluccell{1.63} & \conccell{2.50} &
\tokencell{435} & \divcell{2.43} \\
\bottomrule
\end{tabular}
}
\vspace{4pt}
\caption{\textbf{CoT evaluation on the Human-Refined set.}
Closed-source MLLMs evaluated on the Human-Refined subset
(224 videos, 975 QA pairs).
Answer correctness (Answer), faithfulness (Halluc.), and
reasoning consistency (Div.) are scored on a 1--5 Likert scale
by three judges, with higher scores indicating better performance.
Conciseness (Conc.) is a token-based quintile score
(1--5, higher = more concise) derived from average output
token count (Tokens) relative to the global token distribution.
Colors and normalization match Table~\ref{tab:multi_judge_cot};
stronger shading indicates higher scores or fewer tokens.
Bold marks the best result in each column.}
\label{tab:multi_judge_cot_closed_human}
\end{table}

\subsection{Experimental Setup}
\label{sec:exp_setup}
We use the same input formatting, prompts, and scoring rules for both experimental components: (i) CoT evaluation and (ii) answer-correctness ablations under different grounding cues; see Appendix~\ref{app:evaluation_curation} for additional protocol details.

\paragraph{\textbf{Model inputs and grounding cues.}}
Each example contains video with synchronized audio. Native audio-video models receive the audio-visual stream directly; other models receive sampled frames paired with a Whisper ASR transcript. In ablations, we vary bounding boxes (BB), visual evidence, and audio evidence while keeping all other inputs and prompts fixed.

All frames are resized to $512 \times 384$, and clips are cropped when QA start/end timestamps are available. For clips $\leq 20$s, we feed the full clip: \textsc{Gemini} and \textsc{Qwen3-VL} receive video directly, with \textsc{Qwen3-VL} also receiving the Whisper transcript; \textsc{ChatGPT}, \textsc{Claude}, and \textsc{Grok} receive frames sampled at $5$ FPS plus the transcript, or ``no speech detected'' if transcription is empty.

For clips $>20$s, we use a Socratic sequential-chunk procedure~\cite{zeng2022socraticmodelscomposingzeroshot}. A clipped segment $V$ of duration $T$ is split into $n$ chunks $\{V^{(1)},\dots,V^{(n)}\}$ with model-dependent chunk length. At chunk $k$, the model receives the question, current chunk evidence, timestamped frame/ASR history when applicable, and prior intermediate reasoning $r^{(1)},\dots,r^{(k-1)}$. A final text-only synthesis pass produces:
\begin{equation}
A = M\big([\tau, r^{(1)}, r^{(2)}, \ldots, r^{(n)}, Q]\big),
\end{equation}
where $r^{(k)}$ is the intermediate output for chunk $k$ and $\tau$ is a fixed prompt. Frame-based models use $120$s chunks with frames sampled at $0.5$ FPS and resized to $512 \times 384$, plus Whisper segment-level transcripts. Native-video models use $1800$s (30 min) chunks: \textsc{Gemini} receives audio-preserved chunks and performs internal temporal sampling, while \textsc{Qwen3-VL} uses the same chunking plus the Whisper transcript.

\paragraph{\textbf{Scoring and judge protocol.}}
We evaluate final-answer correctness against the reference answer; \textsc{GPT-5-nano}, \textsc{Grok-4.1-fast}, and \textsc{Claude-Haiku-4.5} score hallucination/faithfulness, divergence, and final answer, while conciseness is computed from output token counts. Given the question, model output, and reference evidence annotations, judges assign 1--5 Likert scores for faithfulness/hallucination, step-to-step divergence, and final answer, measuring evidence support, logical drift/inconsistency, and answer correctness, respectively. Conciseness is measured independently via output token count using the \texttt{cl100k\_base} tokenizer, where responses are binned into quintile-based scores from 1 (most verbose) to 5 (most concise). We average judge scores across the three judge-evaluated dimensions and report inter-judge agreement with the main results; see Appendix~\ref{app:judge_rubrics} for judge-rubric definitions.

On the focused validation subset, model scores correlate strongly (Pearson $r=0.96$--$1.00$) across human-refined, \textsc{Gemini}-, and \textsc{GPT}-generated references, and \textsc{Gemini} does not score highest under its own generated reference on any facet, evidence against reference-source circularity (Appendix~\ref{app:circularity}).

\begin{figure}[!t]
    \centering
    \includegraphics[width=\linewidth]{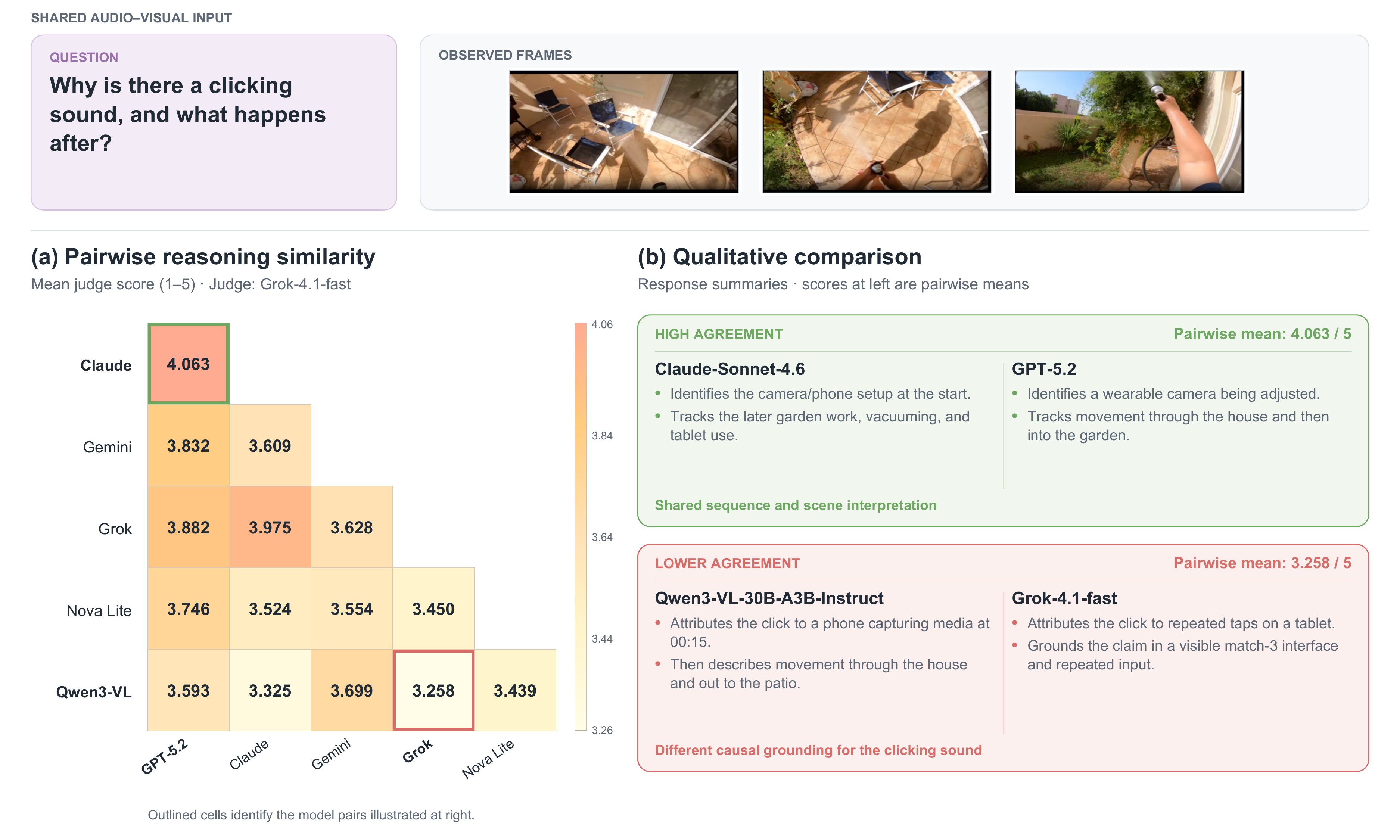}
\caption{\textbf{CoT Faithfulness Analysis.} (Left) Pairwise similarity matrix of model-generated reasoning traces. (Right) Qualitative example where \textsc{Claude} and \textsc{GPT-5.2} show high reasoning convergence ($4.06$), while \textsc{Grok-4.1} and \textsc{Qwen-3VL} provide distinct, granular grounding for the clicking sound and subsequent actions.}
\label{fig:mllm_sim}
\vspace{-0.5cm}
\end{figure}

\subsection{CoT Evaluation}
\label{sec:multi_judge_cot}
Correct final answers do not necessarily imply grounded or reliable reasoning. To directly evaluate reasoning quality, we prompt representative MLLMs to produce both a CoT and a final answer for each question, then score the outputs using the judge protocol in Sec.~\ref{sec:exp_setup}.

As shown in Table~\ref{tab:multi_judge_cot},
\textsc{Gemini-3-Flash} and \textsc{GPT-5.2} achieve similar Answer
scores ($3.83$ and $3.77$) but differ in Hallucination and Divergence.
To control for answer correctness, we examine 2,012 items for which all
judges rate both models' answers as fully correct. GPT-5.2's Hallucination
score is at least one point lower on 250 items and at least 1.5 points
lower on 81; for the latter, all judges agree on the direction in 74\%
of cases. Inspection reveals three recurring failures: identifying the
wrong event, localizing the wrong entity or interval, and retrieving
relevant evidence without connecting it correctly to the answer.

Input representation may contribute to these gaps. Models receiving frames
and ASR rather than native audio claim that no usable audio is available
on 25.3\% of GPT-5.2, 25.0\% of Claude, and 20.9\% of Grok responses,
compared with 3.7\% for Nova and 0.1\% for Gemini-3-Flash. In 85\% of
GPT-5.2's cases, the reference rationale depends on informative
non-speech audio. Since input format and architecture are confounded,
we treat this association as explanatory rather than causal.
Table~\ref{tab:multi_judge_cot_closed_human} reports additional
closed-source models on the human-refined set.

\subsection{Ablation on Grounding Cues}
\label{sec:ablation_grounding}

We quantify the contribution of perceptual grounding signals to answer correctness on the human-verified subset of 224 videos and 975 QA pairs refined during annotation (Sec.~\ref{sec:video_annotation}). To isolate cue effects, we evaluate \textsc{Gemini-3-Flash}, \textsc{GPT-5.6-Luna}, and \textsc{Claude-Sonnet-5} under controlled cue removals. We group cues into \textbf{BB}, spatial anchors for referenced objects injected as structured frame-level annotations; \textbf{Visual}, appearance and motion information from per-QA video segments; and \textbf{Audio}, acoustic evidence from the raw audio track and ASR transcripts. We mask specific modalities by stripping the acoustic stream for \emph{no-audio} settings and removing visual frames for \emph{audio-only} settings. Answer accuracy is judged by \textsc{GPT-5-Nano} as a zero-shot binary evaluator against the benchmark reference, and Table~\ref{tab:ablation_grounding_cues} reports absolute accuracy and the drop $\Delta$ relative to full modality.

\begin{table}[!htbp]
\centering
\caption{%
\textbf{Ablation of Grounding Cues.}
Accuracy on the human-refined test set ($n{=}975$ QA pairs), with
the change from each model's full-modality accuracy shown in parentheses.
Results are judged by \textsc{GPT-5-Nano}.%
}
\label{tab:ablation_grounding_cues}

\scriptsize
\setlength{\tabcolsep}{4pt}
\renewcommand{\arraystretch}{1.12}

\begin{tabular}{lccc ccc}
\toprule
\textbf{Setting} &
\textbf{Visual} &
\textbf{BB} &
\textbf{Audio} &
\makecell{\textbf{Gemini}\\\textbf{3-Flash}} &
\makecell{\textbf{GPT}\\\textbf{5.6-Luna}} &
\makecell{\textbf{Claude}\\\textbf{Sonnet-5}} \\
\midrule

\rowcolor{gray!15}
Full Modality
    & \cmark & \cmark & \cmark
    & 57.7 ({--})
    & 52.5 ({--})
    & 52.6 ({--}) \\

w/o Audio
    & \cmark & \cmark & \xmark
    & 52.9 (-4.8)
    & 48.3 (-4.2)
    & 47.0 (-5.6) \\

w/o Bounding Boxes
    & \cmark & \xmark & \cmark
    & 50.0 (-7.7)
    & 48.9 (-3.6)
    & 49.4 (-3.2) \\

w/o Audio \& Bounding Boxes
    & \cmark & \xmark & \xmark
    & 48.1 (-9.6)
    & 45.0 (-7.5)
    & 45.1 (-7.5) \\

Audio Only
    & \xmark & \xmark & \cmark
    & 16.1 (-41.6)
    & 10.7 (-41.8)
    & 14.2 (-38.4) \\

\bottomrule
\end{tabular}
\end{table}

\paragraph{\textbf{Results.}}
Table~\ref{tab:ablation_grounding_cues} shows that visual evidence is
dominant across all three models: removing video reduces accuracy by
38.4--41.8 percentage points. Removing audio causes a 4.2--5.6-point drop,
while removing bounding boxes causes a 3.2--7.7-point drop; removing both
reduces accuracy by 7.5--9.6 points. The relative importance of audio and
spatial cues varies by model, indicating complementary, model-dependent
benefits. Audio-only results are not directly comparable across models:
\textsc{Gemini-3-Flash} receives native audio, whereas
\textsc{GPT-5.6-Luna} and \textsc{Claude-Sonnet-5} receive ASR transcripts.

\subsection{Ablation on Judge Models}
\label{sec:ablation_judges}

\begin{figure}
    \centering
    \includegraphics[width=0.9\linewidth]{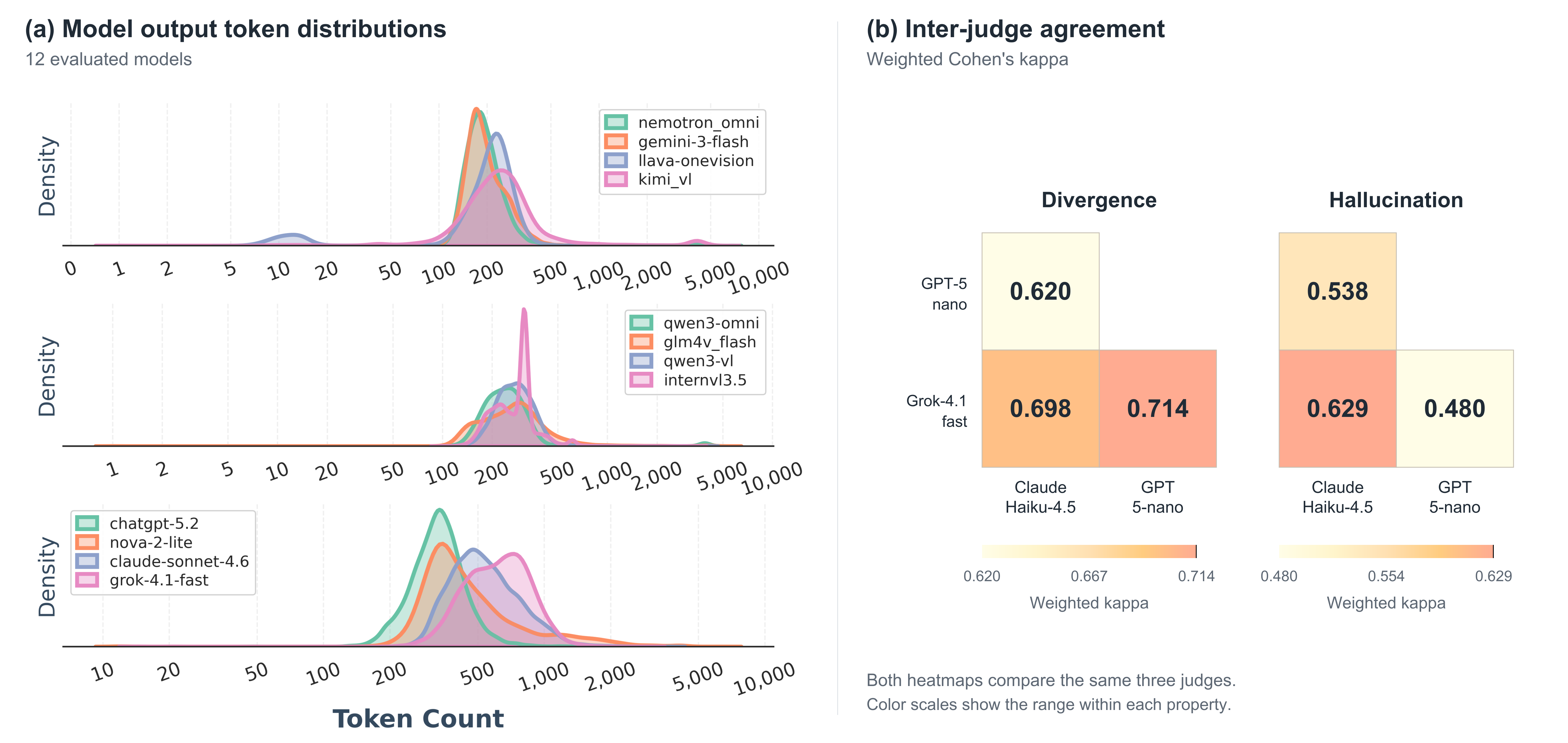}
    \caption{\textbf{Model Output Lengths and Inter-Judge Agreement.}
(a) CoT output length distributions across 12 evaluated MLLMs.
Shorter outputs receive higher conciseness scores by construction.
(b) Pairwise weighted Cohen's $\kappa$ among the three judge models
for Divergence and Hallucination. Each heatmap uses a separate color scale.}
\label{fig:tokens_and_judge_agreement}
\end{figure}

We quantify inter-judge agreement for the judge-evaluated CoT-quality scores defined in Sec.~\ref{sec:exp_setup}, namely Hallucination and Divergence, using pairwise weighted Cohen's $\kappa$ with linear weights over the 1--5 Likert scale. As shown in Fig.~\ref{fig:tokens_and_judge_agreement}, agreement varies by metric. Divergence agreement is highest between \textsc{Grok-4.1-Fast} and \textsc{GPT-5-Nano} ($\kappa{=}0.714$), followed by \textsc{Grok-4.1-Fast} and \textsc{Claude-Haiku-4.5} ($\kappa{=}0.698$). Hallucination agreement is also strong between \textsc{Grok-4.1-Fast} and \textsc{Claude-Haiku-4.5} ($\kappa{=}0.629$). Overall, moderate-to-high agreement for Hallucination and Divergence supports the reliability of our multi-judge protocol for auditing reasoning-trace faithfulness. However, inter-judge agreement measures consistency among LLM judges rather than agreement with human raters: on $100$ sampled model responses, judge-panel and human ratings correlate only moderately (Pearson $r{=}0.50$), with human ratings systematically higher, e.g., hallucination ($4.35$ vs.\ $2.35$; Appendix~\ref{app:human_judge_validation}).







\section{Conclusion}
We introduced \textbf{{\color{cmured} ANCHOR}}, a benchmark and evaluation framework for testing whether MLLMs ground their reasoning in relevant audio-visual evidence rather than linguistic priors or dataset shortcuts. Unlike prior audio-visual QA benchmarks that primarily score final answers, \textbf{\color{cmured} ANCHOR} requires grounded CoT explanations with explicit perceptual anchors, including bounding boxes, representative frames, and salient audio events.

Our construction pipeline combines diverse short- and long-form videos, multiple-choice and open-ended QA pairs, automated SAM3-based grounding proposals, model-generated reasoning traces, and human-in-the-loop refinement through a dedicated annotation toolbox. Experiments reveal a persistent faithfulness gap: strong answer accuracy does not guarantee grounded reasoning. Ablations further show that spatial and visual evidence are central, with performance dropping substantially when bounding boxes or video evidence are removed.

These findings suggest that progress in multimodal reasoning requires moving beyond answer-only evaluation toward benchmarks that directly stress-test perceptual evidence use. While \textbf{\color{cmured} ANCHOR} expands grounded audio-visual evaluation, it remains limited by the scale of human-refined annotations, reliance on pseudo ground-truth traces, and coverage of the source datasets. We hope \textbf{\color{cmured} ANCHOR} supports the development of models that are accurate, interpretable, and trustworthy in real-world audio-visual settings.

\section{Acknowledgements}
This work was financially supported by Amazon. We thank the annotators who made this work possible: Aditya Gupta, Poh-Hsuan Huang, Vaishnavi Khindkar, Sambhav Rohatgi, and Chun-Hsien Lin.

\newpage
\bibliographystyle{abbrv}
\bibliography{main}
\newpage
\appendix
\section{Appendix}
\label{app:evaluation_curation}

This appendix provides additional methodological details for the \textbf{{\color{cmured} ANCHOR}} curation and evaluation pipeline. We include the difficulty assignment rule, the grounded chain-of-thought annotation schema, compact judge-rubric definitions, and the bounding-box target selection procedure. Full prompt text and extended quantitative results are provided in the supplementary material.

\subsection{Difficulty Assignment}
\label{app:difficulty}

Each question--answer pair is assigned a difficulty label using two complementary signals: rollout disagreement and intra-rollout answer diversity. As described in Sec. \ref{sec:diff_rating}, four rollout models each answer a question four times, producing 16 candidate answers. Judge models then evaluate whether each candidate semantically matches the reference answer and assign an overall similarity score to the set of candidate answers on a 1--5 Likert scale. The final difficulty is the maximum severity implied by these two signals.

\begin{algorithm}[h]
\caption{Difficulty Tier Assignment}
\label{alg:difficulty_assignment}
\begin{algorithmic}[1]
\Require $n_{\mathrm{mismatch}}$, $s_{\mathrm{sim}}$
\Ensure difficulty label $d$
\Statex
\Comment{Difficulty tier from rollout disagreement}
\If{$n_{\mathrm{mismatch}} \geq 10$}
    \State $d_{\mathrm{disagree}} \gets \textsc{Difficult}$
\ElsIf{$n_{\mathrm{mismatch}} \geq 5$}
    \State $d_{\mathrm{disagree}} \gets \textsc{Medium}$
\Else
    \State $d_{\mathrm{disagree}} \gets \textsc{Easy}$
\EndIf
\Statex
\Comment{Difficulty tier from intra-rollout similarity}
\If{$s_{\mathrm{sim}} = 5$}
    \State $d_{\mathrm{score}} \gets \textsc{Difficult}$
\ElsIf{$s_{\mathrm{sim}} \in \{3,4\}$}
    \State $d_{\mathrm{score}} \gets \textsc{Medium}$
\Else
    \State $d_{\mathrm{score}} \gets \textsc{Easy}$
\EndIf
\Statex
\Comment{Final label is the maximum-severity tier}
\State Define severity order $\textsc{Easy} < \textsc{Medium} < \textsc{Difficult}$
\State $d \gets \max(d_{\mathrm{disagree}}, d_{\mathrm{score}})$
\State \Return $d$
\end{algorithmic}
\end{algorithm}

This rule is intentionally conservative: a question is considered difficult if either many independent rollouts fail to match the reference answer, or the set of model answers is highly diverse. Conversely, questions are labeled easy only when candidate answers are mostly consistent and semantically aligned with the reference. This procedure helps distinguish examples that require robust multimodal reasoning from examples that can be answered reliably through simple recognition or linguistic priors.

\subsection{Grounded Chain-of-Thought Annotation Schema}
\label{app:cot_schema}

For each question--answer pair, \textbf{{\color{cmured} ANCHOR}} associates the reference answer with a grounded reasoning trace. The initial trace is generated automatically and then refined during human annotation. The trace is structured to explicitly separate visual evidence, audio evidence, and the final answer:

\begin{quote}
\small
\textbf{Visual Cue:} timeframe, description, analysis. \\
\textbf{Audio Cue:} timeframe, description, analysis. \\
\textbf{Final Answer:} concise answer to the question.
\end{quote}

The visual cue records the relevant temporal span, visible entities, actions, scene changes, and spatial evidence needed to answer the question. The audio cue records speech, sound events, music, or other acoustic information that supports the reasoning process. The final answer records the reference response used for answer-correctness evaluation.

Human annotators refine these traces while viewing the video, audio timeline, bounding-box proposals, and editable reasoning fields in the annotation interface. When the automatically generated trace contains unsupported entities, incorrect temporal spans, missing audio evidence, or weak causal links, annotators correct the reasoning text and its grounding links. This process yields a human-refined subset with aligned answer, reasoning, temporal, audio, and spatial annotations.

\subsection{Judge Rubrics}
\label{app:judge_rubrics}

\textbf{{\color{cmured} ANCHOR}} evaluates model outputs along four dimensions: final-answer correctness, hallucination, divergence, and conciseness. The first three dimensions are judged on a 1--5 Likert scale by multiple LLM judges, while conciseness is computed from token counts. Full judge prompts are provided in the supplementary material.

\paragraph{Final-answer correctness.}
This score measures whether the model's final answer is semantically consistent with the reference answer. Judges are instructed to focus on meaning rather than exact wording. Paraphrases, equivalent expressions, and harmless extra details are not penalized if they preserve the core answer. A score of 5 indicates a perfect or near-perfect semantic match, while a score of 1 indicates an unrelated, contradictory, or completely incorrect answer.

\paragraph{Hallucination / faithfulness.}
This score measures whether the model's reasoning introduces unsupported entities, facts, assumptions, or reasoning steps relative to the reference grounded reasoning trace. Judges are instructed not to penalize harmless rephrasing or logically equivalent restatements. A score of 5 indicates that the reasoning is faithful and introduces virtually no unsupported content, while a score of 1 indicates severe hallucination, such as many fabricated objects, contradicted events, or unsupported causal claims.

\paragraph{Divergence.}
This score measures how closely the model's reasoning process aligns with the reference reasoning process in logic, structure, and interpretation of the question. Unlike hallucination, divergence ignores verbosity and unsupported additions unless they change the reasoning trajectory. A score of 5 indicates identical or nearly identical reasoning, while a score of 1 indicates unrelated or incorrect reasoning.

\paragraph{Conciseness.}
Conciseness is measured independently from the judge-evaluated dimensions. We compute the number of output tokens using the same tokenizer across all model outputs and assign a quintile-based score from 1 to 5, where higher scores correspond to shorter and more concise outputs. This avoids conflating verbosity with correctness, hallucination, or reasoning divergence.

\subsection{Bounding-Box Target Selection}
\label{app:bb_target_selection}

To generate spatial grounding proposals, \textbf{{\color{cmured} ANCHOR}} first identifies the visible object or entity most relevant to each question--answer pair and its grounded reasoning trace. The target is constrained to be a concrete noun phrase corresponding to an object, person, animal, vehicle, body part, or other visible entity that can be enclosed by a bounding box. Abstract concepts, actions, emotions, abilities, and dialogue content are not valid targets by themselves. When a question concerns an abstract event or action, the target is instead mapped to the visible entity performing or undergoing that event.

The extracted target label is used to initialize SAM3-based frame-level bounding-box proposals. Human annotators then verify and refine these proposals in the annotation interface. They adjust spatial boundaries, remove spurious detections, propagate boxes across frames when appropriate, and ensure that the final bounding boxes correspond to the perceptual evidence referenced by the grounded reasoning trace.

\subsection{Reference-Source Circularity} \label{app:circularity}
On the focused validation subset (Sec.~\ref{sec:focus_subset}), we score fixed model responses against three reference sources: human-refined, \textsc{Gemini-2.5-Pro}-generated, and \textsc{GPT-5.6-Luna}-generated, using the same judge panel throughout. Model scores correlate strongly across reference-source pairs (Pearson $r=0.96$--$1.00$, mean $0.99$; Spearman $\rho=0.88$--$1.00$, mean $0.95$; each computed over $n{=}9$ models), with a mean absolute score difference of $0.05$ points on the 1--5 scale (max $0.08$). \textsc{Gemini} does not receive its highest score under its own generated reference on any of the three judged facets, instead scoring highest under the human-refined reference.

\subsection{Direct Spatial and Temporal Grounding}
\label{app:direct_grounding}

On the focused validation subset (Sec.~\ref{sec:focus_subset}), we compare boxes generated for objects cited by two models against human-refined tracks. IoU measures spatial overlap, while temporal precision, recall, and F1 measure agreement over visible frames.

\begin{table}[!htbp]
\centering
\caption{\textbf{Direct grounding} on the human-refined subset.}
\label{tab:direct_grounding}
\scriptsize
\setlength{\tabcolsep}{4pt}
\begin{tabular}{lccccc}
\toprule
\textbf{Model} &
\textbf{IoU} $\uparrow$ &
\textbf{IoU$\geq$.5} $\uparrow$ &
\textbf{Temp. P} $\uparrow$ &
\textbf{Temp. R} $\uparrow$ &
\textbf{Temp. F1} $\uparrow$ \\
\midrule
GPT-5.6-Luna
    & \textbf{0.459} & \textbf{48.0\%}
    & \textbf{0.76} & 0.85 & \textbf{0.80} \\
Claude-Haiku-4.5
    & 0.217 & 10.9\%
    & 0.45 & \textbf{0.95} & 0.61 \\
\bottomrule
\end{tabular}
\end{table}

\textsc{GPT-5.6-Luna} achieves stronger spatial and temporal alignment;
\textsc{Claude-Haiku-4.5} attains higher temporal recall but lower
precision. These focused metrics complement semantic evaluation because
exact overlap may penalize alternative valid evidence.

\subsection{Human-Judge Validation}
\label{app:human_judge_validation}

To assess the reliability of our LLM-judge protocol against human judgment, we conducted a human evaluation on 100 model responses sampled from the human-refined subset. Human annotators independently rated answer correctness, hallucination, and divergence from the reference evidence, including cases where the model followed a different but still valid reasoning path. Averaged judge-panel scores correlate moderately with human ratings (Pearson $r=0.50$ pooled across facets; per-facet range $r=0.43$--$0.67$), while exact-category agreement is more modest (weighted Cohen's $\kappa$ ranging from $0.11$ to $0.39$ across facets). Human ratings are systematically higher than judge-panel ratings across all three facets, most notably for hallucination (human mean $4.35$ vs.\ judge-panel mean $2.35$ on the 1--5 scale). We interpret this as evidence of moderate score correlation but limited absolute calibration between human and LLM judges; differences in rubric strictness may contribute to the gap. We treat this discrepancy as a limitation of the multi-judge protocol rather than assuming either evaluator is more reliable.

\subsection{Relationship to the Supplementary Material}
\label{app:supp_relationship}

This appendix includes the compact methodological details most directly needed to interpret the main experimental claims. The supplementary material contains the full implementation-facing details, including complete prompts for taxonomy labeling, pseudo-ground-truth chain-of-thought generation, SAM3 target extraction, rollout generation, difficulty judging, and the LLM judge rubrics. It also contains extended results, including per-model judge-agreement plots, performance by difficulty level, performance by annotation source, and full taxonomy percentages.
\clearpage

\definecolor{prompttext}{RGB}{30, 30, 30}
\definecolor{lightcmu}{RGB}{252, 247, 247}
\definecolor{promptborder}{RGB}{196, 18, 48}
\definecolor{prompttitlebg}{RGB}{245, 225, 225}

\lstdefinestyle{promptstyle}{
    basicstyle=\ttfamily\footnotesize\color{prompttext},
    breaklines=true,
    breakatwhitespace=false,
    columns=fullflexible,
    keepspaces=true,
    showstringspaces=false,
    frame=none,
    xleftmargin=0pt,
    xrightmargin=0pt,
    aboveskip=0pt,
    belowskip=0pt
}

\newtcblisting{judgeprompt}[2][]{%
    enhanced,
    breakable,
    listing only,
    listing engine=listings,
    colback=lightcmu,
    colframe=promptborder,
    coltitle=black,
    title={#2},
    fonttitle=\bfseries,
    boxrule=0.7pt,
    arc=2pt,
    left=8pt,
    right=8pt,
    top=8pt,
    bottom=8pt,
    attach boxed title to top left={xshift=2mm,yshift=-2mm},
    boxed title style={
        colback=prompttitlebg,
        colframe=promptborder,
        boxrule=0.7pt,
        arc=2pt
    },
    listing options={
        style=promptstyle,
        basicstyle=\tiny\ttfamily\color{prompttext}
    },
    #1
}

\raggedbottom
\section{Supplementary Material}

\subsection{Overview}
This supplementary document provides additional details and experiments for
{\color{cmured}\textbf{ANCHOR}}, including:
(i) full judge prompts for evaluation,
(ii) the difficulty-rating procedure used for dataset curation, and
(iii) extended quantitative results and details.

\subsection{Additional Benchmark Details}
We provide the category-wise, subcategory-wise and joint taxonomy distribution for the question-answer pairs in our dataset in Table \ref{tab:taxonomy_percentages_compact}. \textsc{GPT-5.2} is used for this taxonomy labeling with the following prompt:

\begin{judgeprompt}{Taxonomy Labeling}
You are labeling dataset examples into the MAVEN3 taxonomy categories.
Return ONLY the JSON that matches the provided schema.

Top-level categories (choose exactly one):
{TOP_CATEGORIES}

Second-level categories (choose EXACTLY ONE; do NOT output null):
{SUB_CATEGORIES}

Guidance:
- Use the QUESTION/ANSWER as the primary signal.
- Use GEMINI_ANALYSIS only as supporting evidence when it is provided.
- You MUST choose a sub-category even if it is imperfect; pick the closest match.
- "Information Query": factual identification, what/where/who/what sound/what animal/object.
- "Sports": athletics, workouts, sports actions, outdoor adventure.
- "Gaming": video games, game entities, esports, gameplay.
- "Shopping": purchasing, products/brands/prices, store/supermarket, commerce intent.
- "Social Interaction": interpersonal interaction, social situations, reaction content.
- "Sentiment": emotion, opinion, like/dislike, tone/affect judgments.
- "Egocentric Agent": first-person task execution / embodied action planning ("what should I do next", step-by-step manipulation), hands-on agentic behavior.

If unsure between two top categories, pick the best and lower confidence.
If unsure between multiple sub-categories, pick the closest and lower confidence.   
\end{judgeprompt}

\subsection{Prompts for CoT Generation}
\textsc{Gemini-2.5-Pro} is instructed with the following prompt to generate the pseudo ground truth CoT for each question-answer pair.
\begin{judgeprompt}{CoT Generation}
From the given video, analyze deeply using the audio and visual cues, and produce detailed chain-of-thought reasoning.

Question: {question}
Answer: {answer}

Please structure the reasoning as:

Here's the reasoning process:

1. Visual Cue:
- Timeframe:
- Description:
- Analysis:

2. Audio Cue:
- Timeframe:
- Description:
- Analysis:

Final Answer:
\end{judgeprompt}
\subsection{SAM3 Bounding Box}
To get the label for video frame-by-frame bounding box generation using SAM3, \textsc{GPT-4o-mini} is instructed to extract the object of interest from the question, answer and Gemini analysis (the pseudo ground truth CoT generated by \textsc{Gemini-2.5-pro}) with the following prompt:
\begin{judgeprompt}{Prompt for Generating SAM3 BB}
{
"role": "system",
"content": "You return ONLY one or two words. No punctuation. No extra text."
},
{
"role": "user",
"content": f"""You are generating a SINGLE bounding-box label for object tracking (SAM3).

Inputs:
Question: {question}
Answer: {answer}
Gemini-generated chain-of-thought for answering the above question from the video: {gemini_analysis}

Goal: Return ONLY 1-2 words naming a concrete, visible object that is present in the video and can be enclosed by a bounding box and is the object of interest of the gemini analysis / question-answer. The label MUST be a noun phrase (no verbs, no adjectives-only, no abstract concepts, no abilities, no emotions, no reasons, no dialogue).

Rules:
- If the QA is about something abstract, like an ability/action/concept (e.g., "Trample", "winning", "magic", "fear"), DO NOT output that. Instead output the most relevant visible entity performing it, e.g., name (not a Proper noun but a common noun) of a specific character, person, animal, vehicle, card, etc. from the video. Ensure the existence of this specific visible entry in the video from the visual cue of gemini analysis.
- If the QA is about spoken dialogue or "why" reasoning, output the key visible object involved in the event (e.g., "spoon", "baby", "woman", "door", "phone"). If unclear, output the main speaking subject (e.g., "woman" or "man").
- If the QA is about an attribute (color/shade/size), output the object that has the attribute (e.g., "dolphin", "car", "shirt").
- Prefer a specific object over generic ones when possible (e.g., "spoon" > "hand" > "person"). However, if only people are reliably present, output "person".
- If multiple objects are plausible, choose the one that is most central and consistently visible.
- If nothing is visible, output: person

Output format:
Return ONLY the label text, 1-2 words, lowercase, no punctuation, no quotes.
"""
}
\end{judgeprompt}

\subsection{Difficulty-Rating Procedure}
\label{sec:diff_rating_procedure}
We provide additional details on the difficulty-rating procedure used for dataset curation. Algorithm \ref{alg:difficulty_tier} outlines the procedure for deriving the difficulty rating from the LLM judge outputs. Below we provide the prompts for rollout generation and judging (the setup was described in the main paper) that we used.

\begin{judgeprompt}[unbreakable]{Difficulty Judging}
system = (
        "You are a strict evaluator using human-level semantic judgement. "
        "Do NOT do word-by-word matching. Consider paraphrases, synonyms, and concise equivalences."
    )

user_payload = {
    "question": question,
    "ground_truth_answer": ground_truth,
    "candidate_answers": answers16,
    "instructions": [
        "For each candidate answer, decide if it matches the ground-truth answer semantically (true/false).",
        "Then judge how similar the 16 candidate answers are to each other overall, "
        "on a scale 1-5 where 1 = most similar (lowest difficulty), 5 = least similar (highest difficulty).",
        "Return JSON only with keys: matches, similarity_rating, explanation.",
        "matches must be a list of exactly 16 booleans aligned with candidate_answers order.",
    ],
}    
\end{judgeprompt}

\begin{judgeprompt}{Rollout Generation}
"From the given video, analyze using both audio and visual cues and answer:\n"
f"{question}\n\n"
"Return only the answer. Don't be overly precise. Answers should strictly be brief and within a couple of words or a phrase.\n"
"DO NOT RETURN THE CHAIN-OF-THOUGHT, JUST THE ANSWER."   
\end{judgeprompt}

\begin{algorithm}[H]
\caption{Difficulty Tier Assignment}
\label{alg:difficulty_tier}

\begin{algorithmic}[1]

\Require $num\_not\_matching$, $difficulty\_score$
\Ensure $difficulty\_label$

\State \textbf{// Difficulty tier from disagreement}

\If{$num\_not\_matching \ge 10$}
    \State $disagreement\_level \gets$ \texttt{"difficult"}
\ElsIf{$num\_not\_matching \ge 5$}
    \State $disagreement\_level \gets$ \texttt{"medium"}
\Else
    \State $disagreement\_level \gets$ \texttt{"easy"}
\EndIf

\Statex
\State \textbf{// Difficulty tier from judge score}

\If{$difficulty\_score = 5$}
    \State $score\_level \gets$ \texttt{"difficult"}
\ElsIf{$difficulty\_score \in \{3,4\}$}
    \State $score\_level \gets$ \texttt{"medium"}
\Else
    \State $score\_level \gets$ \texttt{"easy"}
\EndIf

\Statex
\State \textbf{// Final difficulty = maximum severity}

\State $difficulty\_priority \gets \{\text{"easy"}:0,\ \text{"medium"}:1,\ \text{"difficult"}:2\}$

\State $difficulty\_label \gets 
\max(disagreement\_level, score\_level,\ key = difficulty\_priority)$

\Return $difficulty\_label$

\end{algorithmic}
\end{algorithm}

\subsection{Judge Prompts for Evaluation}
We provide the exact prompts used for the three LLM judges in our multi-judge evaluation pipeline: \textbf{Final Answer Accuracy}, \textbf{Hallucination}, and \textbf{Divergence}. Each judge returns a Likert-scale rating together with a brief textual justification.

\vspace{0.5em}

\noindent\textbf{Judge 1: Final Answer Accuracy.}
This judge evaluates whether the model's extracted final answer is semantically correct relative to the ground-truth final answer. The prompt is intentionally \emph{slightly lenient}: harmless paraphrases and plausible extra details are not penalized unless they alter the meaning or introduce contradiction.

\begin{judgeprompt}{Judge 1: Final Answer Accuracy}
You are an expert evaluator assessing the **correctness of a model's final answer** relative to a ground-truth final answer for a text-based question.

You will receive:
- A **question** (text only),
- A **model final answer (output)** --- extracted directly from the model's response,
- A **ground-truth final answer (answer)** --- the reference correct answer.

Your task is to evaluate how correct and semantically consistent the model's final answer is compared to the ground-truth final answer.

**Important:** Focus *only* on correctness. Do not penalize for stylistic differences, verbosity, or minor paraphrasing that preserves meaning. **Be slightly lenient**: if the model's answer captures the main idea and does not contradict the ground truth, prefer a better score rather than a worse one. If the model adds extra details beyond the ground truth, do NOT penalize it for being verbose --- reward it slightly if those added details are plausible, non-contradictory, and the core answer remains accurate. Do not consider the reasoning chain.

---

### Evaluation Guidelines (Lenient)

Rate **final answer correctness** on a Likert scale from **1 to 5**:

- **5** -> Perfect match in meaning with the ground truth. Minor wording differences are fine. **Any extra details are consistent (or at least not clearly wrong) and do not change the core answer.**
- **4** -> Largely correct and captures the main point, with only minor omissions or slight inaccuracies. **Extra details may be somewhat speculative but not clearly contradictory.**
- **3** -> Mixed: gets some key parts right but misses important elements or includes noticeable inaccuracies. **Extra details may introduce confusion or mild contradictions.**
- **2** -> Mostly wrong or misses the main point, even if it shares a few surface keywords with the ground truth.
- **1** -> Completely wrong, unrelated, or directly contradicts the ground truth.

When rating, focus on:
- Whether the core claim/conclusion matches the ground truth.
- Whether key entities, facts, or findings are correctly stated.
- Semantic consistency and faithfulness to the ground truth, not surface-level wording.
- **Leniency principle:** if you're unsure whether something is a minor vs. major issue, **err toward the higher score (more correct)** unless there is a clear contradiction.
- **Added details:**
  - **Good:** extra context/examples/clarifications that remain consistent with the ground truth.
  - **Bad:** extra details that clearly conflict with the ground truth or change the meaning.

---

### Input

**Question:**
"""{question}"""

**Model Final Answer (output):**
"""{output}"""

**Ground-Truth Final Answer (answer):**
"""{answer}"""

---

### Output Format

The **first line** of your response must contain only the JSON rating in the following format:
{{"rating": <integer between 1 and 5>}}

Subsequent lines should provide a **brief textual justification** explaining why you assigned that rating, including whether any added details helped or hurt correctness.

Example:
{{"rating": 4}}
The model captures the main conclusion correctly and adds extra explanation that is consistent with the ground truth, though it slightly overstates one detail.
\end{judgeprompt}

\vspace{0.8em}

\noindent\textbf{Judge 2: Hallucination.}
This judge measures whether the predicted chain of thought introduces unsupported entities, facts, assumptions, or reasoning steps relative to the ground-truth chain of thought.

\begin{judgeprompt}{Judge 2: Hallucination}
You are an expert evaluator trained to assess the **hallucination level** of a model's reasoning in text-based tasks.

You will receive:
- A **question** (text only),
- A **model-generated chain of thought (output)**,
- A **ground-truth chain of thought (answer)**.

Your task is to evaluate how much the model's reasoning (**output**) includes **unsupported, fabricated, or inaccurate details** that are not grounded in the provided **answer**.
Hallucination refers to **added entities, assumptions, facts, steps, or interpretations** that cannot be justified by the ground truth.

**Be careful not to penalize harmless rephrasings, synonymous wording, or logically equivalent reasoning.**
Small differences in wording or structure do **not** count as hallucination if they preserve the meaning and stay faithful to the context.

---

### Evaluation Guidelines

Rate **hallucination** on a Likert scale from **1 to 5**, based primarily on the **number and severity of added or contradicted objects, facts, or reasoning steps**:

- **5** -> Virtually no hallucination. The output may rephrase or paraphrase ideas, but introduces *no new factual objects* or incorrect claims.
- **4** -> Very minor hallucination. One or two mild additions appear, but they do not materially change the reasoning or introduce contradictions.
- **3** -> Moderate hallucination. A few added or altered facts or reasoning steps appear that are not supported by the ground truth.
- **2** -> High hallucination. Several important unsupported or contradictory claims are introduced.
- **1** -> Severe hallucination. The output introduces *many* fabricated objects or contradictory claims, or diverges substantially from the ground truth's factual basis.

When rating, focus on:
- Whether **objects**, facts, or reasoning steps in the **output** appear in or are justified by the **answer**.
- The **count** and **impact** of new, unsupported entities or claims.
- Whether the output's additions change the meaning, introduce speculation, or contradict the answer.
- **Do not penalize** paraphrasing, rewording, or alternative but valid restatements of the same objects/ideas.

---

### Input

**Question:**
"""{question}"""

**Model Output (Predicted Chain of Thought):**
"""{output}"""

**Ground Truth Chain of Thought:**
"""{answer}"""

---

### Output Format

The **first line** of your response must contain only the JSON rating in the following format:
{{"rating": <integer between 1 and 5>}}

Subsequent lines should provide a **brief textual justification** explaining why you assigned that rating.

Example:
{{"rating": 2}}
The output introduces several new entities and speculative steps not present in the ground truth, leading to a high hallucination score.
\end{judgeprompt}

\vspace{0.8em}

\noindent\textbf{Judge 3: Divergence.}
This judge measures how closely the model's reasoning aligns with the reference reasoning process, while explicitly ignoring conciseness and hallucination.

\begin{judgeprompt}{Judge 3: Divergence}
You are an expert evaluator trained to assess how closely a model's reasoning aligns with a reference reasoning process in text-based tasks.

You will receive:
- A **question** (text only),
- A **model-generated chain of thought (output)**,
- A **ground-truth chain of thought (answer)**.

Your task is to evaluate the *divergence* between the model's reasoning (**output**) and the reference reasoning (**answer**) --- that is, how much the model's reasoning process deviates in logic, structure, and interpretation of the question.

**Important:** *Ignore conciseness and hallucination issues entirely. This evaluation measures **only divergence**, independently from other facets.*

---

### Evaluation Guidelines

Assign a **divergence rating** on a Likert scale from **1 to 5**:

- **5** -> Identical or nearly identical reasoning (minimal divergence).
- **4** -> Slight divergence in focus or emphasis, but overall similar reasoning flow.
- **3** -> Noticeable differences in reasoning steps or logic, though partially aligned.
- **2** -> Major divergence in reasoning structure, logic, or conclusions.
- **1** -> Completely unrelated or incorrect reasoning.

When rating, focus on:
- Alignment of logical steps and intermediate conclusions.
- Consistency in interpretation of the question.
- Similarity of causal or deductive reasoning structure.
- *Ignore verbosity, formatting, or hallucination issues.*

---

### Input

**Question:**
"""{question}"""

**Model Output (Predicted Chain of Thought):**
"""{output}"""

**Ground Truth Chain of Thought:**
"""{answer}"""

---

### Output Format

The **first line** of your response must contain only the JSON rating in the following format:
{{"rating": <integer between 1 and 5>}}

Subsequent lines should provide a **brief textual justification** explaining why you assigned that rating.

Example:
{{"rating": 3}}
The model's reasoning partially aligns with the ground truth but introduces different intermediate steps and omits key logical links.
\end{judgeprompt}

\subsection{Additional Experiments}
We report extended results beyond those included in the main paper.

\subsection{Judge Agreement}
We report per-model Cohen's $\kappa$ scores to provide a more fine-grained view of inter-judge agreement beyond the aggregate results in the main paper. As shown in Figure~\ref{fig:judge_agreement_kappa}, these plots illustrate how consistently the judge models rate each evaluated model across the \textit{Divergence}, \textit{Hallucination}, and \textit{Final Answer} dimensions.

\begin{figure*}[t]
    \centering

    \begin{subfigure}[t]{0.31\textwidth}
        \centering
        \includegraphics[width=\linewidth]{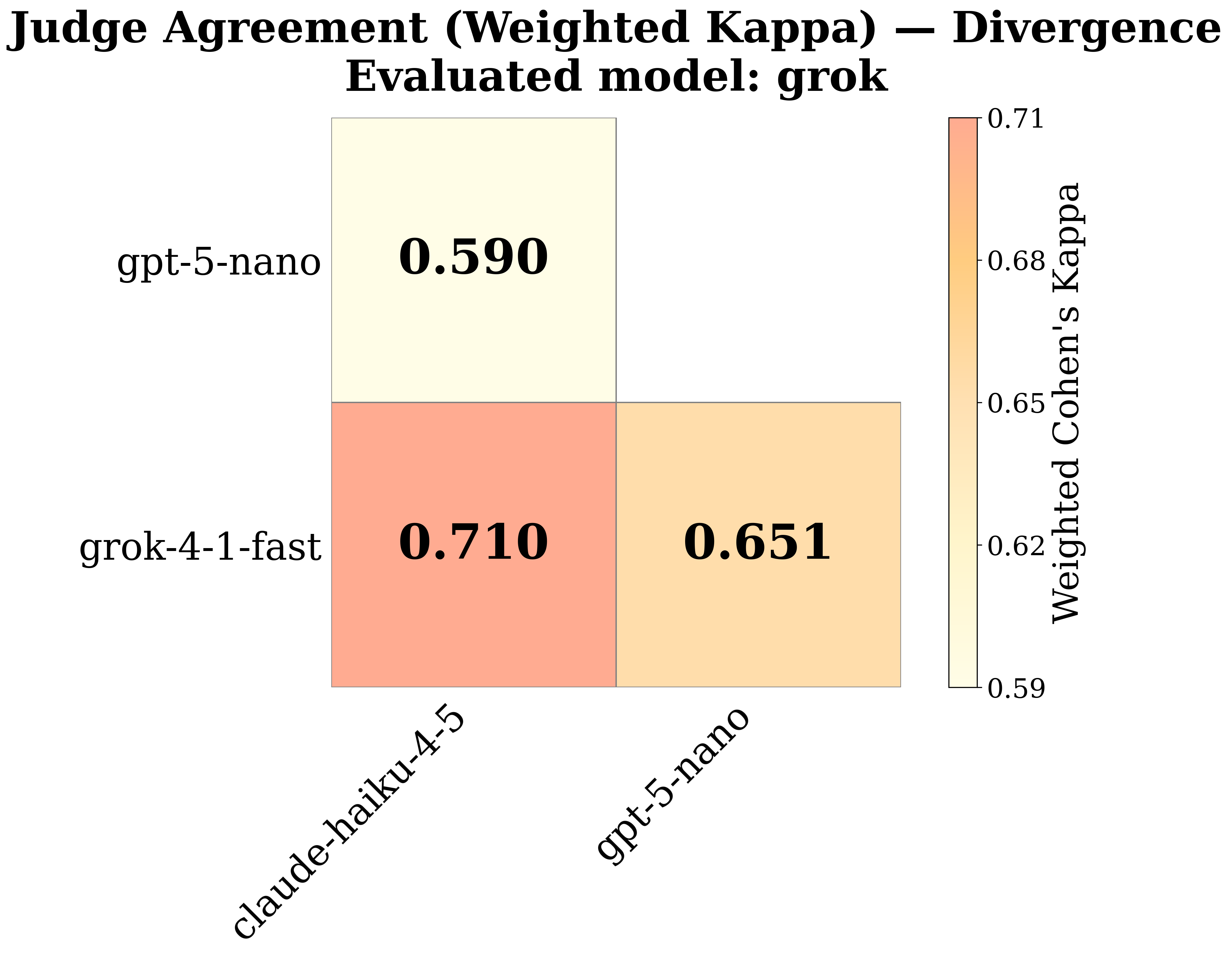}
    \end{subfigure}\hfill
    \begin{subfigure}[t]{0.31\textwidth}
        \centering
        \includegraphics[width=\linewidth]{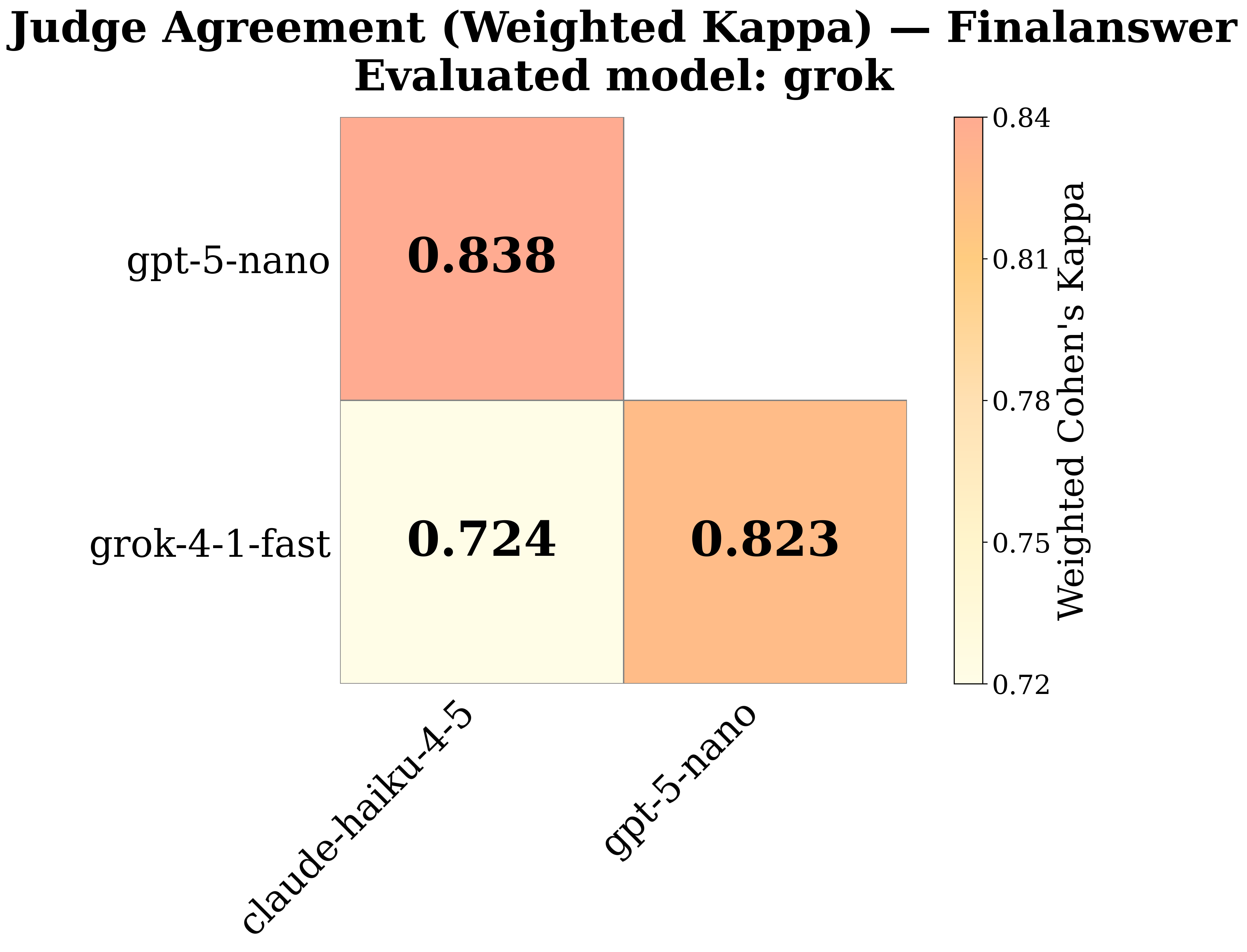}
    \end{subfigure}\hfill
    \begin{subfigure}[t]{0.31\textwidth}
        \centering
        \includegraphics[width=\linewidth]{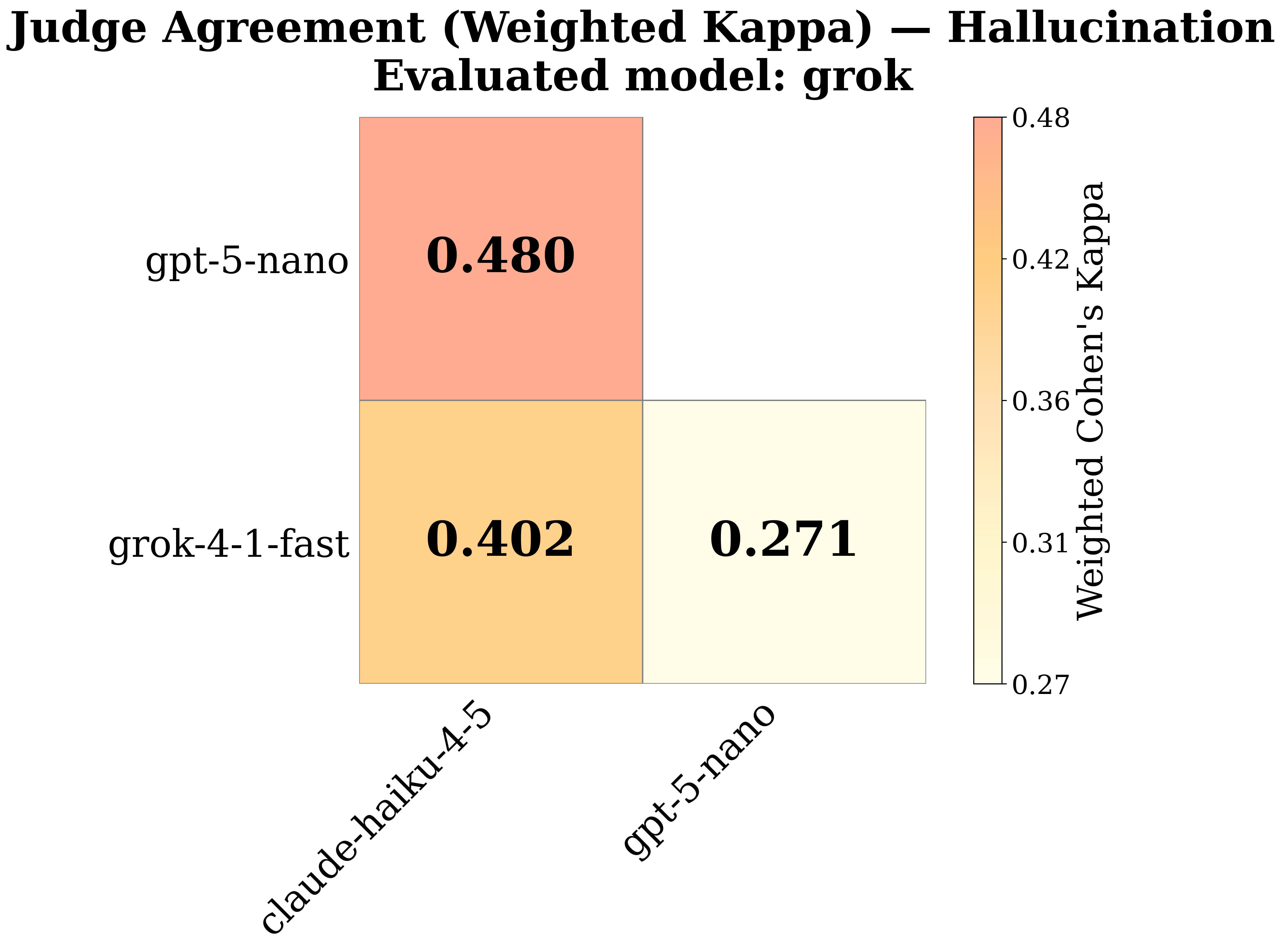}
    \end{subfigure}

    \vspace{0.5em}

    \begin{subfigure}[t]{0.31\textwidth}
        \centering
        \includegraphics[width=\linewidth]{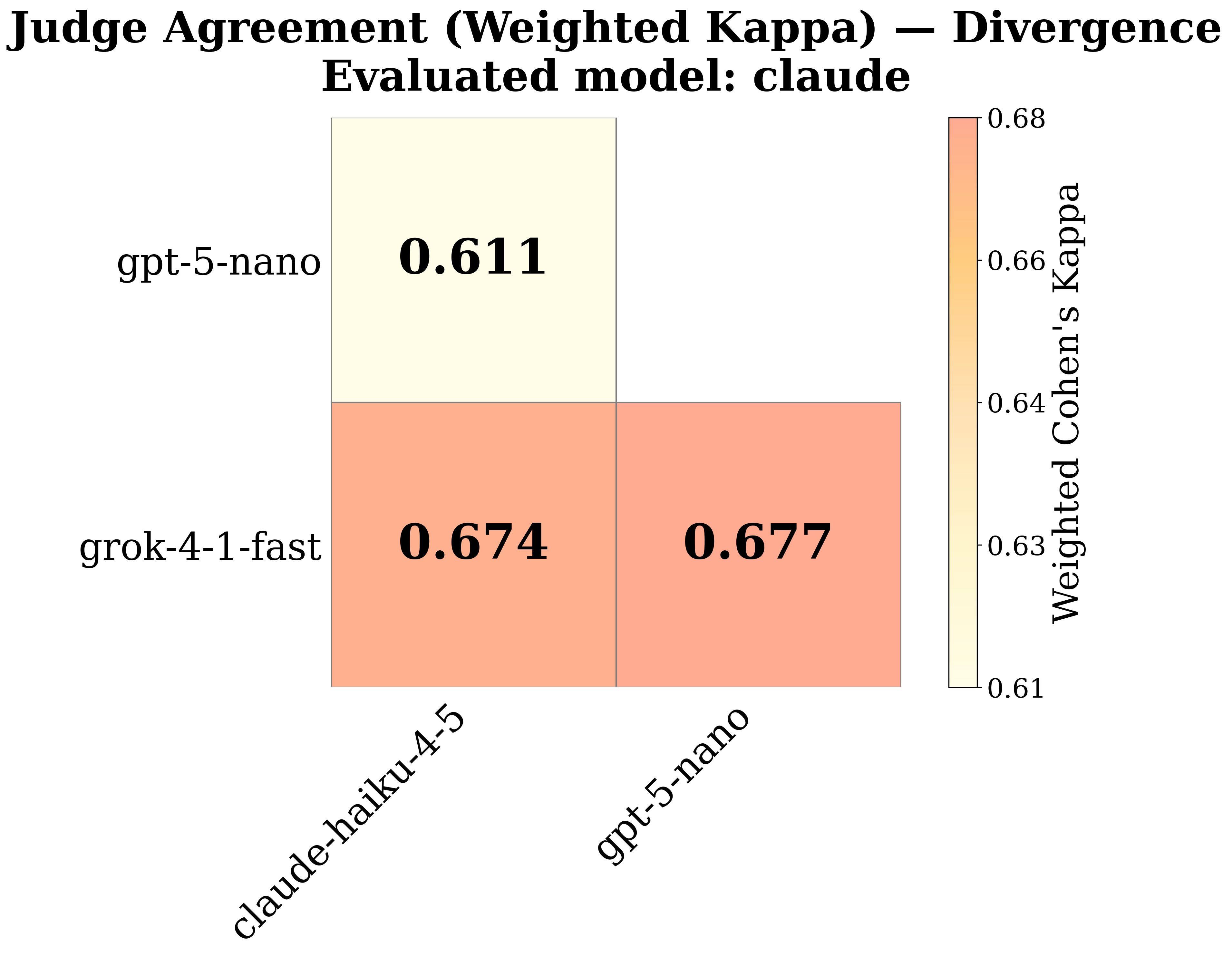}
    \end{subfigure}\hfill
    \begin{subfigure}[t]{0.31\textwidth}
        \centering
        \includegraphics[width=\linewidth]{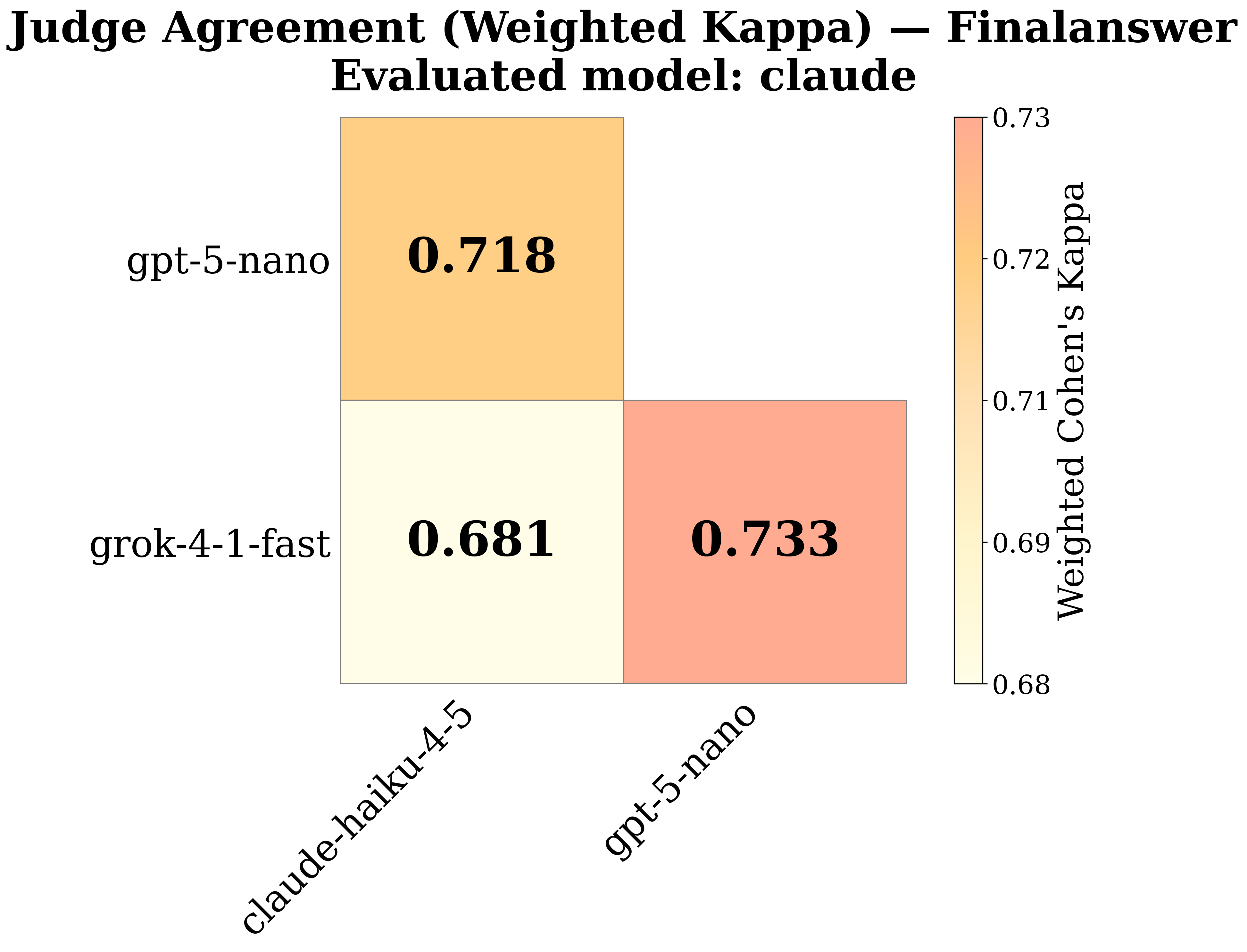}
    \end{subfigure}\hfill
    \begin{subfigure}[t]{0.31\textwidth}
        \centering
        \includegraphics[width=\linewidth]{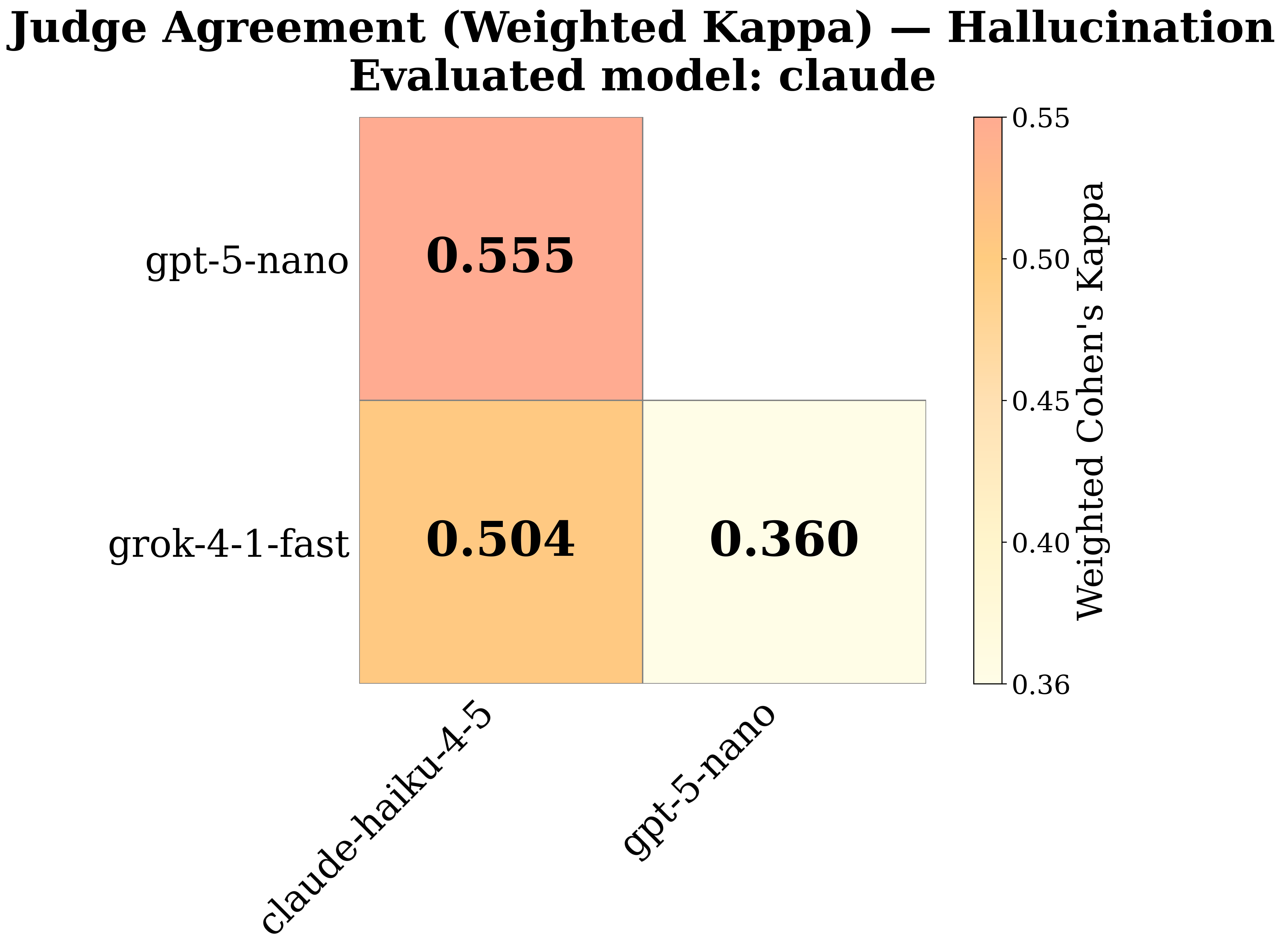}
    \end{subfigure}

    \vspace{0.5em}

    \begin{subfigure}[t]{0.31\textwidth}
        \centering
        \includegraphics[width=\linewidth]{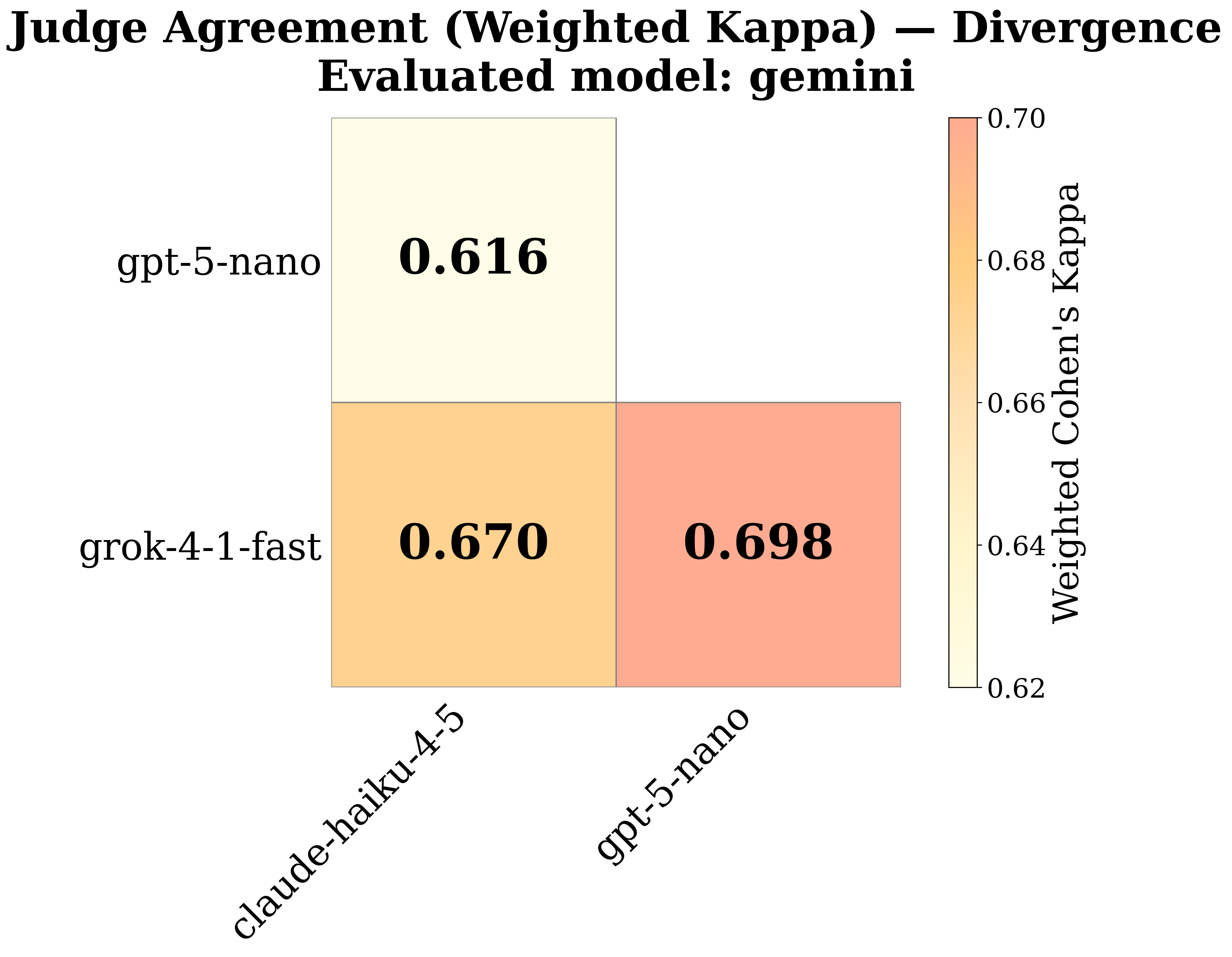}
    \end{subfigure}\hfill
    \begin{subfigure}[t]{0.31\textwidth}
        \centering
        \includegraphics[width=\linewidth]{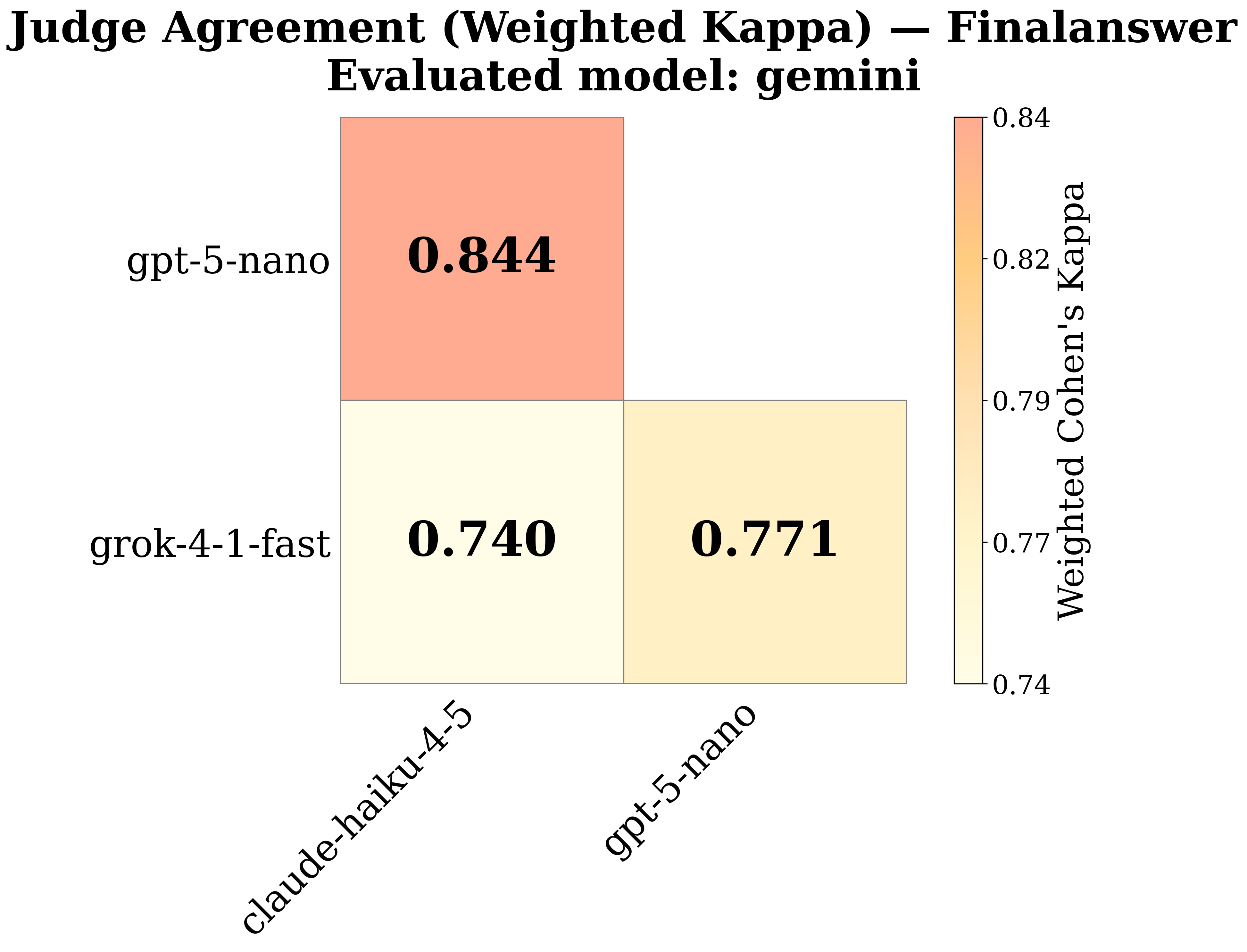}
    \end{subfigure}\hfill
    \begin{subfigure}[t]{0.31\textwidth}
        \centering
        \includegraphics[width=\linewidth]{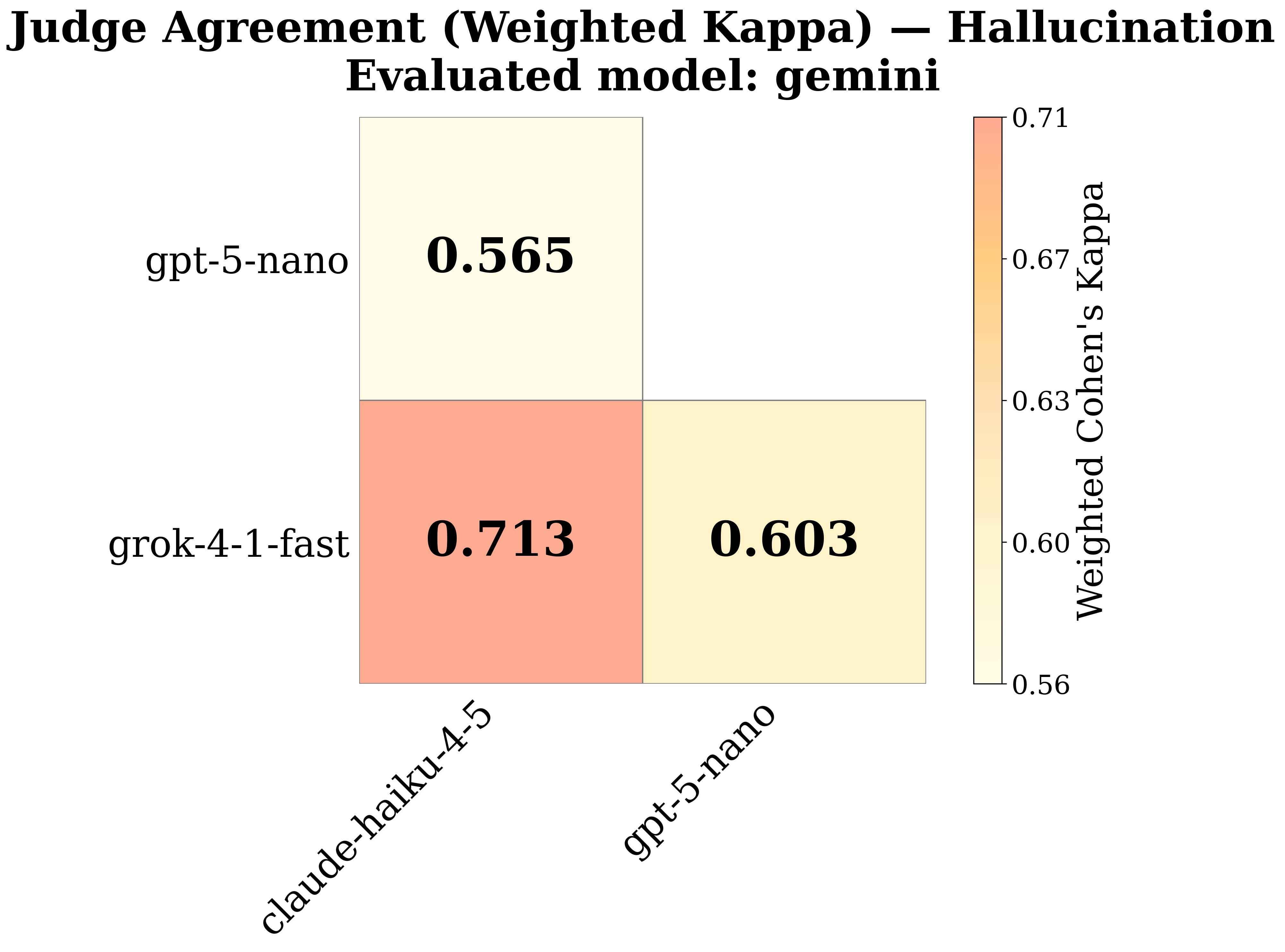}
    \end{subfigure}

    \vspace{0.5em}

    \begin{subfigure}[t]{0.31\textwidth}
        \centering
        \includegraphics[width=\linewidth]{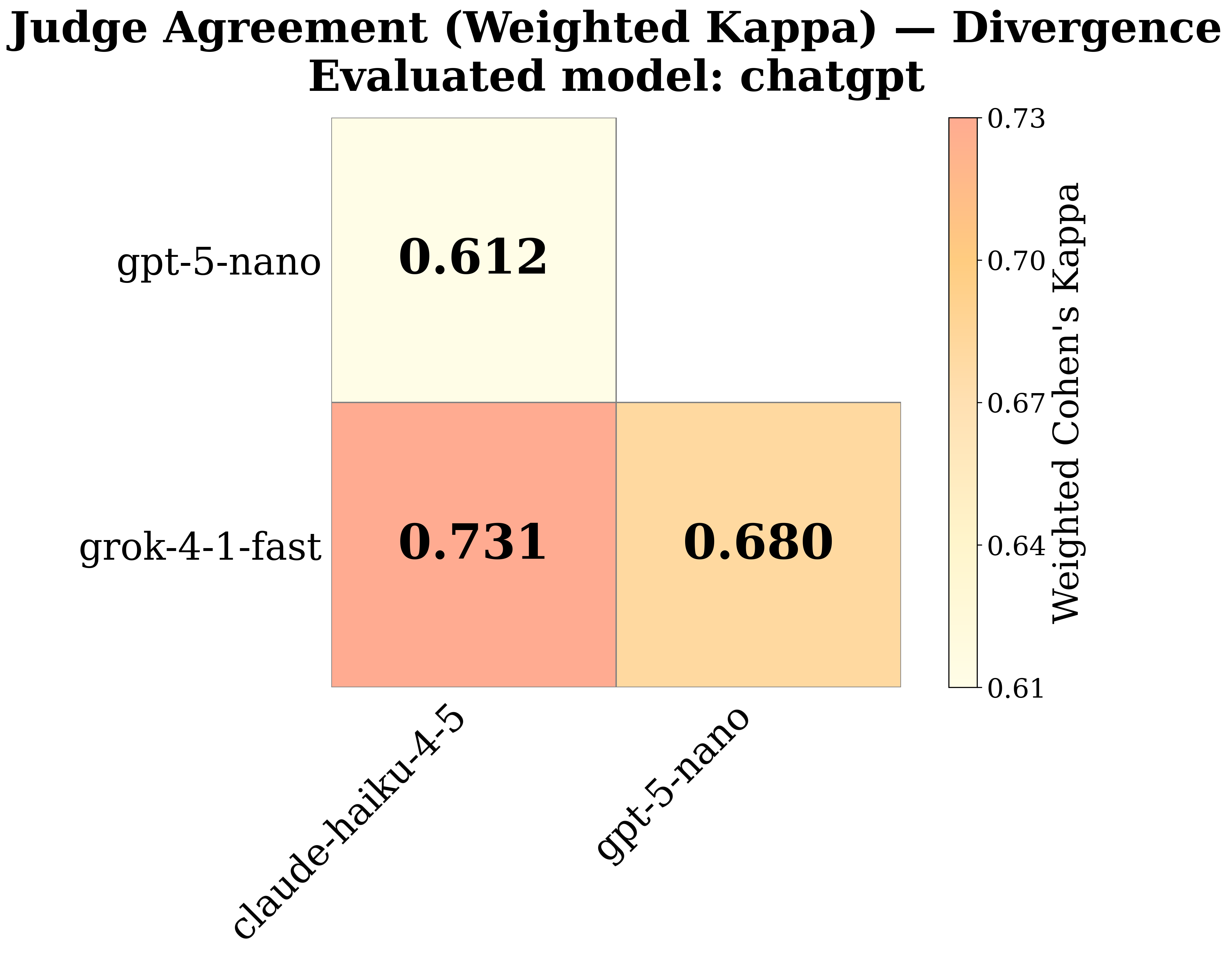}
    \end{subfigure}\hfill
    \begin{subfigure}[t]{0.31\textwidth}
        \centering
        \includegraphics[width=\linewidth]{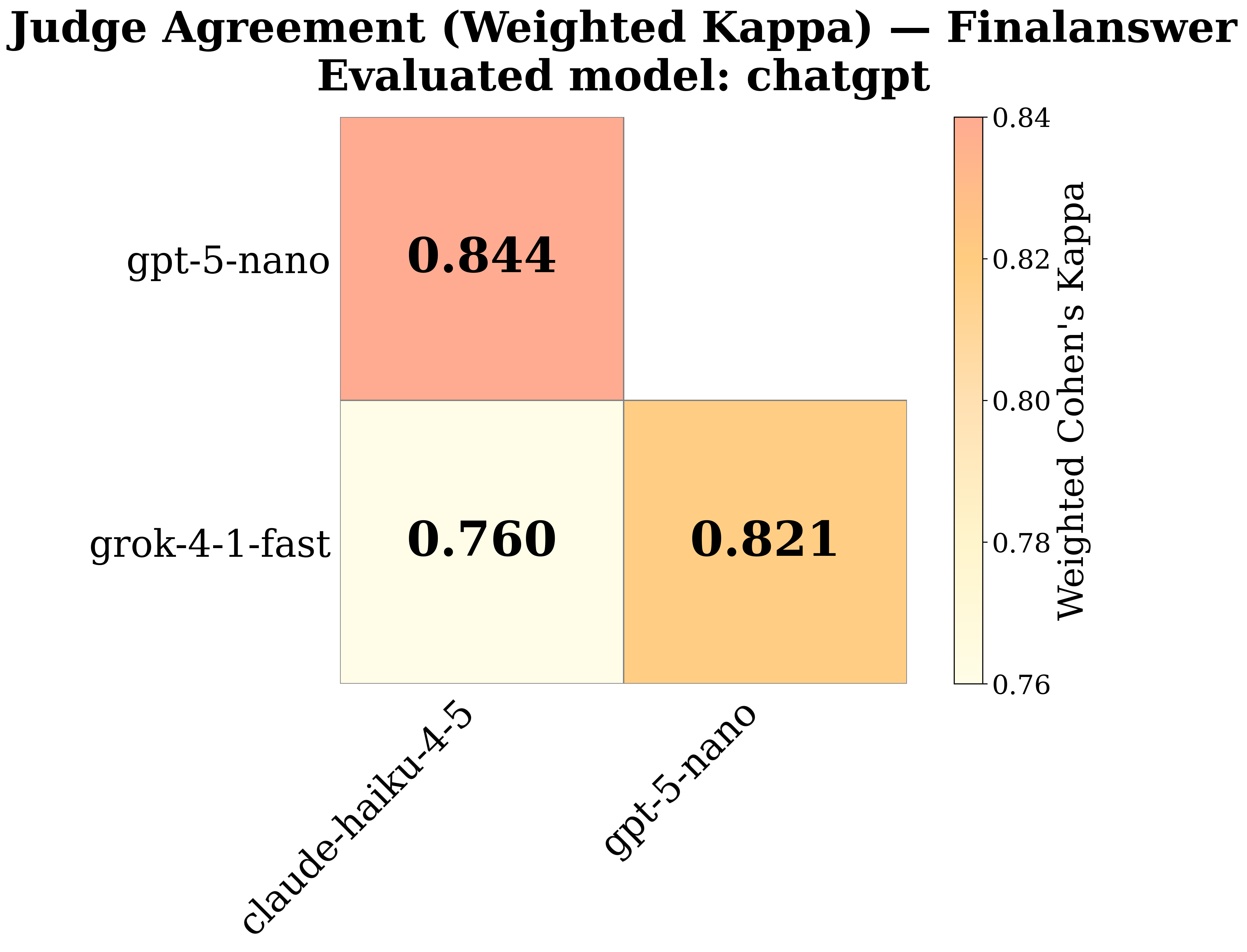}
    \end{subfigure}\hfill
    \begin{subfigure}[t]{0.31\textwidth}
        \centering
        \includegraphics[width=\linewidth]{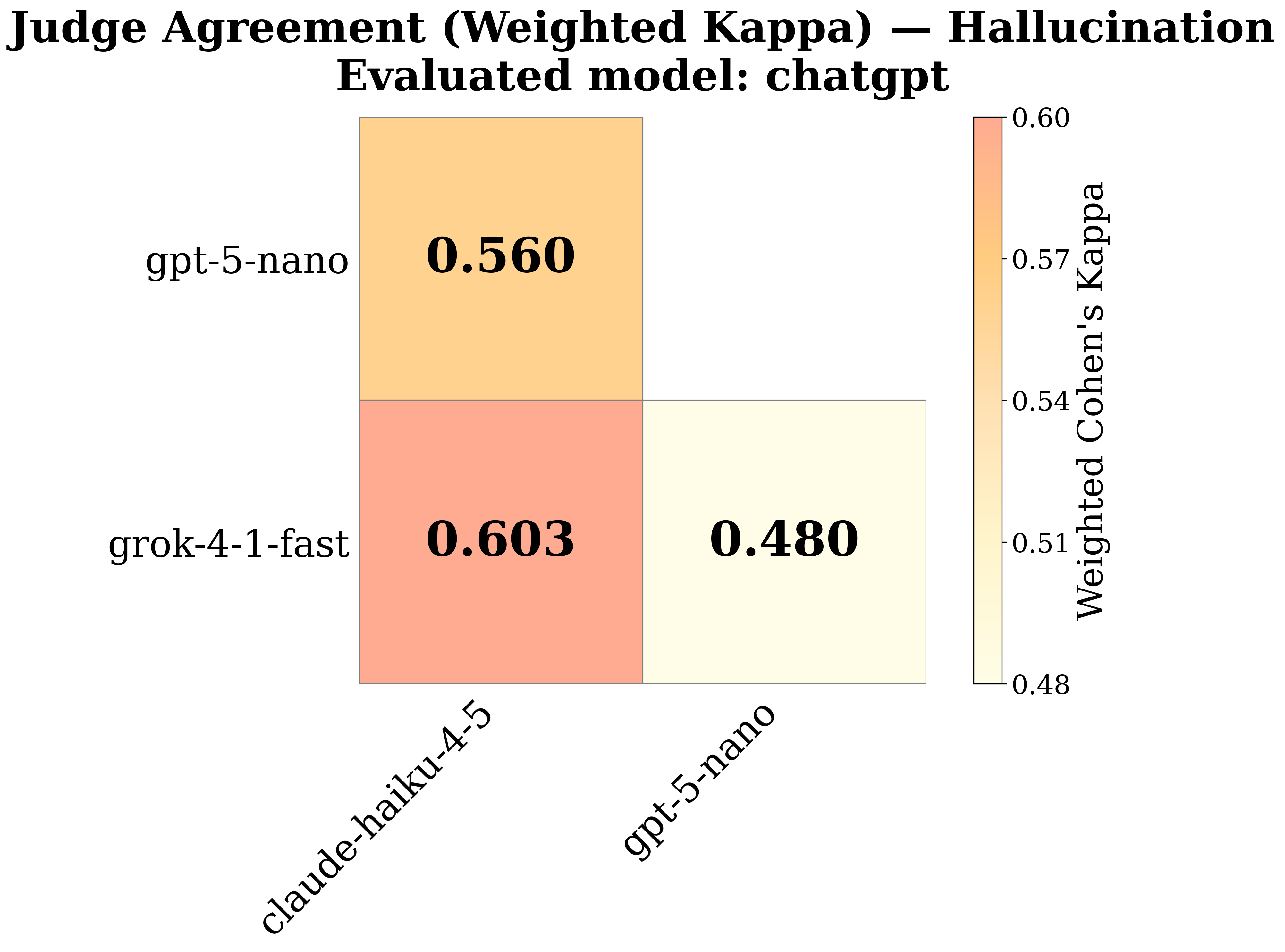}
    \end{subfigure}

    \vspace{0.5em}

    \begin{subfigure}[t]{0.31\textwidth}
        \centering
        \includegraphics[width=\linewidth]{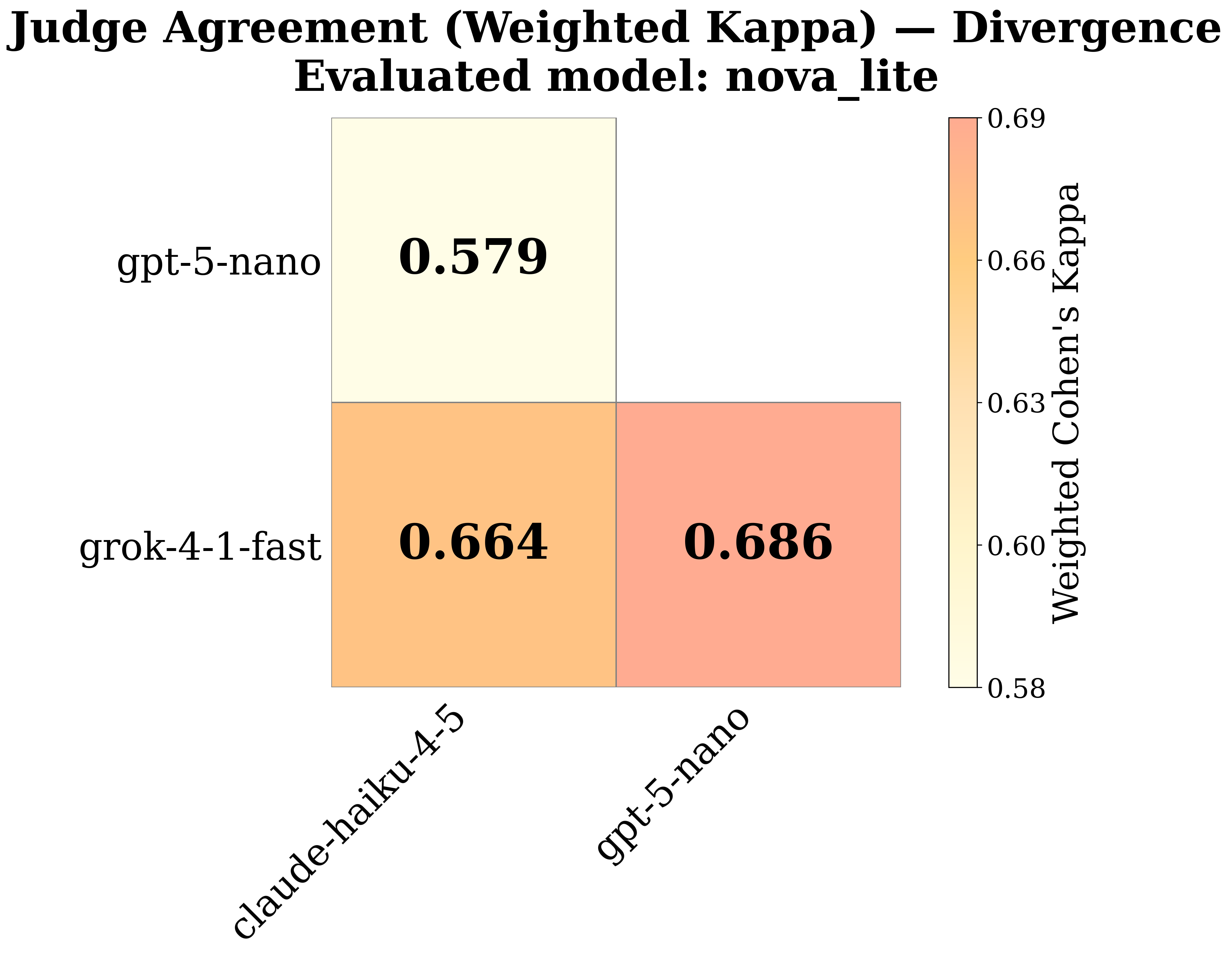}
    \end{subfigure}\hfill
    \begin{subfigure}[t]{0.31\textwidth}
        \centering
        \includegraphics[width=\linewidth]{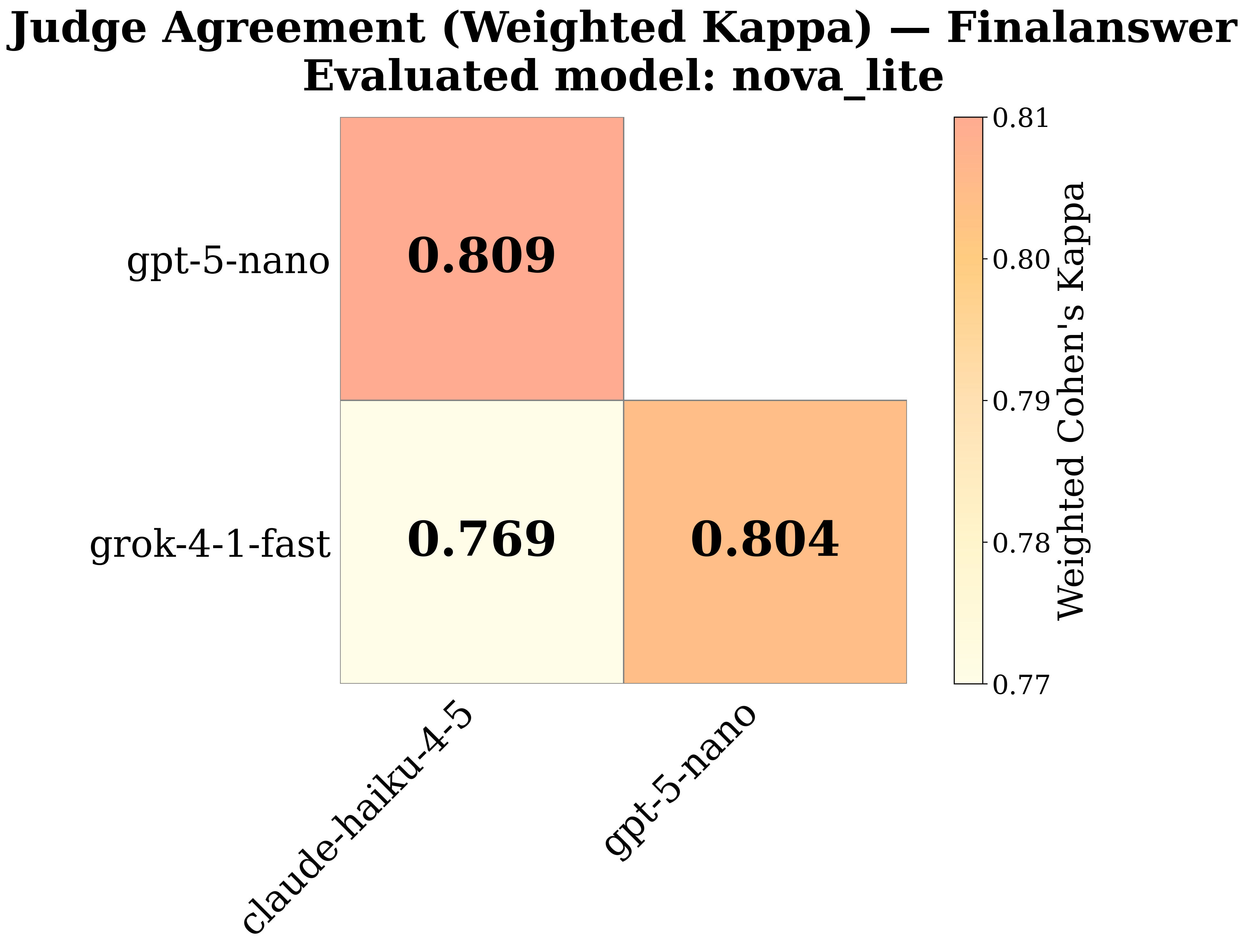}
    \end{subfigure}\hfill
    \begin{subfigure}[t]{0.31\textwidth}
        \centering
        \includegraphics[width=\linewidth]{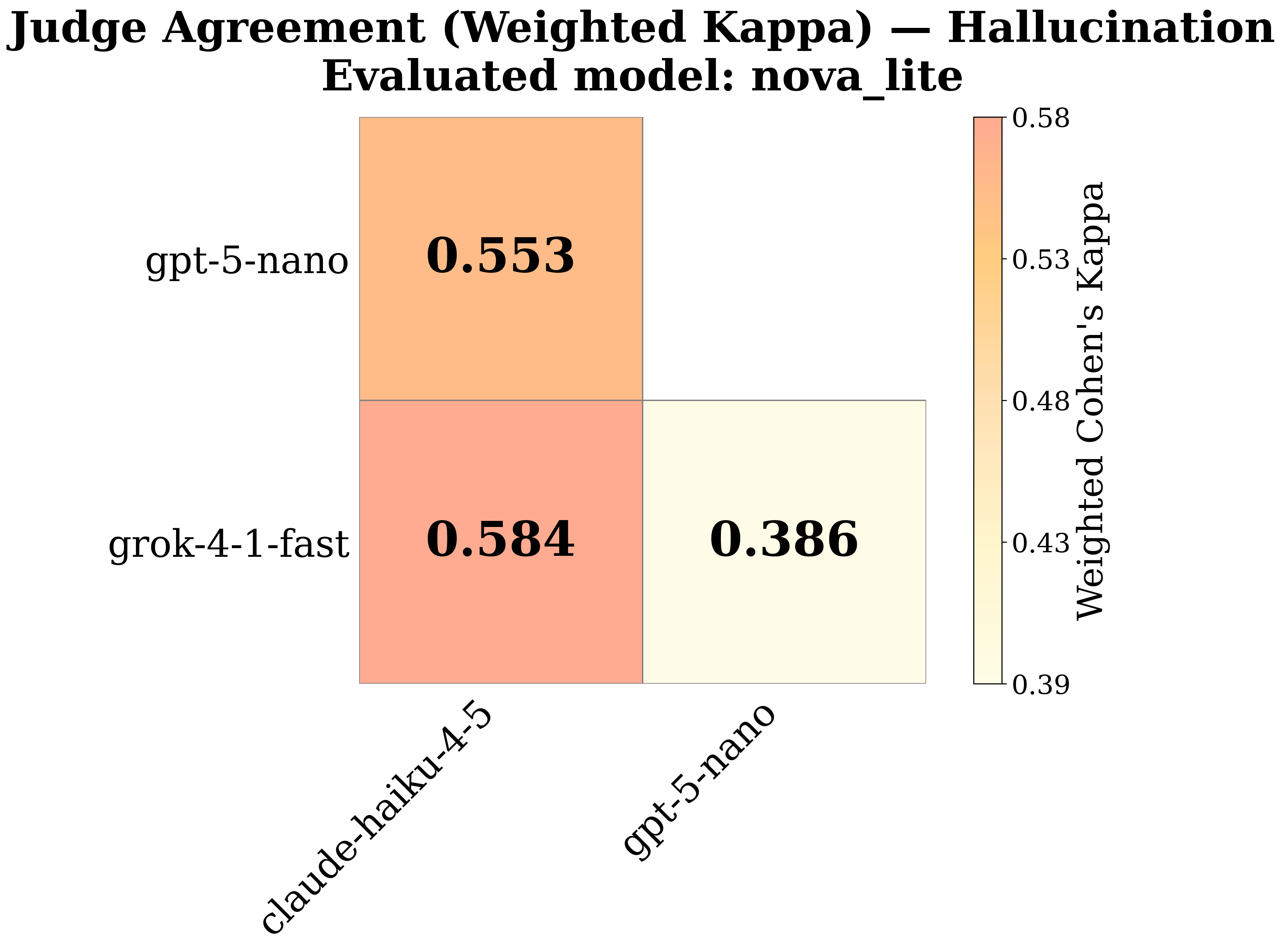}
    \end{subfigure}

    \caption{\textbf{Per-model inter-judge agreement measured by Cohen's $\kappa$.}
    Rows denote models and columns denote evaluation dimensions: \textit{Divergence}, \textit{Final Answer}, and \textit{Hallucination}.}
    \label{fig:judge_agreement_kappa}
\end{figure*}

\begin{figure*}[t]
    \ContinuedFloat
    \centering

    \begin{subfigure}[t]{0.31\textwidth}
        \centering
        \includegraphics[width=\linewidth]{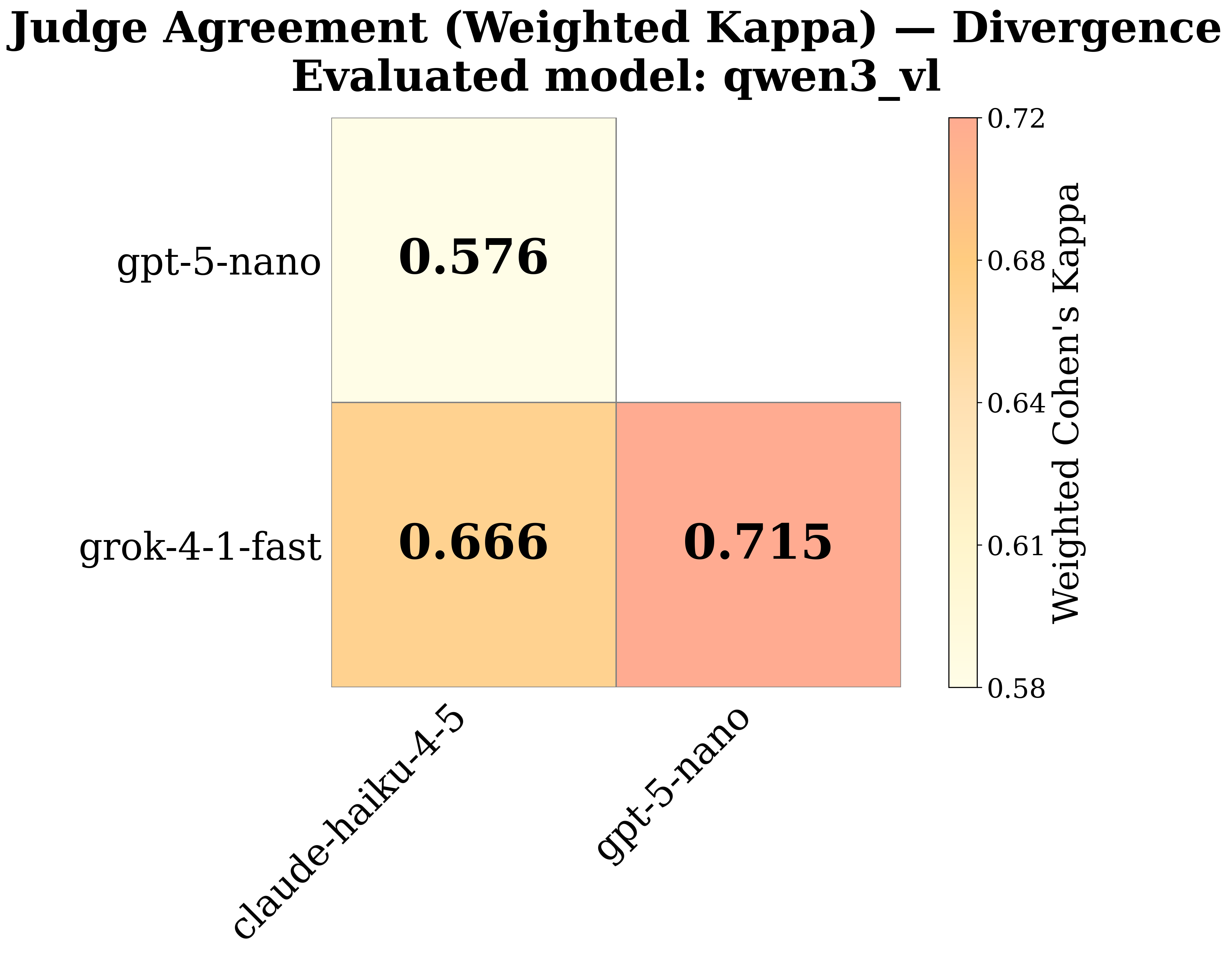}
    \end{subfigure}\hfill
    \begin{subfigure}[t]{0.31\textwidth}
        \centering
        \includegraphics[width=\linewidth]{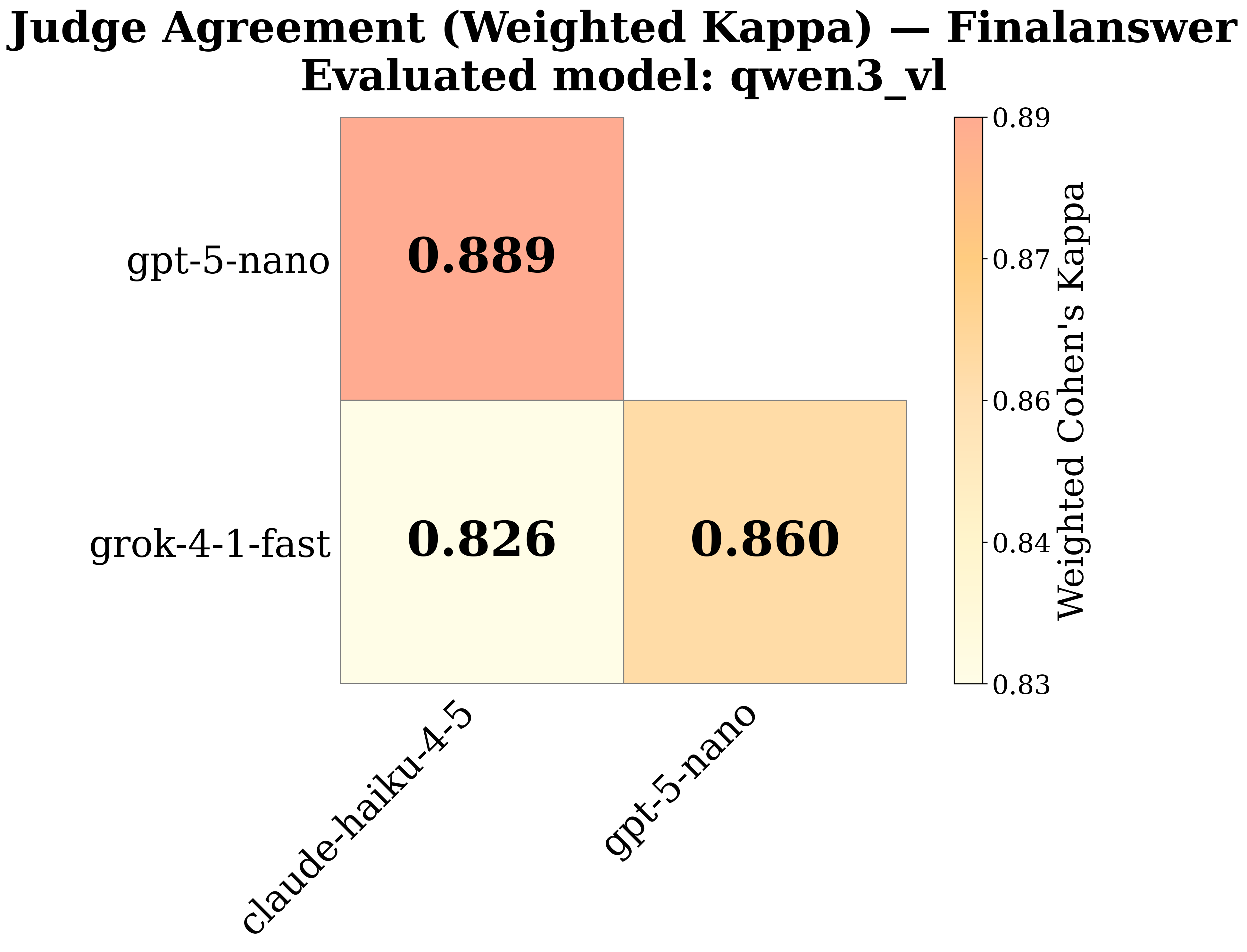}
    \end{subfigure}\hfill
    \begin{subfigure}[t]{0.31\textwidth}
        \centering
        \includegraphics[width=\linewidth]{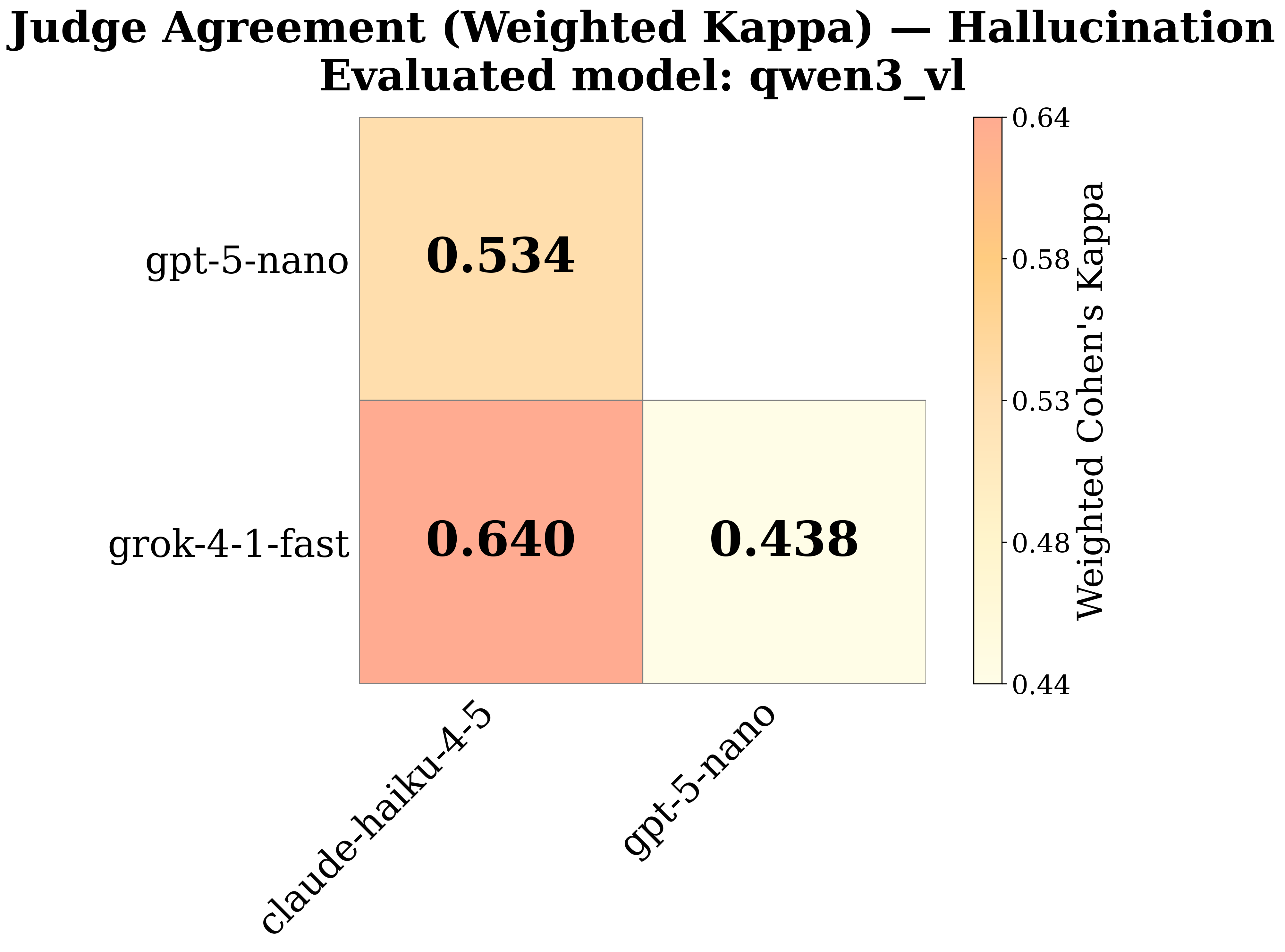}
    \end{subfigure}

    \vspace{0.5em}

    \begin{subfigure}[t]{0.31\textwidth}
        \centering
        \includegraphics[width=\linewidth]{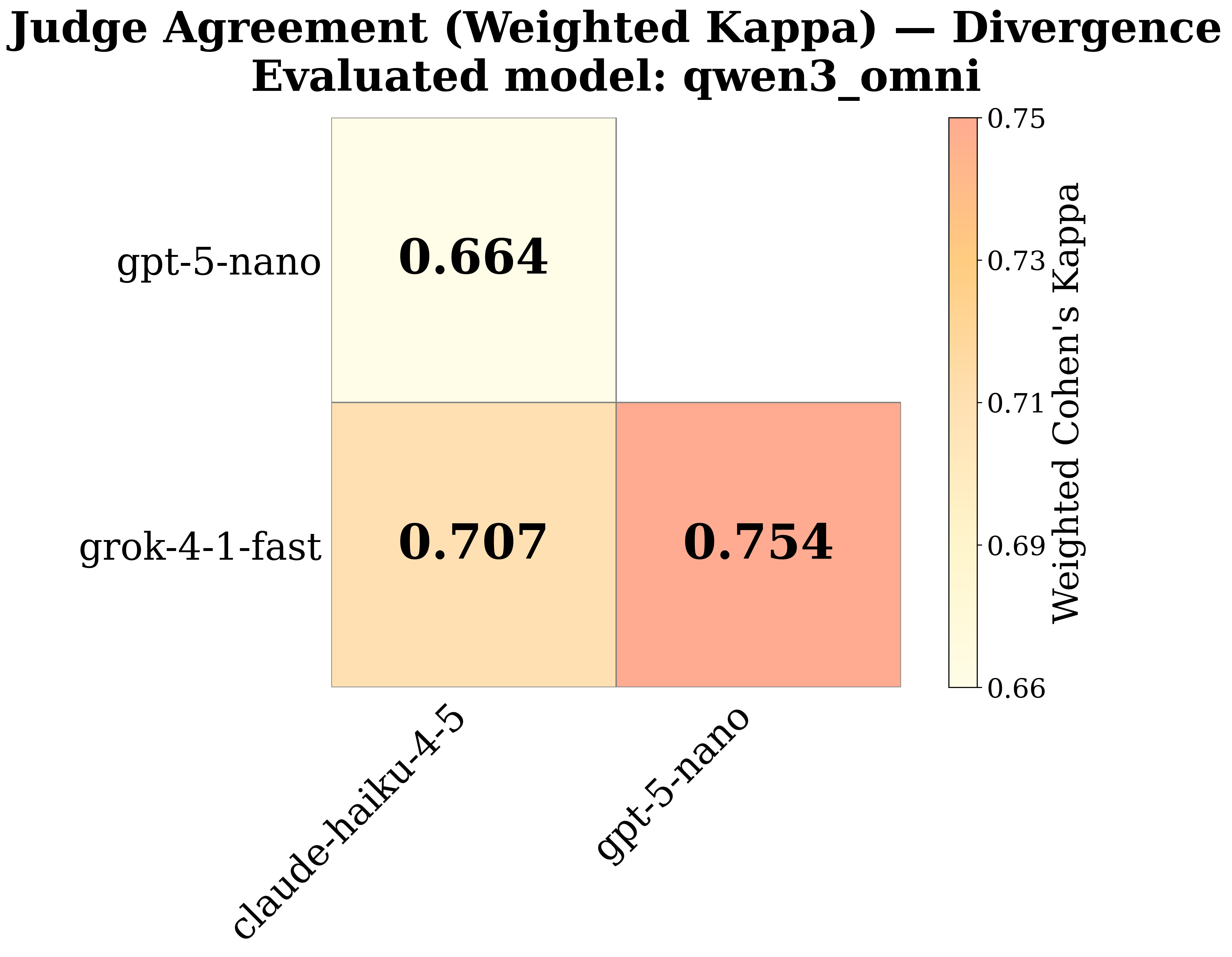}
    \end{subfigure}\hfill
    \begin{subfigure}[t]{0.31\textwidth}
        \centering
        \includegraphics[width=\linewidth]{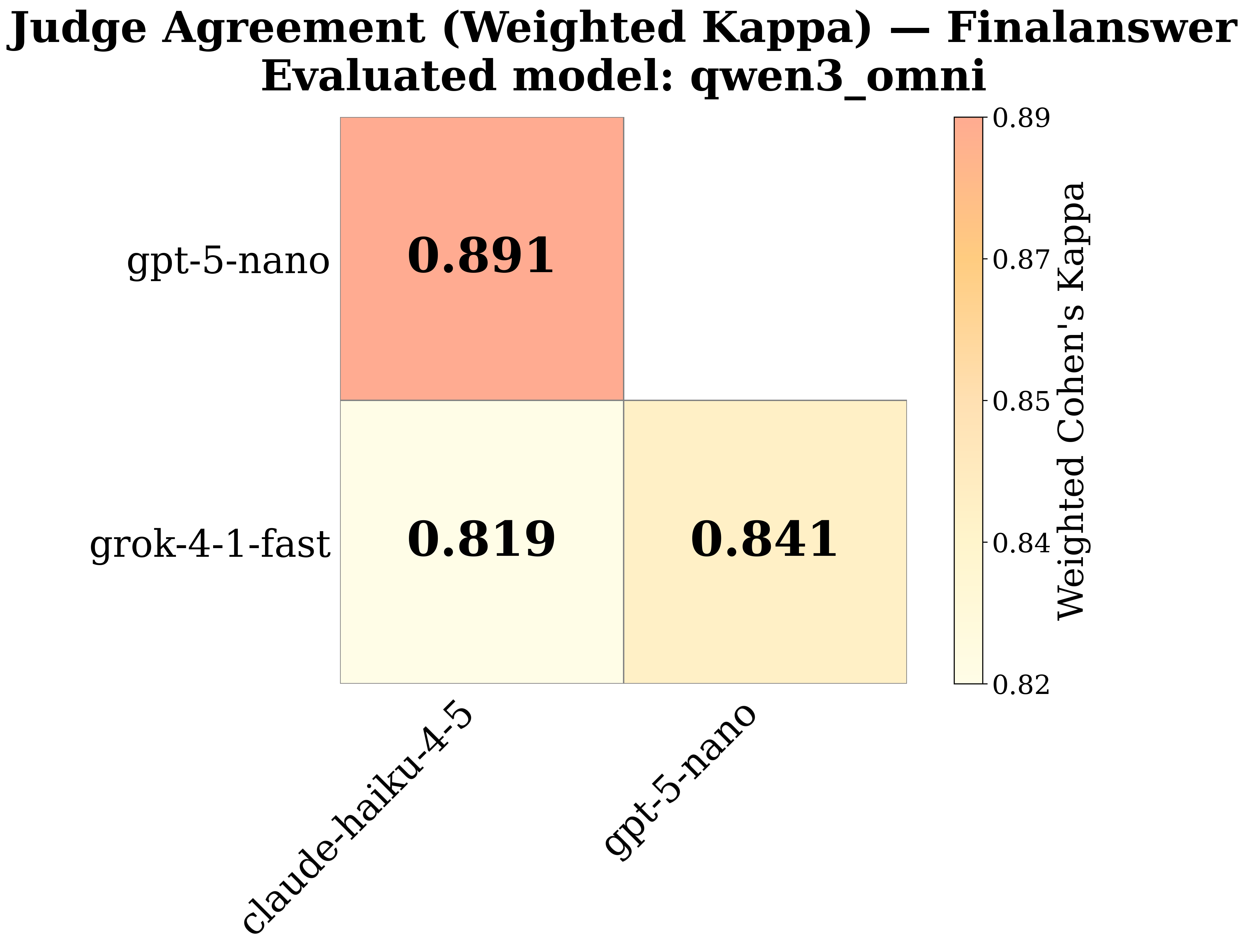}
    \end{subfigure}\hfill
    \begin{subfigure}[t]{0.31\textwidth}
        \centering
        \includegraphics[width=\linewidth]{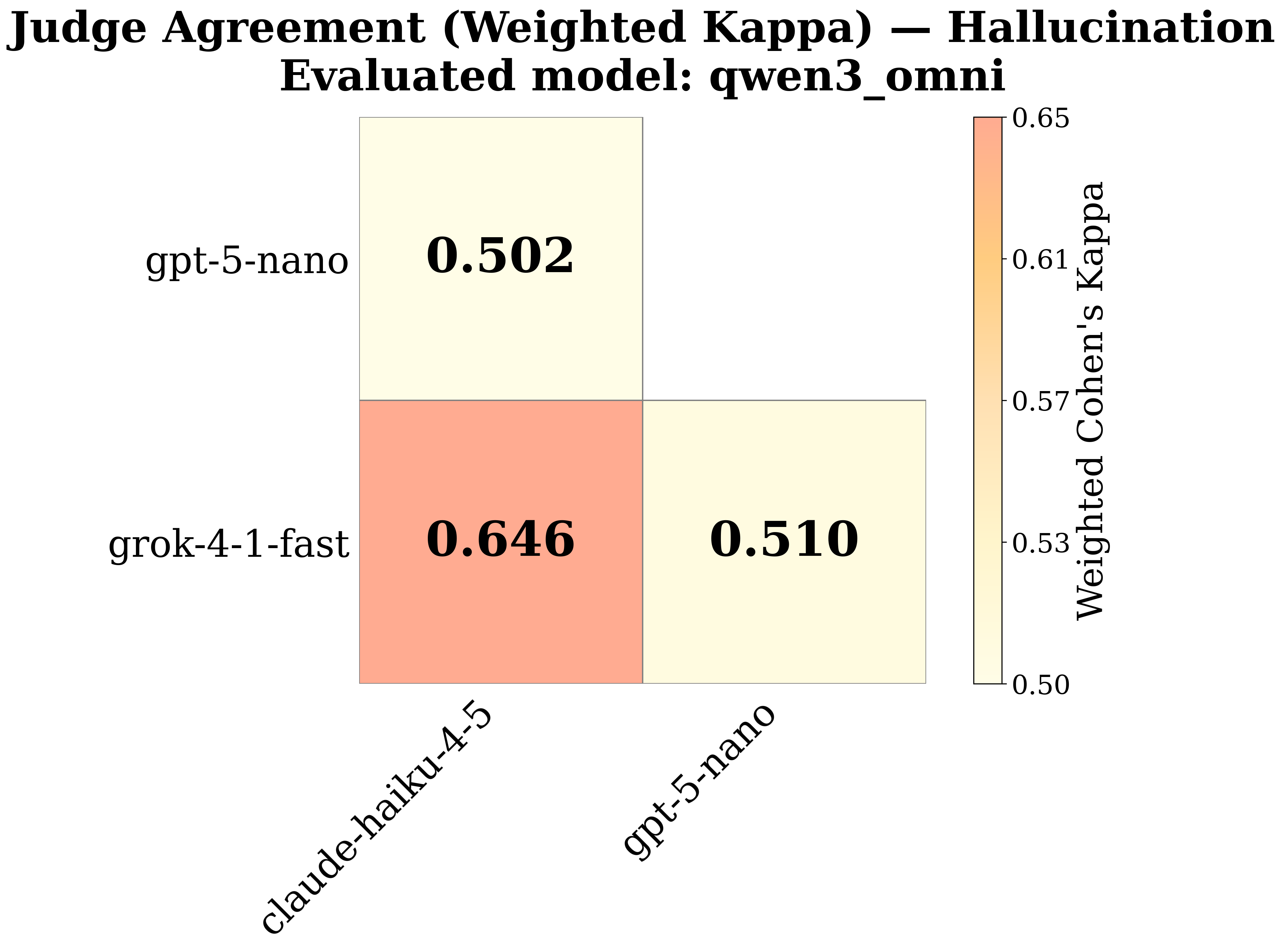}
    \end{subfigure}

    \vspace{0.5em}

    \begin{subfigure}[t]{0.31\textwidth}
        \centering
        \includegraphics[width=\linewidth]{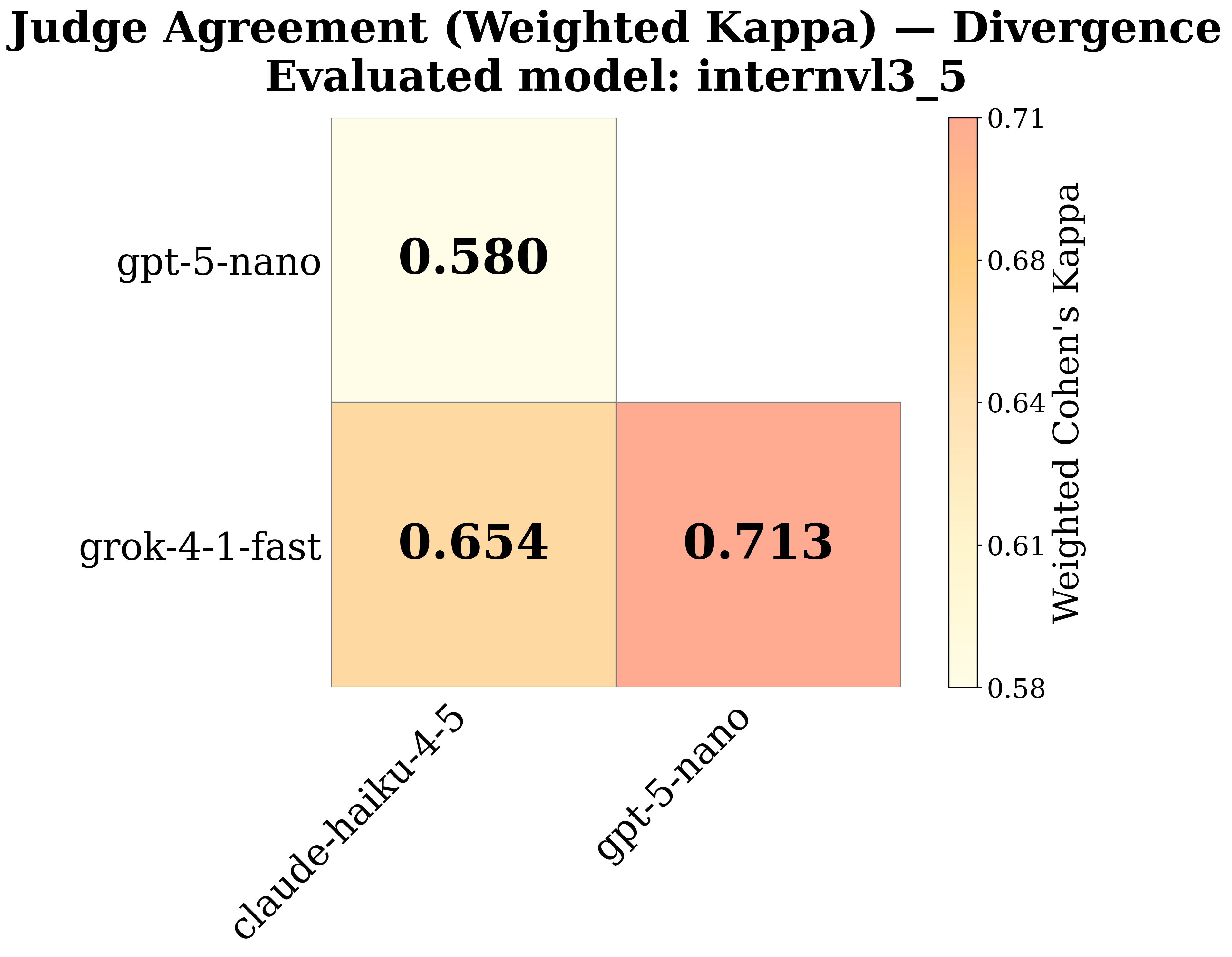}
    \end{subfigure}\hfill
    \begin{subfigure}[t]{0.31\textwidth}
        \centering
        \includegraphics[width=\linewidth]{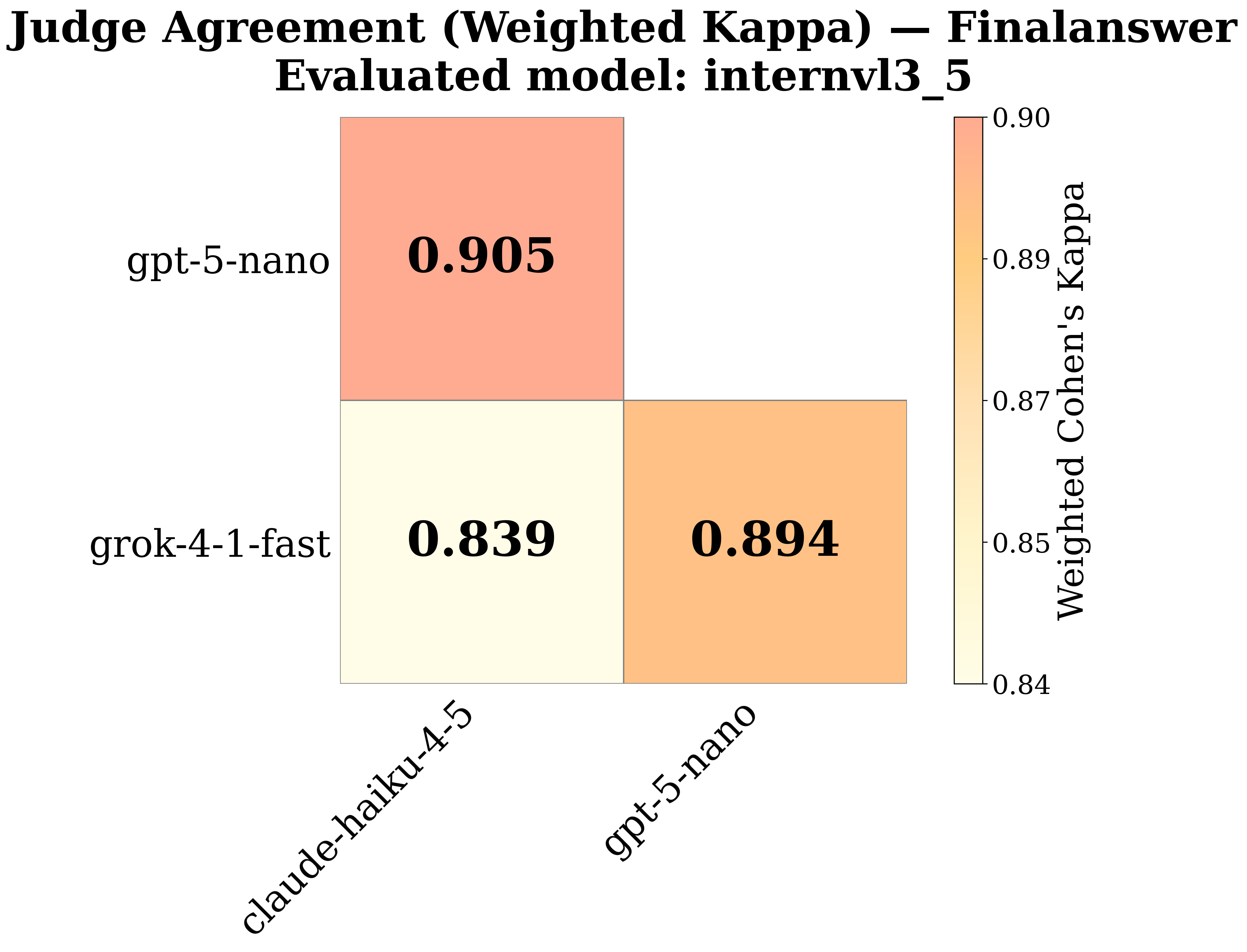}
    \end{subfigure}\hfill
    \begin{subfigure}[t]{0.31\textwidth}
        \centering
        \includegraphics[width=\linewidth]{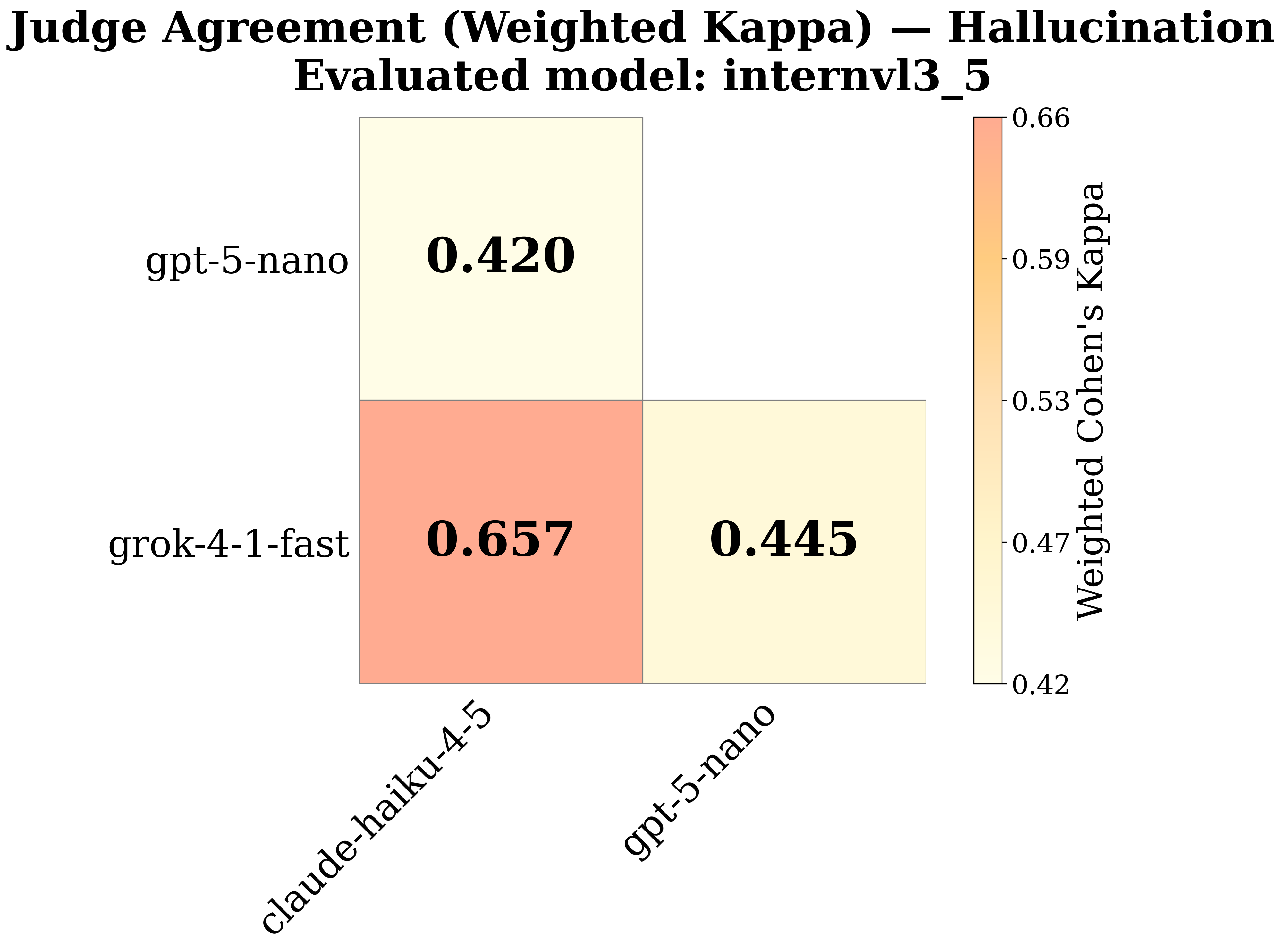}
    \end{subfigure}

    \vspace{0.5em}

    \begin{subfigure}[t]{0.31\textwidth}
        \centering
        \includegraphics[width=\linewidth]{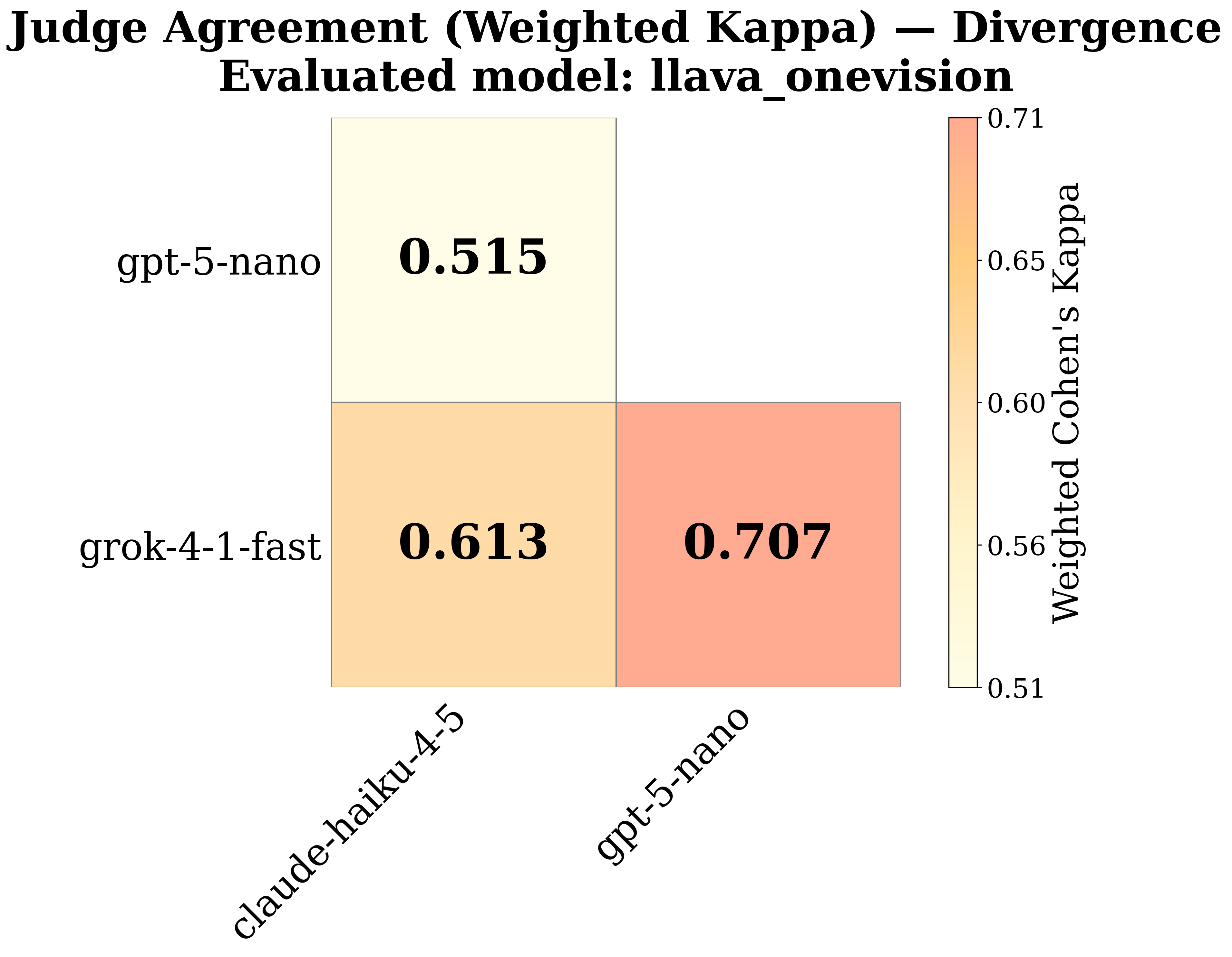}
    \end{subfigure}\hfill
    \begin{subfigure}[t]{0.31\textwidth}
        \centering
        \includegraphics[width=\linewidth]{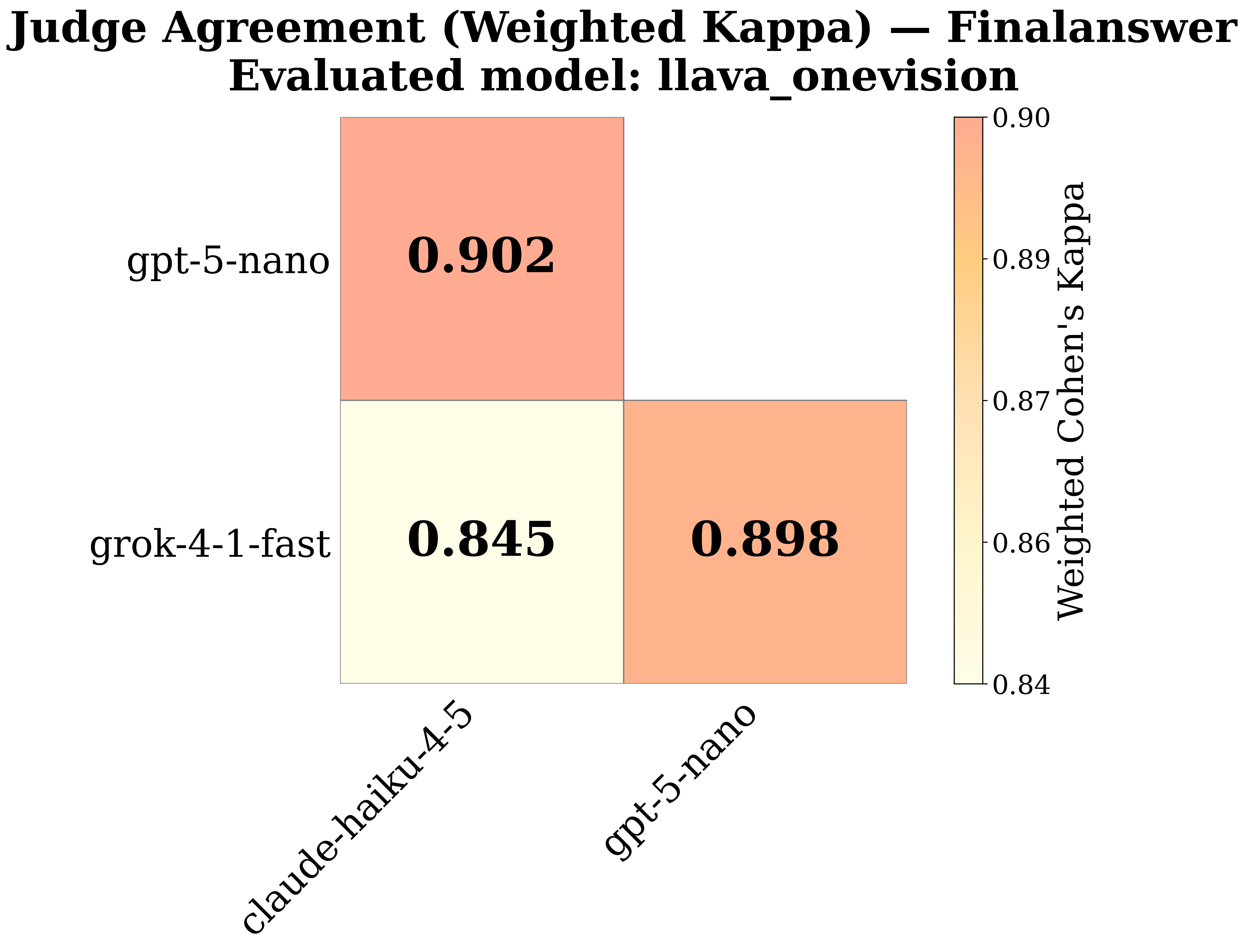}
    \end{subfigure}\hfill
    \begin{subfigure}[t]{0.31\textwidth}
        \centering
        \includegraphics[width=\linewidth]{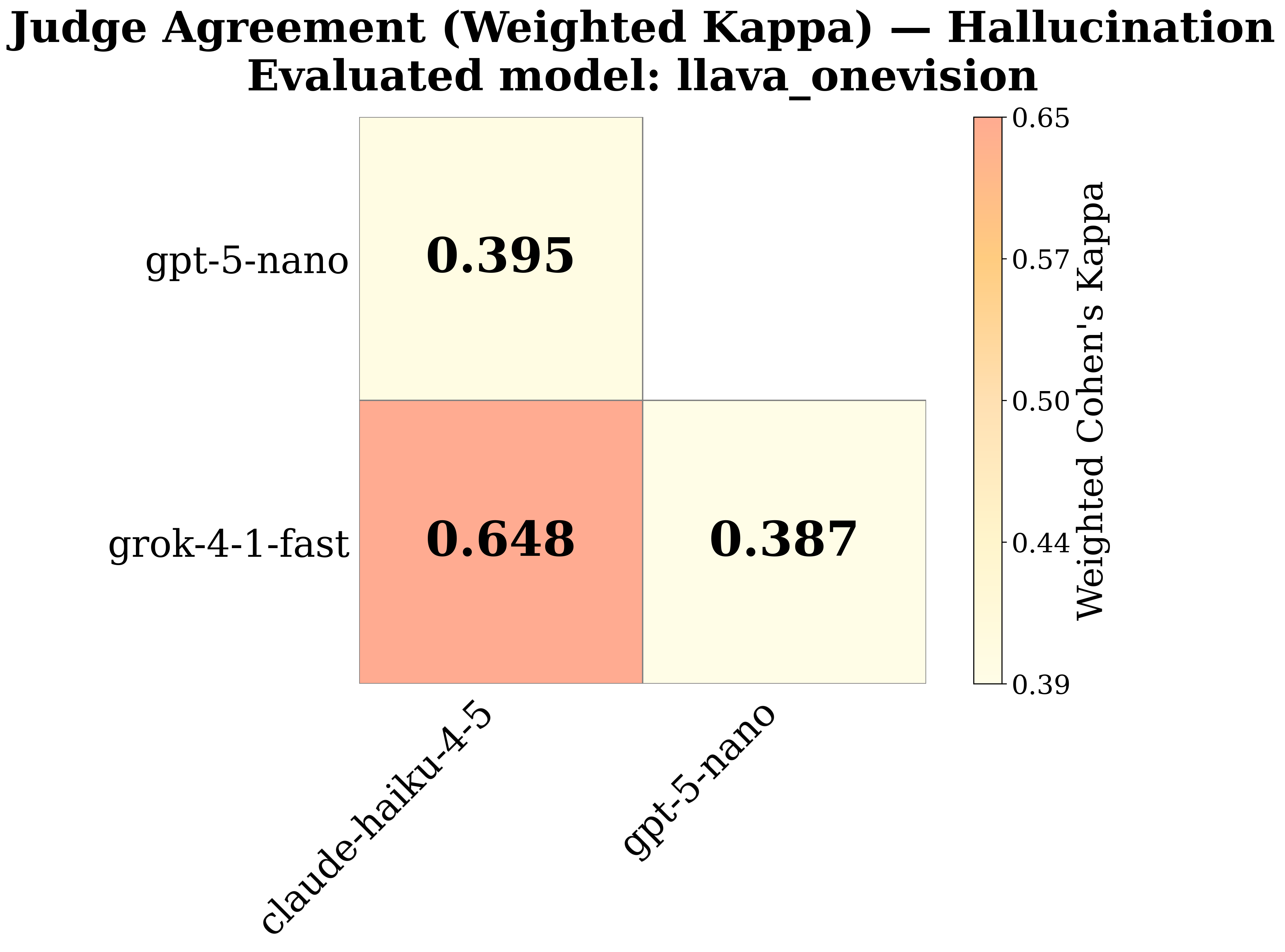}
    \end{subfigure}

    \caption{\textbf{Per-model inter-judge agreement measured by Cohen's $\kappa$ continued.}}
\end{figure*}

\begin{figure*}[t]
    \ContinuedFloat
    \centering


    \begin{subfigure}[t]{0.31\textwidth}
        \centering
        \includegraphics[width=\linewidth]{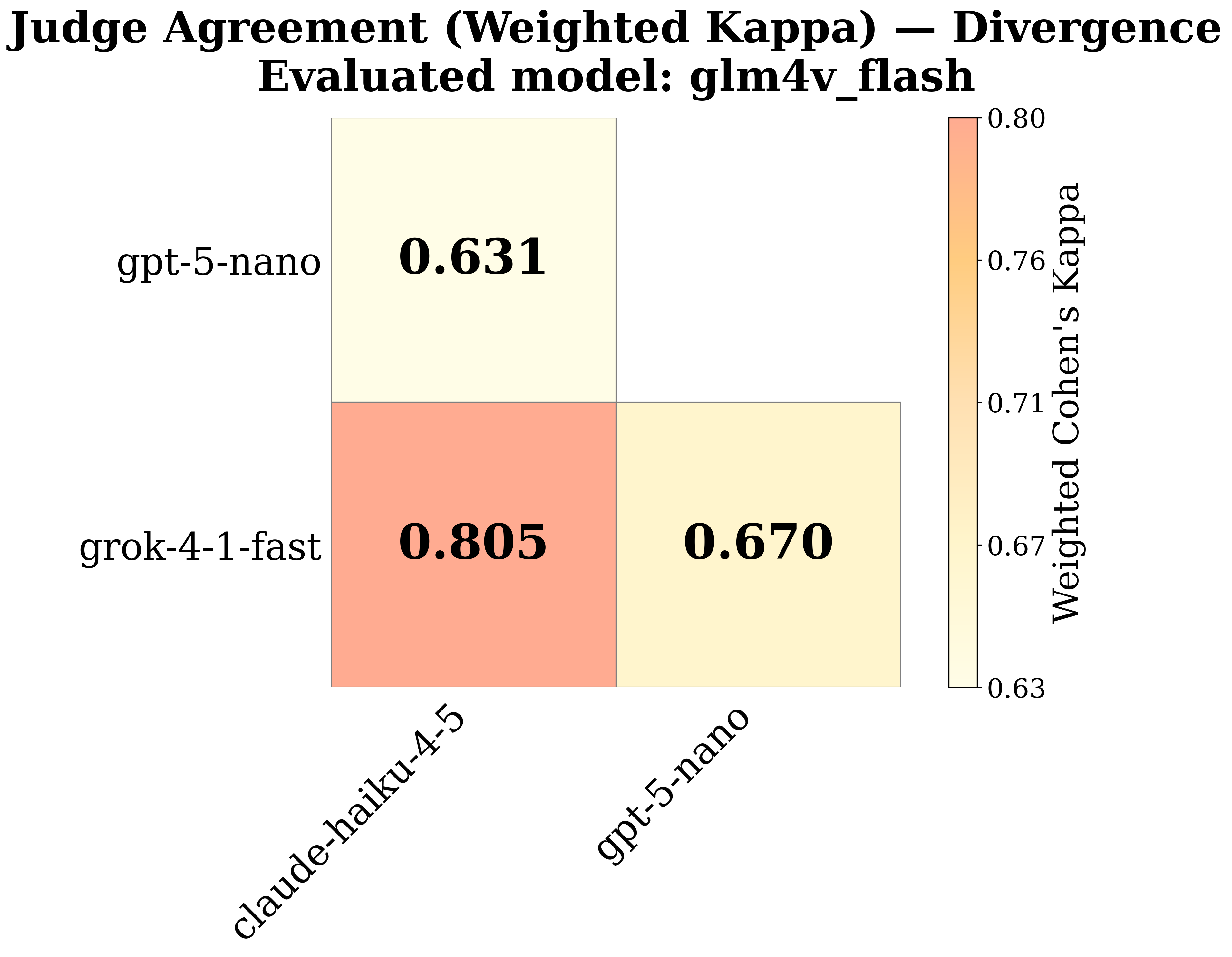}
    \end{subfigure}\hfill
    \begin{subfigure}[t]{0.31\textwidth}
        \centering
        \includegraphics[width=\linewidth]{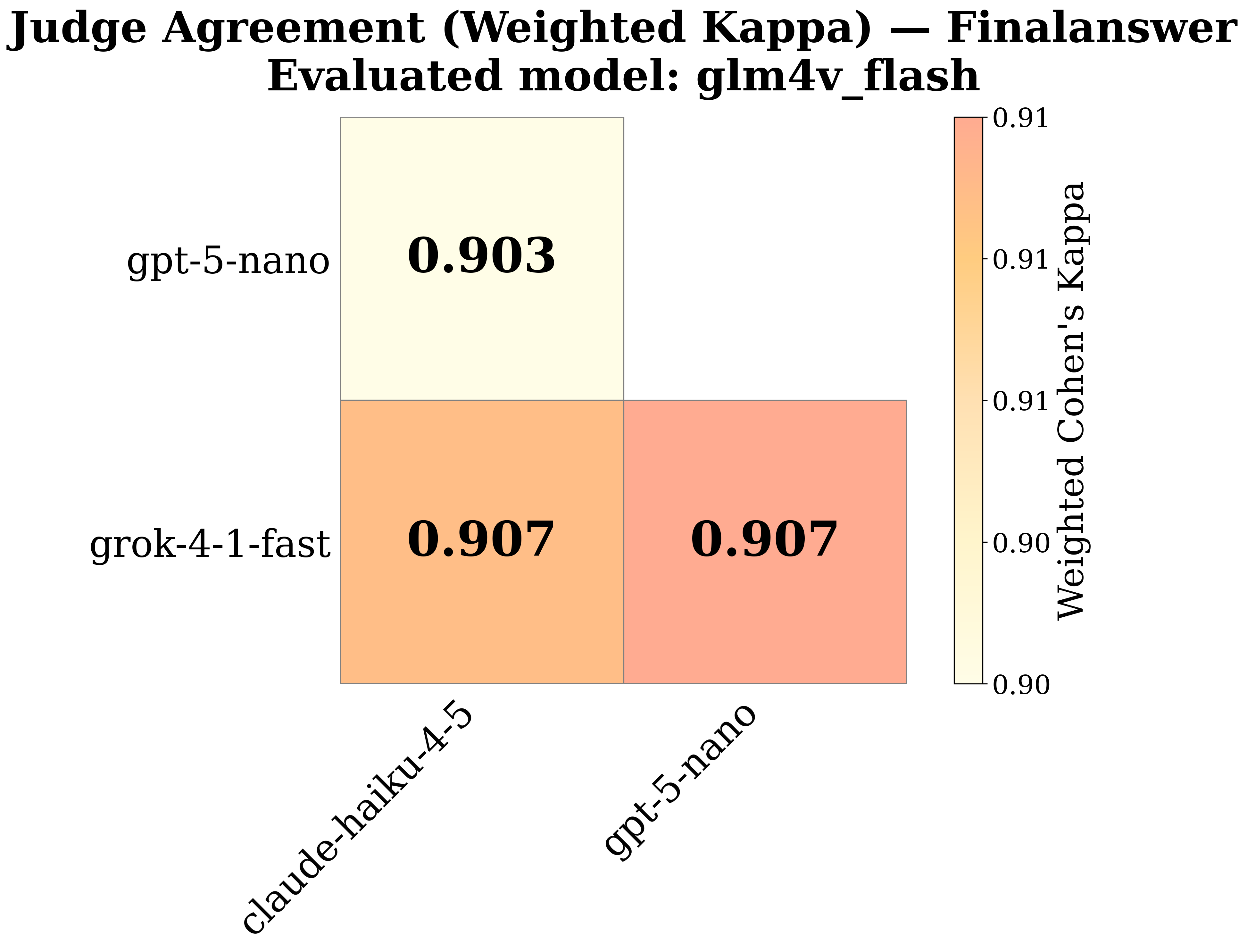}
    \end{subfigure}\hfill
    \begin{subfigure}[t]{0.31\textwidth}
        \centering
        \includegraphics[width=\linewidth]{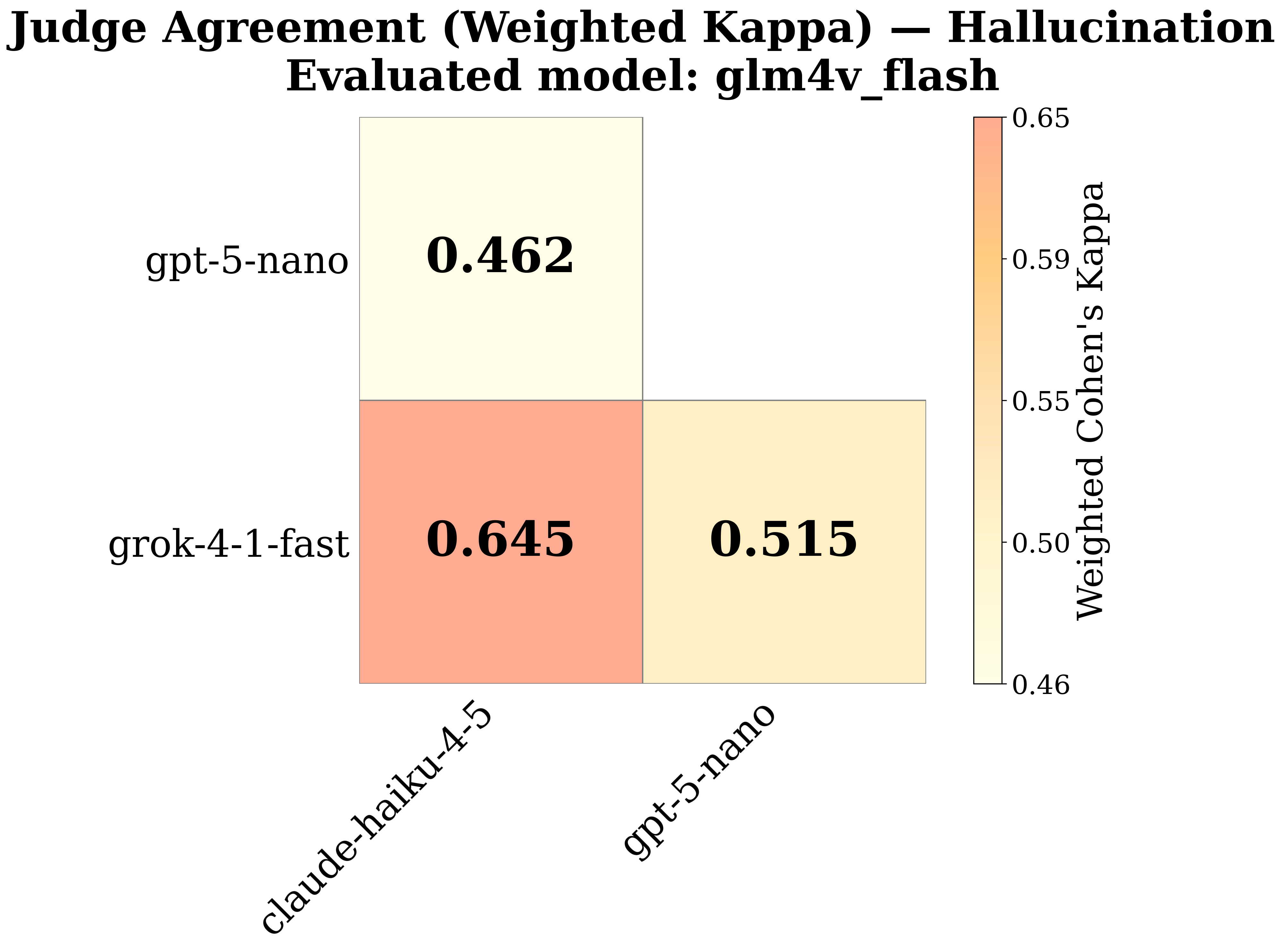}
    \end{subfigure}

    \vspace{0.5em}

    \begin{subfigure}[t]{0.31\textwidth}
        \centering
        \includegraphics[width=\linewidth]{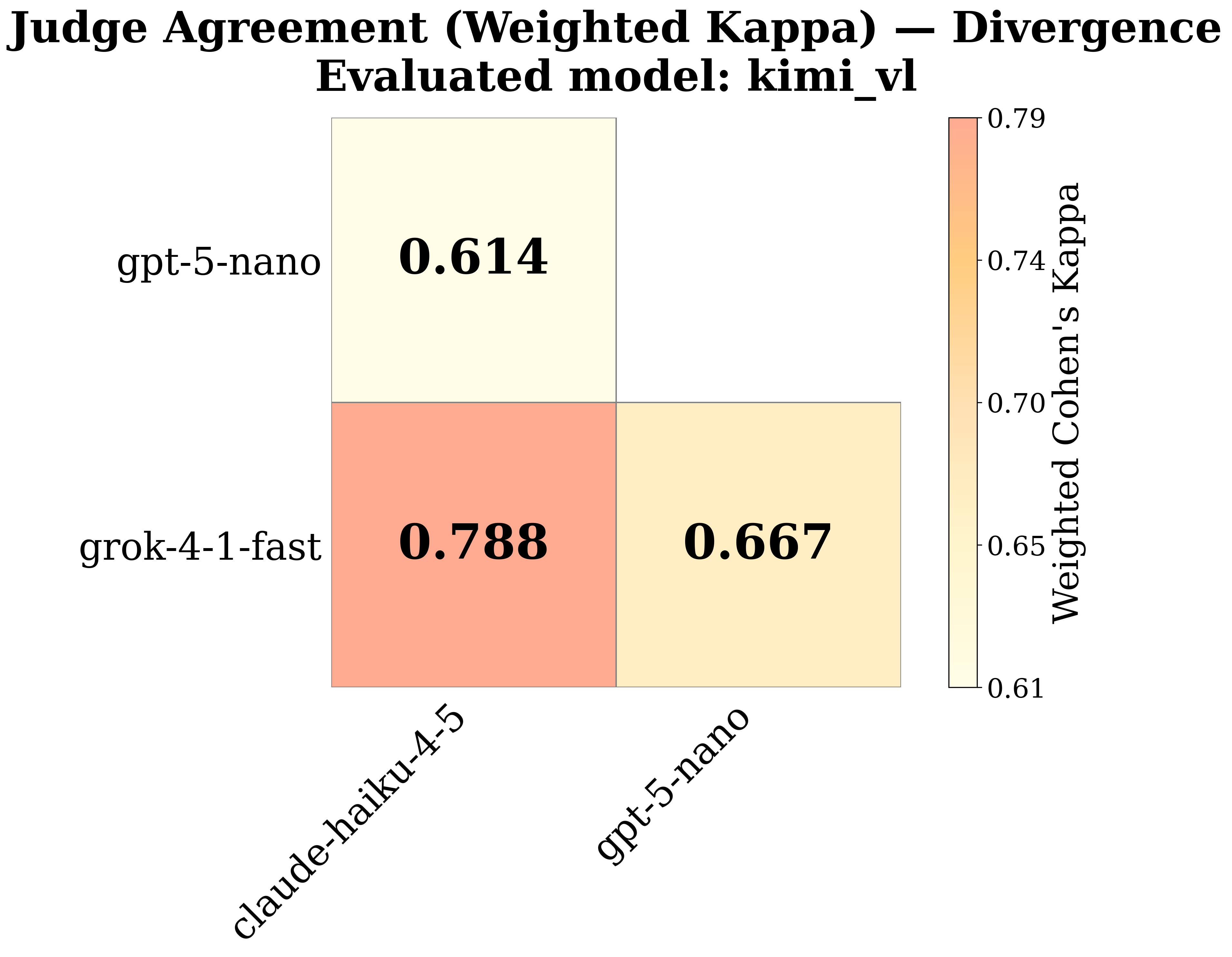}
    \end{subfigure}\hfill
    \begin{subfigure}[t]{0.31\textwidth}
        \centering
        \includegraphics[width=\linewidth]{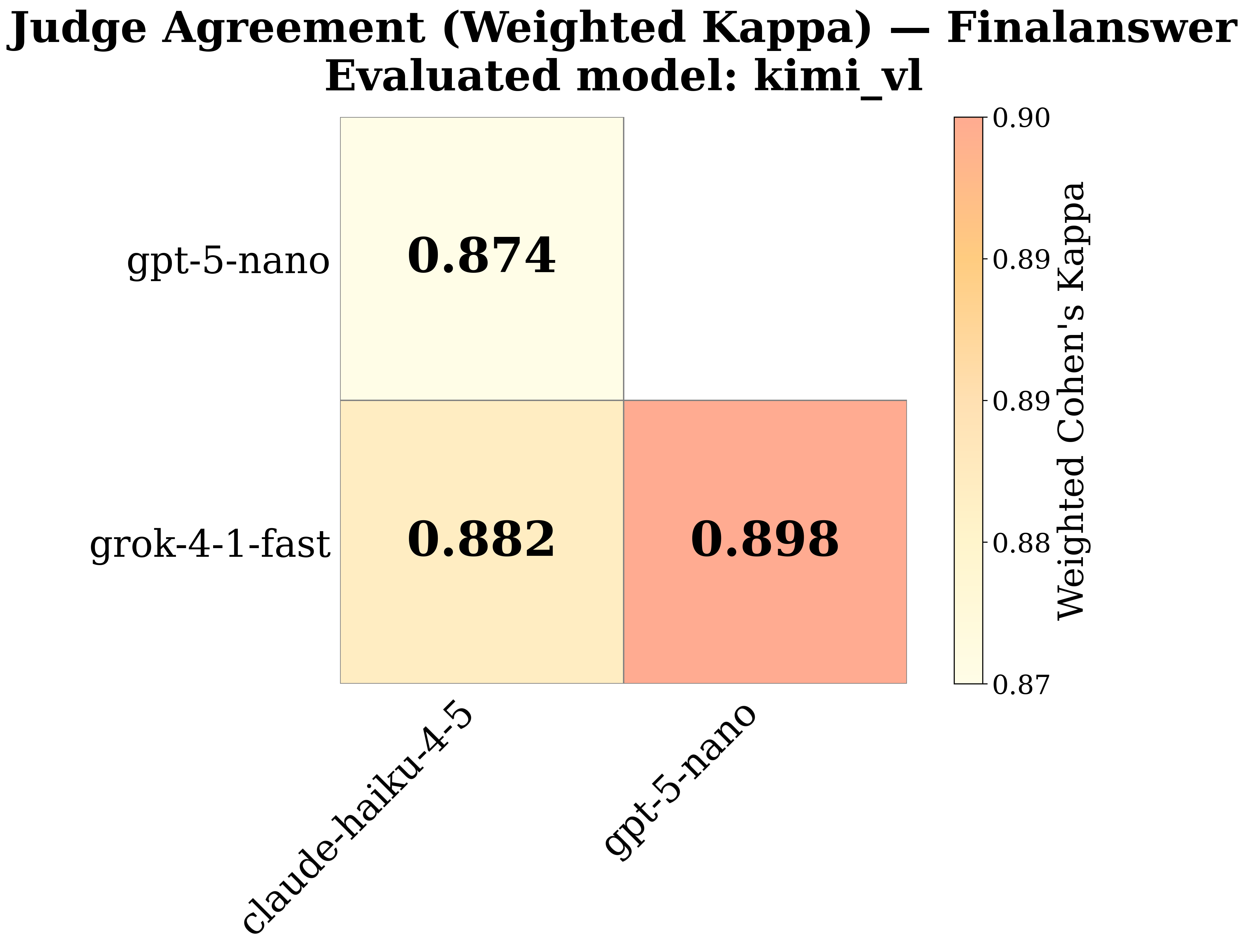}
    \end{subfigure}\hfill
    \begin{subfigure}[t]{0.31\textwidth}
        \centering
        \includegraphics[width=\linewidth]{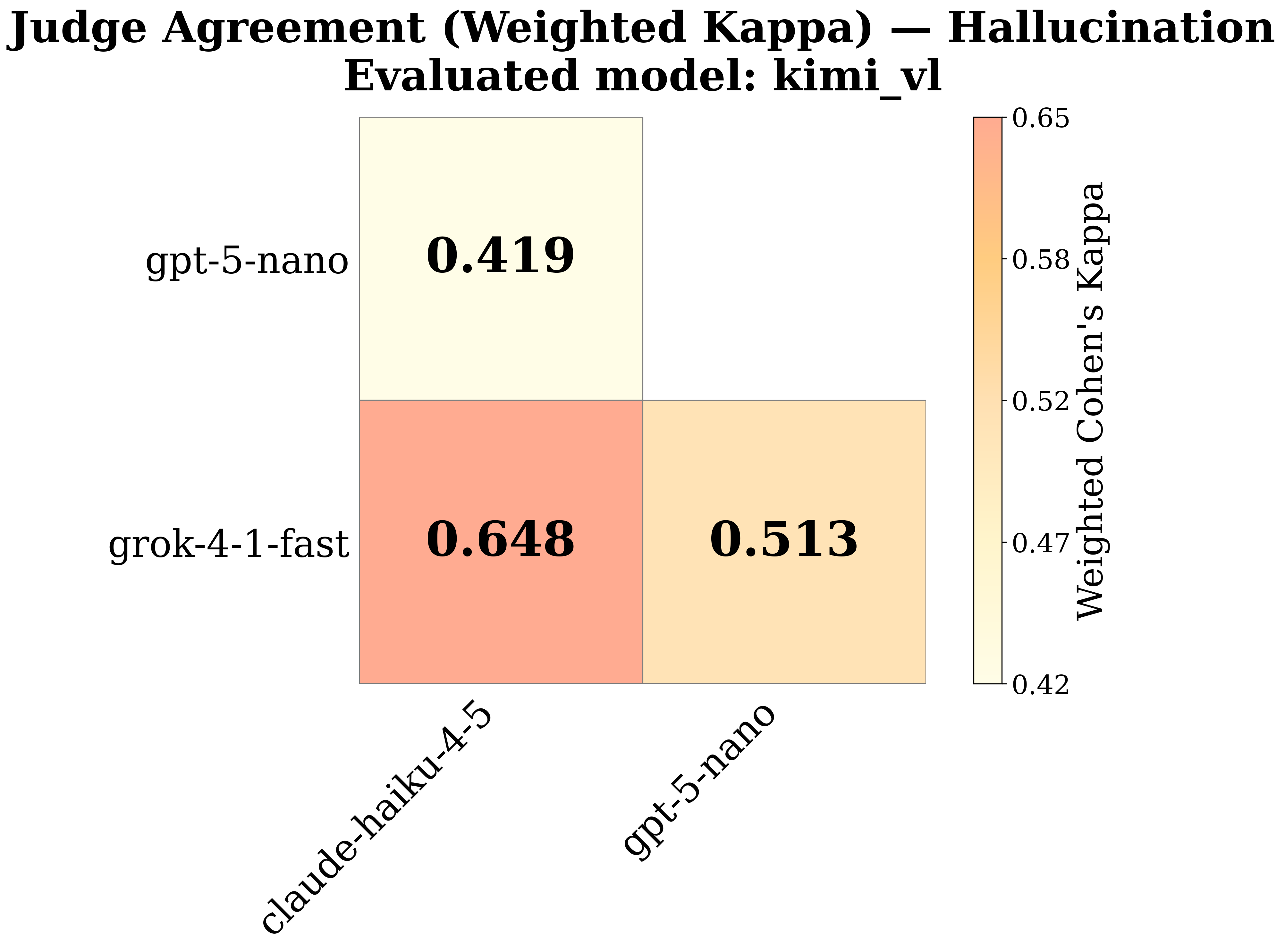}
    \end{subfigure}

    \vspace{0.5em}

    \begin{subfigure}[t]{0.31\textwidth}
        \centering
        \includegraphics[width=\linewidth]{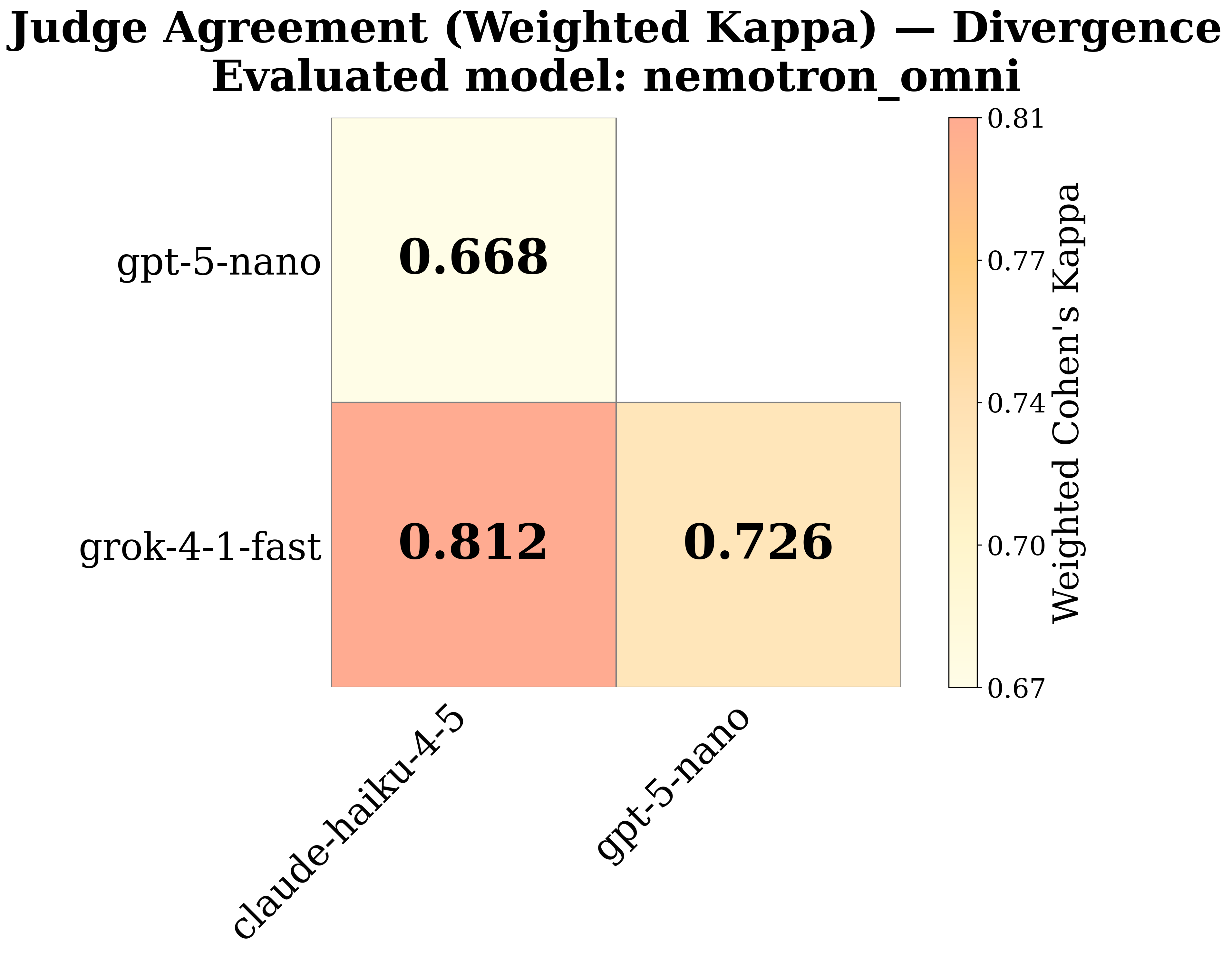}
    \end{subfigure}\hfill
    \begin{subfigure}[t]{0.31\textwidth}
        \centering
        \includegraphics[width=\linewidth]{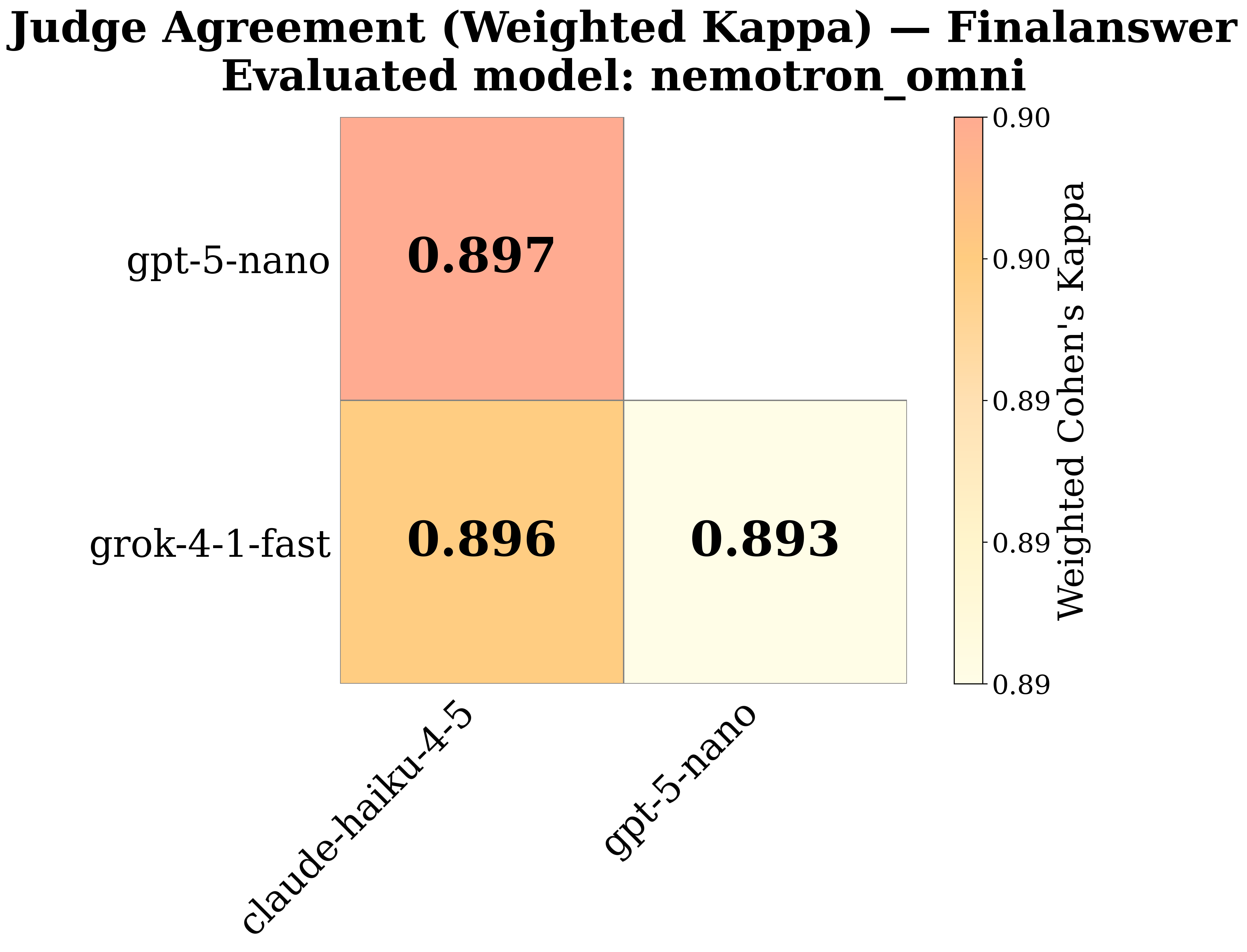}
    \end{subfigure}\hfill
    \begin{subfigure}[t]{0.31\textwidth}
        \centering
        \includegraphics[width=\linewidth]{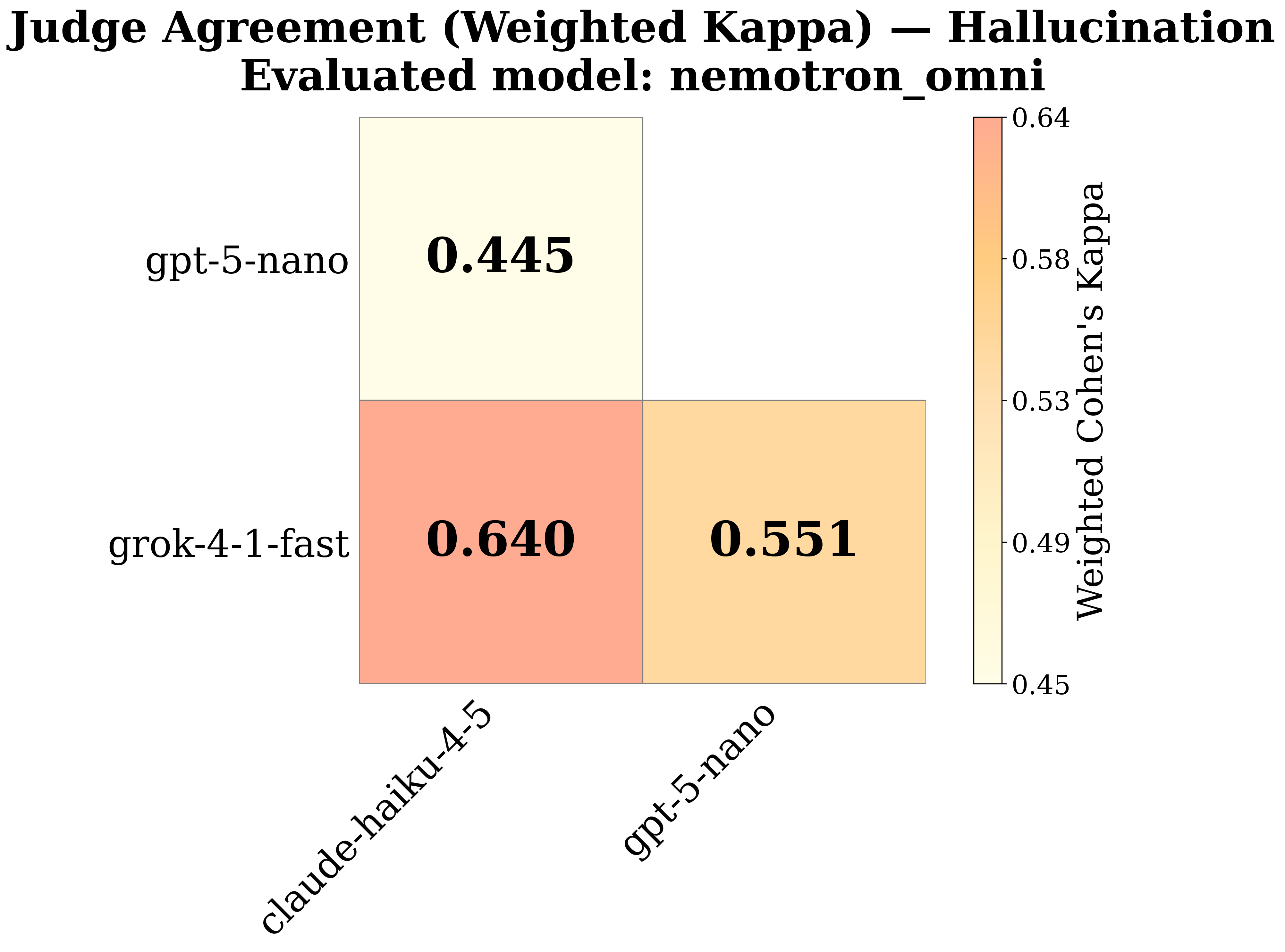}
    \end{subfigure}

    \caption{\textbf{Per-model inter-judge agreement measured by Cohen's $\kappa$ continued.}}
\end{figure*}

\subsection{Ablation \#1: Performance by Difficulty Level}
As part of dataset curation, we assign each QA pair a difficulty label following the procedure described in Section~\ref{sec:diff_rating_procedure}. To examine how model performance varies with question difficulty, we report per-facet results across the \textit{Easy}, \textit{Medium}, and \textit{Difficult} splits in Table~\ref{tab:score-by-difficulty}.

\begin{table*}[ht]
\centering
\scriptsize
\setlength{\tabcolsep}{6pt}
\renewcommand{\arraystretch}{1.12}
\begin{tabular}{llccc}
\toprule
\textbf{Facet} & \textbf{Model} & \textbf{Easy} & \textbf{Medium} & \textbf{Difficult} \\
\midrule

\multirow{12}{*}{Final Answer}
& ChatGPT 5.2              & 4.57 & 4.04 & 3.27 \\
& Claude 4.6 Sonnet        & 4.11 & 3.47 & 2.96 \\
& Gemini-3-Flash           & 4.50 & 4.10 & 3.38 \\
& Grok 4.1 Fast Reasoning  & 4.25 & 3.64 & 2.86 \\
& Nova-2-Lite              & 4.05 & 3.28 & 2.68 \\
& Qwen3-VL                 & 4.47 & 3.56 & 2.88 \\
& InternVL3.5              & 3.99 & 2.88 & 2.45 \\
& LLaVA-OneVision          & 3.47 & 2.21 & 2.00 \\
& Qwen3-Omni               & 4.37 & 3.44 & 2.78 \\
& GLM-4.6-V-Flash          & 4.35 & 3.49 & 2.76 \\
& Kimi-VL-A3B-Instruct     & 3.95 & 3.18 & 2.52 \\
& Nemotron-3-Nano-Omni-30B & 4.38 & 3.56 & 2.85 \\
\midrule

\multirow{12}{*}{Conciseness}
& ChatGPT 5.2              & 3.20 & 3.08 & 3.19 \\
& Claude 4.6 Sonnet        & 1.76 & 1.80 & 1.81 \\
& Gemini-3-Flash           & 4.64 & 4.64 & 4.64 \\
& Grok 4.1 Fast Reasoning  & 1.42 & 1.52 & 1.53 \\
& Nova-2-Lite              & 2.46 & 2.43 & 2.40 \\
& Qwen3-VL                 & 3.84 & 3.81 & 3.77 \\
& InternVL3.5              & 4.22 & 4.04 & 4.15 \\
& LLaVA-OneVision          & 4.64 & 4.58 & 4.66 \\
& Qwen3-Omni               & 4.26 & 4.23 & 4.24 \\
& GLM-4.6-V-Flash          & 2.48 & 2.50 & 2.54 \\
& Kimi-VL-A3B-Instruct     & 2.97 & 2.83 & 2.83 \\
& Nemotron-3-Nano-Omni-30B & 3.63 & 3.74 & 3.71 \\
\midrule

\multirow{12}{*}{Tokens}
& ChatGPT 5.2              & 326 & 355 & 343 \\
& Claude 4.6 Sonnet        & 560 & 595 & 566 \\
& Gemini-3-Flash           & 198 & 208 & 204 \\
& Grok 4.1 Fast Reasoning  & 657 & 683 & 622 \\
& Nova-2-Lite              & 492 & 596 & 534 \\
& Qwen3-VL                 & 275 & 288 & 294 \\
& InternVL3.5              & 240 & 269 & 262 \\
& LLaVA-OneVision          & 195 & 214 & 199 \\
& Qwen3-Omni               & 243 & 249 & 253 \\
& GLM-4.6-V-Flash          & 303 & 346 & 306 \\
& Kimi-VL-A3B-Instruct     & 301 & 365 & 338 \\
& Nemotron-3-Nano-Omni-30B & 201 & 220 & 212 \\

\midrule

\multirow{12}{*}{Divergence}
& ChatGPT 5.2              & 3.22 & 2.92 & 2.63 \\
& Claude 4.6 Sonnet        & 2.85 & 2.61 & 2.40 \\
& Gemini-3-Flash           & 3.29 & 3.00 & 2.74 \\
& Grok 4.1 Fast Reasoning  & 2.71 & 2.43 & 2.23 \\
& Nova-2-Lite              & 2.84 & 2.45 & 2.21 \\
& Qwen3-VL                 & 3.06 & 2.50 & 2.24 \\
& InternVL3.5              & 2.85 & 2.16 & 2.06 \\
& LLaVA-OneVision          & 2.42 & 1.77 & 1.77 \\
& Qwen3-Omni               & 3.23 & 2.63 & 2.35 \\
& GLM-4.6-V-Flash          & 3.00 & 2.51 & 2.21 \\
& Kimi-VL-A3B-Instruct     & 2.80 & 2.30 & 2.07 \\
& Nemotron-3-Nano-Omni-30B & 3.28 & 2.59 & 2.33 \\
\midrule

\multirow{12}{*}{Hallucination}
& ChatGPT 5.2              & 2.19 & 2.11 & 1.98 \\
& Claude 4.6 Sonnet        & 1.71 & 1.67 & 1.64 \\
& Gemini-3-Flash           & 2.37 & 2.23 & 2.12 \\
& Grok 4.1 Fast Reasoning  & 1.53 & 1.49 & 1.46 \\
& Nova-2-Lite              & 1.88 & 1.75 & 1.63 \\
& Qwen3-VL                 & 2.10 & 1.79 & 1.67 \\
& InternVL3.5              & 2.20 & 1.72 & 1.68 \\
& LLaVA-OneVision          & 2.00 & 1.51 & 1.53 \\
& Qwen3-Omni               & 2.46 & 2.00 & 1.81 \\
& GLM-4.6-V-Flash          & 2.25 & 1.95 & 1.79 \\
& Kimi-VL-A3B-Instruct     & 2.28 & 1.88 & 1.77 \\
& Nemotron-3-Nano-Omni-30B & 2.71 & 2.16 & 2.01 \\
\bottomrule
\end{tabular}
\caption{\textbf{Model performance by facet and difficulty level.} We group QA pairs into \textit{Easy}, \textit{Medium}, and \textit{Difficult} subsets using the difficulty-labeling procedure described in Section~\ref{sec:diff_rating_procedure}, and report per-facet scores for each model across these splits. Conciseness is a token-based quintile score, and Tokens reports the average output token count.}
\label{tab:score-by-difficulty}
\end{table*}

\subsection{Ablation \#2: Annotation Separation}
To examine how annotation source affects evaluation outcomes, we compare model performance across three subsets: human-only (\textbf{H}), human+AI (\textbf{H+AI}), and AI-only (\textbf{AI}). As shown in Table~\ref{tab:score-by-split}, performance remains broadly consistent across splits, while also revealing modest differences in answer quality and reasoning behavior depending on the annotation source.

\begin{table*}[ht]
\centering
\scriptsize
\setlength{\tabcolsep}{6pt}
\renewcommand{\arraystretch}{1.12}
\begin{tabular}{llccc}
\toprule
\textbf{Facet} & \textbf{Model} & \textbf{H} & \textbf{H+AI} & \textbf{AI} \\
\midrule

\multirow{12}{*}{Final Answer}
& GPT-5.2                  & 3.45 & 3.77 & 3.81 \\
& Claude 4.6 Sonnet        & 3.18 & 3.37 & 3.40 \\
& Gemini-3-Flash           & 3.44 & 3.83 & 3.88 \\
& Grok 4.1 Fast Reasoning  & 2.70 & 3.39 & 3.48 \\
& Nova-2-Lite              & 2.90 & 3.18 & 3.22 \\
& Qwen3-VL                 & 3.07 & 3.45 & 3.50 \\
& Qwen3-Omni               & 3.05 & 3.35 & 3.39 \\
& InternVL3.5              & 2.36 & 2.99 & 3.04 \\
& LLaVA-OneVision          & 2.08 & 2.45 & 2.48 \\
& GLM-4.6-V-Flash          & 3.00 & 3.33 & 3.37 \\
& Kimi-VL-A3B-Instruct     & 2.71 & 3.04 & 3.08 \\
& Nemotron-3-Nano-Omni-30B & 3.11 & 3.41 & 3.45 \\
\midrule

\multirow{12}{*}{Conciseness}
& GPT-5.2                  & 2.91 & 2.72 & 2.69 \\
& Claude 4.6 Sonnet        & 1.56 & 1.50 & 1.49 \\
& Gemini-3-Flash           & 4.09 & 4.47 & 4.44 \\
& Grok 4.1 Fast Reasoning  & 1.59 & 1.30 & 1.30 \\
& Nova-2-Lite              & 2.50 & 2.01 & 1.96 \\
& Qwen3-VL                 & 3.43 & 3.40 & 3.41 \\
& Qwen3-Omni               & 3.92 & 3.90 & 3.91 \\
& InternVL3.5              & 4.15 & 3.48 & 3.45 \\
& LLaVA-OneVision          & 4.47 & 4.25 & 4.24 \\
& GLM-4.6-V-Flash          & 2.02 & 2.52 & 2.58 \\
& Kimi-VL-A3B-Instruct     & 2.54 & 2.83 & 2.86 \\
& Nemotron-3-Nano-Omni-30B & 3.51 & 3.65 & 3.66 \\
\midrule

\multirow{12}{*}{Tokens}
& GPT-5.2                  & 339 & 342 & 343 \\
& Claude 4.6 Sonnet        & 553 & 571 & 574 \\
& Gemini-3-Flash           & 200 & 203 & 204 \\
& Grok 4.1 Fast Reasoning  & 653 & 644 & 643 \\
& Nova-2-Lite              & 435 & 545 & 559 \\
& Qwen3-VL                 & 298 & 287 & 285 \\
& Qwen3-Omni               & 279 & 291 & 293 \\
& InternVL3.5              & 250 & 283 & 285 \\
& LLaVA-OneVision          & 219 & 202 & 201 \\
& GLM-4.6-V-Flash          & 316 & 310 & 309 \\
& Kimi-VL-A3B-Instruct     & 315 & 337 & 340 \\
& Nemotron-3-Nano-Omni-30B & 207 & 210 & 210 \\

\midrule

\multirow{12}{*}{Divergence}
& GPT-5.2                  & 2.86 & 2.83 & 2.83 \\
& Claude 4.6 Sonnet        & 2.52 & 2.56 & 2.56 \\
& Gemini-3-Flash           & 2.85 & 2.93 & 2.94 \\
& Grok 4.1 Fast Reasoning  & 2.25 & 2.39 & 2.41 \\
& Nova-2-Lite              & 2.39 & 2.42 & 2.43 \\
& Qwen3-VL                 & 2.44 & 2.50 & 2.51 \\
& Qwen3-Omni               & 2.54 & 2.66 & 2.67 \\
& InternVL3.5              & 1.95 & 2.34 & 2.37 \\
& LLaVA-OneVision          & 1.80 & 1.96 & 1.97 \\
& GLM-4.6-V-Flash          & 2.40 & 2.48 & 2.48 \\
& Kimi-VL-A3B-Instruct     & 2.23 & 2.31 & 2.32 \\
& Nemotron-3-Nano-Omni-30B & 2.49 & 2.64 & 2.66 \\
\midrule

\multirow{12}{*}{Hallucination}
& GPT-5.2                  & 2.10 & 2.05 & 2.04 \\
& Claude 4.6 Sonnet        & 1.65 & 1.66 & 1.66 \\
& Gemini-3-Flash           & 2.17 & 2.20 & 2.20 \\
& Grok 4.1 Fast Reasoning  & 1.43 & 1.48 & 1.49 \\
& Nova-2-Lite              & 1.74 & 1.72 & 1.72 \\
& Qwen3-VL                 & 1.79 & 1.80 & 1.80 \\
& Qwen3-Omni               & 1.88 & 2.04 & 2.06 \\
& InternVL3.5              & 1.65 & 1.87 & 1.89 \\
& LLaVA-OneVision          & 1.57 & 1.67 & 1.67 \\
& GLM-4.6-V-Flash          & 1.86 & 1.94 & 1.95 \\
& Kimi-VL-A3B-Instruct     & 1.82 & 1.93 & 1.94 \\
& Nemotron-3-Nano-Omni-30B & 2.07 & 2.22 & 2.24 \\
\bottomrule
\end{tabular}
\caption{\textbf{Model performance by facet and annotation source.} We report per-facet scores for each model on the human-only (\textbf{H}), human+AI (\textbf{H+AI}), and AI-only (\textbf{AI}) subsets. Conciseness is a token-based quintile score, and Tokens reports the average output token count.}
\label{tab:score-by-split}
\end{table*}

\begin{table}[p]
\centering
\caption{Taxonomy percentages of the QA pairs from our dataset}
\label{tab:taxonomy_percentages_compact}
\begin{tabular}{llll}
\hline
Section & Top category & Subcategory & Percentage (\%) \\
\hline
Top-level & Information Query & -- & 69.26 \\
Top-level & Sports & -- & 28.80 \\
Top-level & Gaming & -- & 1.05 \\
Top-level & Shopping & -- & 0.33 \\
Top-level & Sentiment & -- & 0.22 \\
Top-level & Egocentric Agent & -- & 0.19 \\
Top-level & Social Interaction & -- & 0.15 \\
\hline
Subcategory & -- & Sports \& Adventure & 29.67 \\
Subcategory & -- & Movies, TV \& Animation & 28.31 \\
Subcategory & -- & Pets \& Animals & 19.05 \\
Subcategory & -- & Science \& Knowledge & 10.64 \\
Subcategory & -- & Life \& Practical Skills & 7.19 \\
Subcategory & -- & Humanities \& Society & 2.48 \\
Subcategory & -- & Technology \& Gaming & 1.39 \\
Subcategory & -- & Social Trends \& Reactions & 0.70 \\
Subcategory & -- & Business \& Commerce & 0.57 \\
\hline
Joint & Sports & Sports \& Adventure & 28.80 \\
Joint & Information Query & Movies, TV \& Animation & 28.26 \\
Joint & Information Query & Pets \& Animals & 19.04 \\
Joint & Information Query & Science \& Knowledge & 10.64 \\
Joint & Information Query & Life \& Practical Skills & 7.00 \\
Joint & Information Query & Humanities \& Society & 2.48 \\
Joint & Gaming & Technology \& Gaming & 1.05 \\
Joint & Information Query & Sports \& Adventure & 0.87 \\
Joint & Information Query & Social Trends \& Reactions & 0.41 \\
Joint & Information Query & Technology \& Gaming & 0.33 \\
Joint & Shopping & Business \& Commerce & 0.33 \\
Joint & Information Query & Business \& Commerce & 0.25 \\
Joint & Egocentric Agent & Life \& Practical Skills & 0.18 \\
Joint & Sentiment & Social Trends \& Reactions & 0.15 \\
Joint & Social Interaction & Social Trends \& Reactions & 0.14 \\
Joint & Sentiment & Movies, TV \& Animation & 0.05 \\
Joint & Egocentric Agent & Technology \& Gaming & 0.01 \\
Joint & Sentiment & Pets \& Animals & 0.01 \\
Joint & Social Interaction & Life \& Practical Skills & 0.01 \\
\hline
\end{tabular}
\end{table}

\end{document}